\documentclass[11pt]{article}

\usepackage[a4paper,margin=1in]{geometry}
\usepackage[T1]{fontenc}
\usepackage[utf8]{inputenc}

\usepackage{graphicx}
\usepackage{multirow}
\usepackage{amsmath,amssymb,amsfonts}
\usepackage{amsthm}
\usepackage{mathrsfs}
\usepackage[title]{appendix}
\usepackage{xcolor}
\usepackage{textcomp}
\usepackage{booktabs}
\usepackage{tabularx}
\usepackage{array}
\usepackage{adjustbox}
\usepackage{caption}
\usepackage{algorithm}
\usepackage{algorithmicx}
\usepackage{algpseudocode}
\usepackage{listings}
\usepackage{authblk}
\usepackage[numbers,sort&compress]{natbib}
\usepackage{hyperref}

\hypersetup{
	colorlinks=true,
	linkcolor=blue,
	citecolor=blue,
	urlcolor=blue
}

\newcommand{\thinhline}{\specialrule{0.3pt}{0.1ex}{0.1ex}}

\theoremstyle{plain}

\theoremstyle{definition}

\theoremstyle{remark}

\title{Astra: a generalizable report generation foundation model for 3D computed tomography}

\author[1]{Zhuhao Wang\textsuperscript{\textdagger}}
\author[2]{Fang Chen\textsuperscript{\textdagger,*}}
\author[3,4]{Chaohui Yu\textsuperscript{\textdagger}}
\author[1]{Zihan Li\textsuperscript{\textdagger}}
\author[1]{Yuchao Zheng}
\author[3,4]{Jing Wang}
\author[8]{Xuan Yang}
\author[1]{Jia Guo}
\author[5]{Zhenlu Yang}
\author[5]{Xingju Zheng}
\author[1]{Yihua Sun}
\author[1]{Haojie Han}
\author[6]{Xiaoxiao Qin}
\author[6]{Zhan Feng}
\author[6]{Wenbo Xiao}
\author[7]{Chao Zhu}
\author[7]{Yuehua Li}
\author[1]{Shipeng Zhang}
\author[3,4,9]{Hao Luo}
\author[5]{Yunsong Peng\textsuperscript{*}}
\author[3]{Fan Wang\textsuperscript{*}}
\author[1,2]{Hongen Liao\textsuperscript{*}}

\affil[1]{School of Biomedical Engineering, Tsinghua University}
\affil[2]{School of Biomedical Engineering, Shanghai Jiao Tong University}
\affil[3]{DAMO Academy, Alibaba Group}
\affil[4]{Hupan Laboratory}
\affil[5]{Department of Radiology, Guizhou Provincial People's Hospital}
\affil[6]{Department of Radiology, The First Affiliated Hospital, Zhejiang University School of Medicine}
\affil[7]{Department of Radiology, Shanghai Sixth People's Hospital Affiliated to Shanghai Jiao Tong University School of Medicine}
\affil[8]{Department of Biomedical Engineering, National University of Singapore}
\affil[9]{College of Computer Science and Technology, Zhejiang University}

\affil[]{\footnotesize
	\textsuperscript{\textdagger}These authors contributed equally to this work.\\
	\textsuperscript{*}Correspondence: 
	chen-fang@sjtu.edu.cn; 
	pys@mail.ustc.edu.cn;\\
	fan.w@alibaba-inc.com; 
	liao@tsinghua.edu.cn
}

\date{}

\begin{document}
	
	\maketitle

	\begin{abstract}
	Interpreting computed tomography (CT) requires review of hundreds of volumetric slices and remains time-intensive and expertise-dependent. Automated CT report generation offers a promising route to improving clinical efficiency, yet the field still lacks a generalizable CT report generation foundation model that supports multi-region reporting and remains robust across external real-world cohorts. Intrinsic inconsistencies in reporting style and diagnostic terminology across cohorts make naive joint training difficult. 
	Here we present Astra, a generalizable CT report generation foundation model developed on 90,678 thoracoabdominal CT-report pairs collected from five sites worldwide (CTRgDB), comprising 353,671 abnormalities spanning eight organ systems. By harmonizing report style and further refining diagnostic consistency via reinforcement learning, Astra achieves style-consistent and diagnostically accurate report generation across diverse anatomical regions and institutions. Evaluated on CTRgDB and six external cohorts, Astra achieves state-of-the-art performance with a 38.4\% average improvement in fine-grained diagnostic metrics (P $<$ 0.001). Deployed at external clinical sites without any site-specific fine-tuning, Astra accelerated chest report drafting by 29.6\% and improved abdominal report completeness by 11.3\% among junior and mid-level radiologists (P $<$ 0.001). Furthermore, Astra demonstrates broad utility as a foundation for CT AI development, improving downstream diagnostic performance and scaling vision-language pretrain through high-quality report synthesis. Overall, Astra serves as a broadly accessible clinical assistant and a pivotal infrastructure for the next generation of AI-powered healthcare. The code for Astra is publicly available at \url{https://github.com/zh-Wang-Med/Astra}.
	\end{abstract}

	\section{Introduction}
	Computed tomography (CT) is a cornerstone of modern clinical imaging. Its ability to resolve complex anatomy in three dimensions makes it indispensable for emergency trauma assessment, oncological staging and cardiovascular evaluation~\cite{otoni2017role}. Worldwide, an estimated 300 million CT examinations are performed annually~\cite{schockel2020developments}, of which thoracic and abdominal scans account for 25\% and 45\%, respectively~\cite{kanal2017us}.
	However, accurate interpretation of multi-organ lesions across hundreds of CT slices demands substantial expertise and considerable time, with an average of 20 minutes required to generate a single diagnostic report~\cite{udare2022radiologist}. Such complexity not only burdens individual radiologists but also entails a steep learning curve. Consequently, as imaging data grows exponentially, the global supply of radiologists has severely lagged behind clinical demand, as evidenced by China's ratio of merely one radiologist per 70,000 people~\cite{meng2023growing}.

	Artificial intelligence has emerged as a powerful tool for automated CT interpretation and diagnosis. Numerous deep learning models have demonstrated remarkable performance across a range of tasks including disease classification, grading, and prognosis~\cite{zhang2020clinically,hu2025ai,hu2025aa,cao2023large,milam2023current}.
	Nevertheless, most of these systems remain narrow, single-task tools that rely on dense annotations for a limited set of disease categories, making them difficult to scale and unable to replicate the comprehensive diagnostic workflow of a radiologist. Automated CT report generation offers a compelling solution to this challenge, capable of synthesizing comprehensive radiological findings from scans into structured, clinically interpretable narratives, while requiring only readily available radiology reports for supervision rather than costly fine-grained annotations.

	Recent advances in multimodal large language models (MLLMs) have driven progress in CT report generation, giving rise to broad paradigms. The first fine-tunes MLLMs on cohort-specific CT report generation tasks, forming expert models~\cite{braingpt,reg2rg,hamamci2024ct2rep,diallama,deng2025mvketr,ctagrg,ctgraph,BTB3D,3dctgpt}. By incorporating specialized architectures~\cite{deng2025mvketr}, segmentation~\cite{reg2rg} or classification priors~\cite{ctagrg}, and external medical knowledge~\cite{ctgraph}, these models achieve strong performance when training and test distributions are closely aligned. However, they are typically restricted to single anatomical regions and development cohorts. Most existing efforts focus on chest CT~\cite{reg2rg,ctgraph,deng2025mvketr}, whereas abdominal CT remains underexplored despite its greater anatomical complexity and broader disease spectrum. Extending expert models across regions often requires separate region-specific training, increasing data and deployment costs. Their robustness to unseen institutions also remains uncertain, owing to visual domain shifts in scanners and contrast phases, as well as substantial variation in reporting style, conventions and descriptive granularity~\cite{delbrouck2025automated}. Thus, although expert models perform well on narrowly defined benchmarks, their single-region and single-cohort design limits their utility in real-world clinical practice and constrains broader integration with other AI systems.
	
	The second paradigm leverages general-purpose MLLMs, such as GPT-4~\cite{gpt4} and Gemini~\cite{team2023gemini}, by representing CT volumes as sequences of 2D slices and generating reports through in-context learning or prompting. Radiology generalist models, including RadFM~\cite{radfm} and M3D~\cite{bai2024m3d}, and medical generalist models, such as HuluMed~\cite{jiang2025hulu}, further incorporate 3D CT inputs and domain-focused training on radiology or medical datasets. Compared with expert models, these generalist models offer the potential for multi-region reporting and improved robustness across heterogeneous cohorts. Nevertheless, this potential remains difficult to realize in practice. Naively pooling heterogeneous multi-center data can introduce conflicting textual supervision, as institutions often differ in report style and diagnostic terminology. Specifically, visually similar findings may be organized in different orders or expressed using inconsistent clinical terms. Moreover, report generation is usually optimized as one task among many vision-language objectives, where inter-task conflicts may compromise generation quality. Together, these limitations highlight the need for a dedicated 3D CT report generation model that supports multi-region reporting while remaining robust and generalizable across diverse, real-world clinical cohorts.

	To address this critical need, we introduce Astra, a generalizable report generation foundation model for 3D CT. Astra is trained on CTRgDB, a large-scale and harmonized database comprising 90,678 thoracoabdominal CT–report pairs and 353,671 abnormalities spanning eight organ systems. Through supervised fine-tuning and reinforcement learning, Astra learns to generate style-consistent and fine-grained CT reports across anatomical regions and institutions. We systematically evaluate Astra across three dimensions: methodological generalizability, clinical utility, and foundational extensibility. Methodological generalizability is established through rigorous benchmarking across CTRgDB and six independent out-of-distribution clinical cohorts, where Astra consistently surpasses existing state-of-the-art architectures. Clinical utility is corroborated via a real-world human-AI collaboration study, where Astra, on average, accelerates chest CT report drafting by 29.6\% and enhances abdominal CT report completeness by 11.3\% across diverse levels of clinical expertise. 
	
	Finally, foundational extensibility is evidenced by the capacity to catalyze broader AI development paradigms, where Astra facilitates diagnostic model development through ensemble strategies with pretrained vision encoder and scales vision-language pretraining by synthesizing diagnostic reports for previously unreported scans. Taken together, Astra represents a transformative step toward generalizable foundation models for 3D CT report generation, moving beyond the traditional single-cohort paradigm and highlighting their broader potential across clinical and research settings.

	\section{Results}
	
	\begin{center}
		\includegraphics[width=0.99\textwidth]{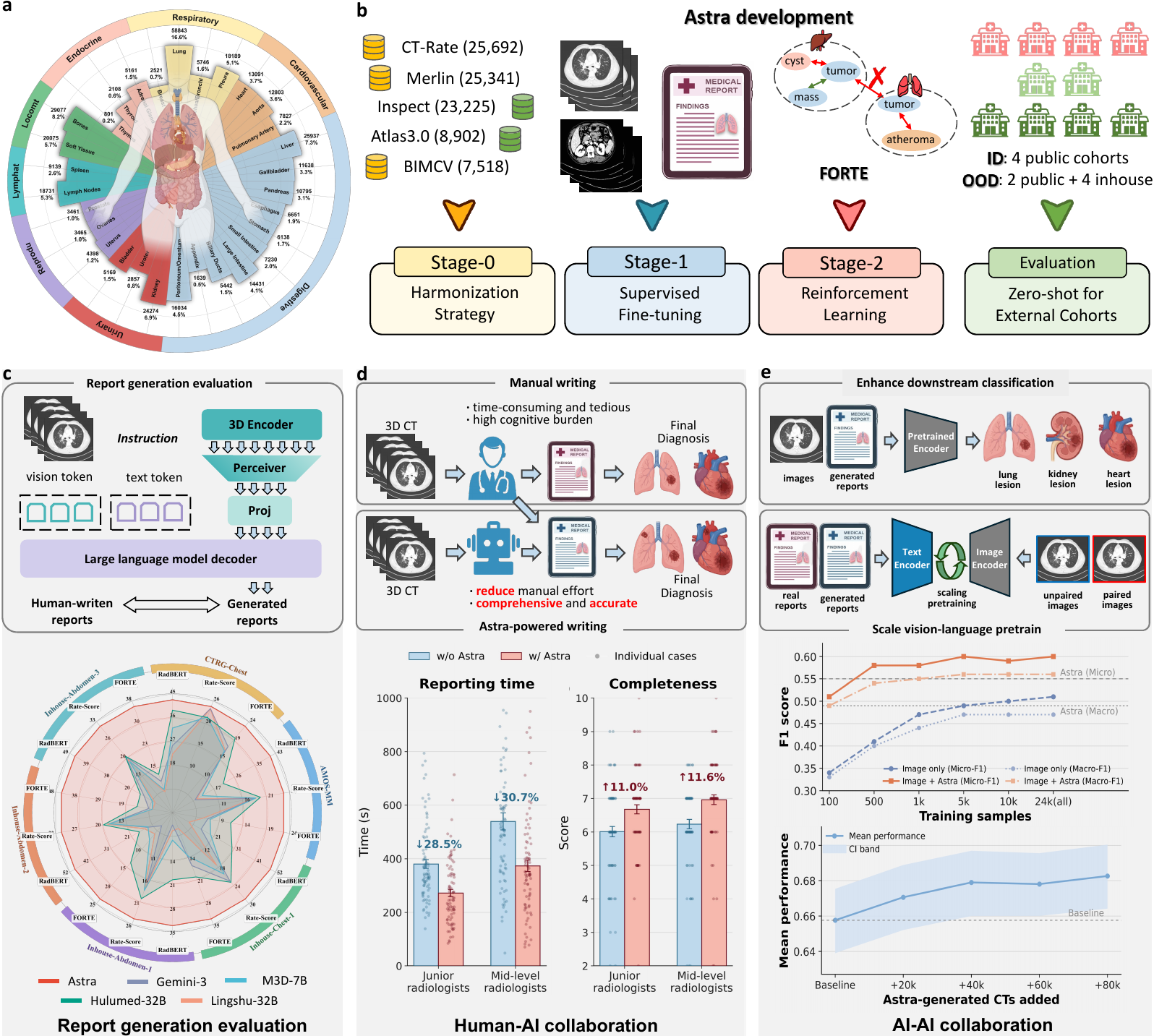}
		\captionof{figure}{\textbf{Overview of Astra.}
			\textbf{a}, Distribution of 353,671 abnormalities across eight organ systems in CTRgDB.
			\textbf{b}, Overview of Astra development and evaluation, including database construction, supervised fine-tuning, reinforcement learning with region-wise attribute-level rewards, and evaluation across ten cohorts spanning anatomical regions, clinical settings and institutions.
			\textbf{c}, Zero-shot report generation pipeline of Astra and its quantitative performance across six external validation cohorts.
			\textbf{d}, Human--AI collaboration showing that radiologist revision of Astra-generated drafts improved chest CT reporting efficiency and abdominal CT report quality.
			\textbf{e}, AI--AI collaboration, in which Astra enhances pretrained CT encoders for downstream classification and generates pseudo-labels for CT scans without paired reports to scale CT vision-language pretraining.} 
	\end{center}
	
	\subsection{Overview of Astra framework}
	Astra was built around two components: a region-wise, abnormality-focused report harmonization strategy to unify reporting style, and reinforcement learning-based post-training to mitigate terminology heterogeneity and enhance fine-grained diagnostic capability (Fig.~1b). Report harmonization was performed using a state-of-the-art closed-source large language model, leveraging its strong capability in medical text processing~\cite{kirchler2026large_llmprocess1,sandmann2025benchmark_llmprocess2,tordjman2025comparative_llmprocess3,hu2024zero_llmprocess4,lopez2025clinical_llmprocess5}. Following Reg2RG~\cite{reg2rg} and Merlin~\cite{merlin}, we defined 10 anatomical regions for chest CT and 13 for abdominal CT. The large language model extracted region-specific diagnostic descriptions from original reports and reorganized them into a fixed anatomical order, reducing variability arising from institution-specific templates and reporting sequences. It was further used to remove negative mentions and clinician communications, and to rewrite comparative statements. These elements are often driven by institutional templates~\cite{nguyen2023pragmatic}, patient complaints, examination indications or prior imaging studies, rather than findings directly visible on the current CT scan. Their removal or rewriting therefore focused textual supervision on current imaging abnormalities and reduced non-visual textual noise during training. Representative chest and abdominal CT report cases are shown in Supplementary Fig.~A1. This process yielded CTRgDB, a harmonized CT report generation database with a fixed template and consistent reporting style. Supervised fine-tuning on CTRgDB enabled Astra to acquire preliminary cross-region and cross-cohort generalizability while generating stylistically consistent CT reports.
	
	The reinforcement learning-based post-training stage was implemented using group relative policy optimization (GRPO)~\cite{guo2025deepseek} to reduce interference from inconsistent diagnostic terminology and further improve fine-grained abnormality characterization. Specifically, for each CT input, Astra generated multiple candidate reports through rollouts, each scored by a reward function and group-relative advantages were then computed to steer the model toward clinically preferred responses. 
	The core of this stage was the reward design. Inspired by FORTE~\cite{braingpt}, our reward function assessed whether abnormalities are correctly described along four attributes: Degree (size and intensity), Landmark (anatomical location), Feature (disease traits and primary imaging findings), and Impression (the final diagnosis).
	Within each anatomical region, these attribute-specific keywords were extracted from both the generated and ground-truth reports and compared through curated synonym mappings, enabling clinically equivalent expressions to be matched despite differences in wording. 
	We updated the keyword scope and synonym mapping tables for both chest CT and abdominal CT reports, and introduced a new region-based keyword matching strategy to reduce keyword mismatching. Detailed implementation is provided in the Methods section.
	Overall, this reward function avoids penalizing correct diagnoses expressed with synonymous terminology, a limitation of conventional supervised fine-tuning, and guides Astra toward more precise region-level and lesion-level abnormality characterization.
	
	\begin{figure}[!htbp]
		\centering
		\includegraphics[width=0.9\textwidth]{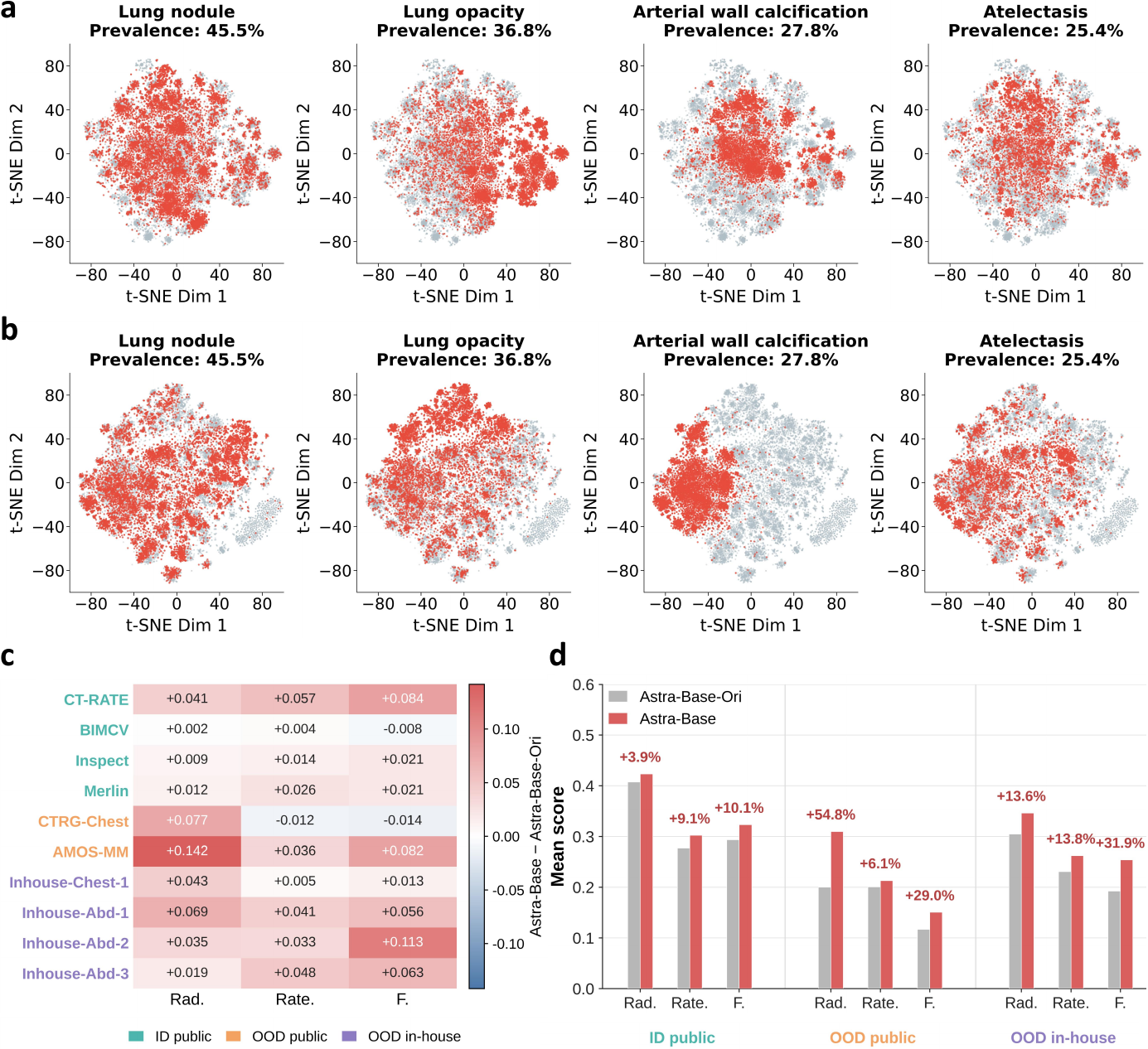}
		\caption{\textbf{Evaluation of the harmonization strategy.} 
			\textbf{a}, t-SNE visualization of report text embeddings from the original CT-Rate reports. Embeddings were extracted using Qwen3-Embedding for the four most prevalent diseases. Red and grey points denote positive and negative cases, respectively.
			\textbf{b}, t-SNE visualization of report text embeddings from harmonized CT-Rate reports using Qwen3-Embedding. The harmonization strategy improved the semantic separation between positive and negative cases.
			\textbf{c,d}, Performance comparison of Astra-Base-Ori, trained with original reports, and Astra-Base, trained with harmonized reports, across four in-distribution public cohorts, two out-of-distribution public cohorts and four out-of-distribution in-house cohorts. Rad., Rate. and F. denote RadBERT, RaTE-Score and FORTE, respectively. Report harmonization improved clinical accuracy, particularly on out-of-distribution cohorts.}
		\label{fig:harmonization}
	\end{figure}
	\subsection{CTRgDB lays the foundation for generalizable CT report generation}
	To support Astra development, we curated CTRgDB, a large-scale 3D CT report generation dataset integrating high-quality volumetric CT scans and paired diagnostic reports from Merlin~\cite{merlin}, Atlas3.0~\cite{atlas}, CT-Rate~\cite{ctrate}, Inspect~\cite{huang2023inspect}, and BIMCV~\cite{chen2024bimcv}. CTRgDB comprised 90,678 3D CT scan--report pairs spanning diverse thoracic and abdominal diseases, with 353,671 abnormalities across eight major organ systems. Multi-organ, multi-region and multi-disease involvement was common, with a mean of 3.9 affected organs per case (supplementary Fig.~A2--A6). This complexity was mirrored by the corresponding reports: abdominal CT reports contained, on average, 4.85 lesion-degree descriptors, 9.97 anatomical landmarks and 7.33 morphological descriptors, whereas chest CT reports contained 4.28, 7.75 and 4.45, respectively (Supplementary Fig.~A7). These characteristics establish CTRgDB as a rich training resource for fine-grained 3D CT report generation, extending beyond isolated lesion recognition toward comprehensive abnormality description.
	
	To assess whether CTRgDB provides stronger supervision than naive data pooling, we compared models fine-tuned on the harmonized CTRgDB with those trained on a simple mixture of the same source datasets. In in-distribution evaluations, CTRgDB yielded the largest gains on datasets dominated by template-driven content, especially negative findings and boilerplate descriptions. On CT-Rate, for example, training on CTRgDB improved Rate-Score~\cite{zhao2024ratescore} by 20.2\% and FORTE~\cite{braingpt} by 33.3\% over training on the uncurated mixture (Fig.~2c), whereas gains were smaller on Inspect and BIMCV, where template-driven negative mentions were less prevalent. The benefit also extended to out-of-distribution settings. Across two external public datasets and four real-world clinical cohorts, CTRgDB-trained models consistently improved both global disease-level diagnosis and fine-grained abnormality characterization (supplementary Tab.~B1--B4). To further characterize the effect of harmonization, we extracted textual features from harmonized and original CT-Rate reports using Qwen3-Embedding~\cite{zhang2025qwen3_embed} and visualized them with t-SNE~\cite{tsne}. For abnormalities such as lung opacity, arterial wall calcification and atelectasis, harmonized reports formed more compact and better-separated semantic clusters (Fig.~2a,b). Together, these results indicate that CTRgDB converts heterogeneous clinical reports into scalable supervision for generalizable 3D CT report generation.
	
	\begin{figure}[!htbp]
		\centering
		\includegraphics[width=0.99\textwidth]{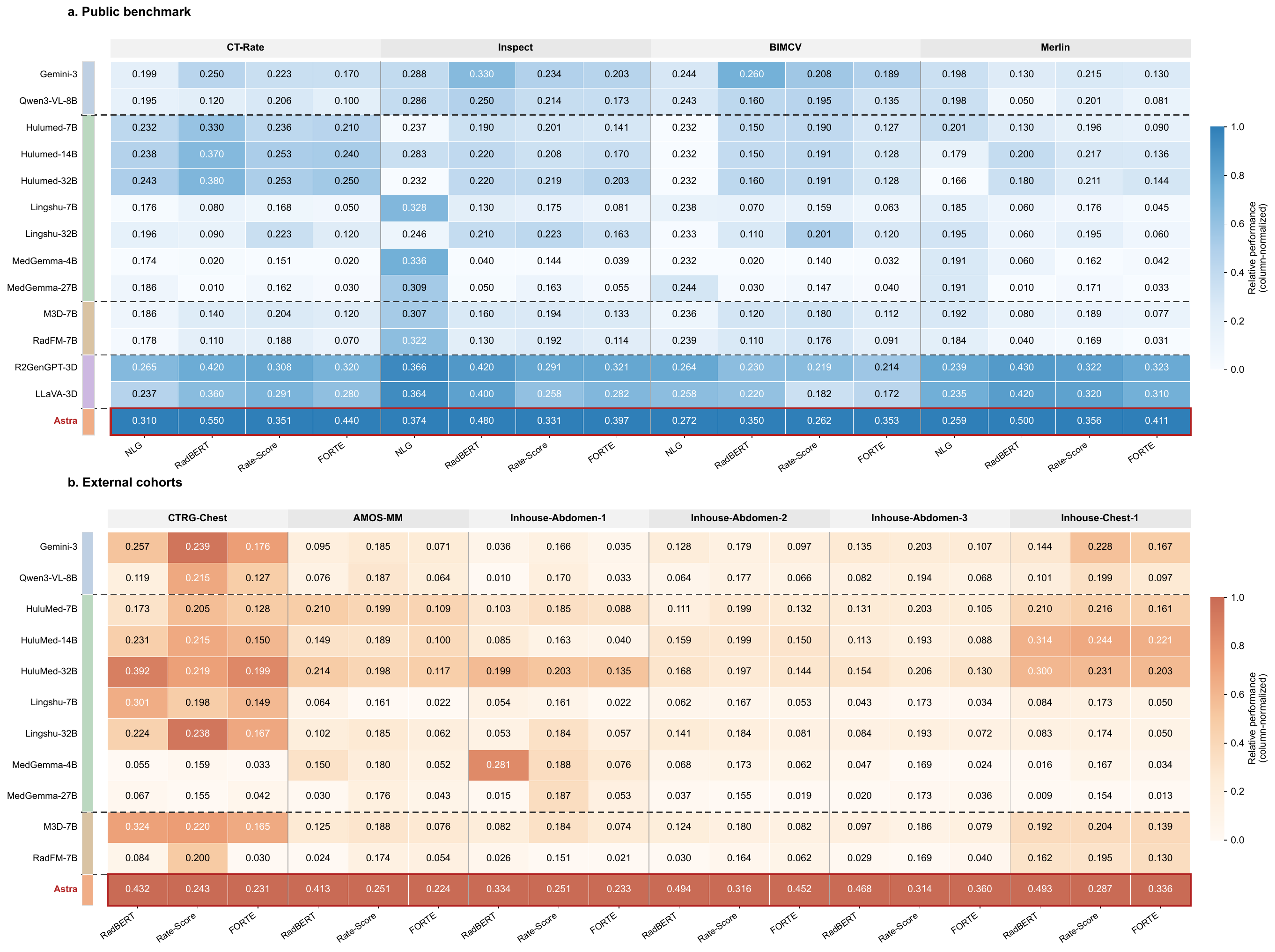}
		\caption{\textbf{Evaluation on public benchmark and external cohorts.} 
			\textbf{a}, Evaluation on CT-Rate ($n=1564$), Inspect ($n=1000$), BIMCV ($n=1505$) and Merlin ($n=1000$). 
			\textbf{b}, Evaluation on CTRG-Chest ($n=324$), AMOS-MM ($n=400$), Inhouse-Abdomen-1 ($n=400$), Inhouse-Abdomen-2 ($n=400$), Inhouse-Abdomen-3 ($n=400$), Inhouse-Chest ($n=400$). 
			NLG means average score across BLEU-4, ROUGE-L, METEOR. Astra outperforms current generalist models supporting volumetric 3D CT analysis, including general-purpose models (e.g., Gemini-3), medical generalist models (e.g., HuluMed) and radiology-oriented models (e.g., M3D). It also surpasses dataset-specific expert models fine-tuned on each benchmark with the same state-of-the-art visual encoder and large language model. Astra enables more fine-grained CT diagnostic report generation and maintains consistently strong performance on external cohorts.}
		\label{fig:evaluation}
	\end{figure}
	\subsection{Astra advances fine-grained CT diagnostic report generation}
	We evaluated Astra across chest and abdominal CT report generation datasets, including CT-Rate, Merlin, Inspect and BIMCV, with Merlin focused on abdominal CT and the others on chest imaging. Comparisons included all available MLLMs capable of 3D CT volumetric analysis, categorized as general-purpose MLLMs, represented by Gemini-3~\cite{team2023gemini} and Qwen3-VL~\cite{yang2025qwen3}; medical generalist MLLMs, including HuluMed~\cite{jiang2025hulu}, Lingshu~\cite{xu2025lingshu} and Med-Gemma~\cite{sellergren2025medgemma}; and radiology-specialized MLLMs, including RadFM~\cite{radfm} and M3D~\cite{bai2024m3d}. We also adapted R2GenGPT~\cite{wang2023r2gengpt} and LLaVA~\cite{llavarad} into dataset-specific 3D expert baselines, denoted R2GenGPT-3D and LLaVA-3D, using the same pretrained 3D visual encoder and large language model decoder as Astra. 
	For all 2D-based MLLMs, we used a standardized multi-slice input strategy for fair comparison.
	
	As shown in Fig.~3a, Astra consistently outperformed open-source, closed-source generalist models and dataset-specific expert models across both natural language generation (NLG) metrics and RadBERT-based metrics, which assess disease-level diagnostic accuracy. On CT-Rate, Astra achieved NLG and RadBERT scores of 0.250 and 0.550, exceeding R2GenGPT-3D (0.177 and 0.421) and HuluMed-32B (0.157 and 0.381). Similar improvements were observed on Inspect, where Astra scored 0.374 and 0.483 versus 0.365 and 0.417 for R2GenGPT-3D and 0.231 and 0.216 for HuluMed-32B, and on Merlin, where Astra reached 0.258 and 0.497 versus 0.238 and 0.432 for R2GenGPT-3D and 0.166 and 0.177 for HuluMed-32B. Across datasets, general-purpose and medical generalist MLLMs remained substantially inferior to dataset-specific expert models, indicating that high-fidelity clinical report generation remains under-optimized in current generalist systems. This limitation was especially evident for abdominal CT, likely reflecting its under-representation in existing medical generalist MLLMs. In contrast, Astra surpassed dataset-specific expert models while retaining broad generalizability, with particularly pronounced gains on abdominal CT.
	
	We next assessed fine-grained lesion description using the FORTE and Rate-Score metrics, which quantify whether reports accurately capture lesion degree, anatomical landmarks, imaging features, and overall impression. Astra achieved the best performance on all evaluated datasets, exceeding the second-best model on the FORTE metric by 37.5\% on CT-Rate, 23.7\% on Inspect, 65.7\% on BIMCV and 32.2\% on Merlin. These findings indicate that Astra improves not only overall report quality and diagnostic accuracy, but also the fidelity of fine-grained lesion characterization relevant to clinical decision-making.

	\subsection{Astra generalizes robustly across external datasets}
	To evaluate the external generalizability of Astra, we reserved AMOS-MM~\cite{amos} and CTRG-Chest~\cite{ctrg} as out-of-distribution (OOD) benchmarks and further assembled four independent real-world datasets from three clinical sites, comprising one chest CT cohort and three abdominal CT cohorts. These four real-world datasets were denoted as Inhouse-Abdomen-1, Inhouse-Abdomen-2, Inhouse-Abdomen-3, and Inhouse-Chest-1. Notably, CTRG-Chest was constructed from 2D screenshots, whereas all other datasets consisted of native volumetric 3D CT data. 
	These external cohorts differed markedly in voxel spacing, slice number, disease spectrum and report length, enabling a comprehensive assessment of Astra across heterogeneous imaging and reporting distributions. (supplementary Fig.~A8)

	We evaluated Astra on external 3D chest and abdominal CT datasets using RadBERT for disease-level diagnostic accuracy and Rate-Score and FORTE for fine-grained lesion description. Compared with general-purpose, medical generalist and radiology-specialized MLLMs, Astra achieved the best overall performance across all external datasets, with substantial gains over the second-best model across all three metrics (Fig.~3b). The advantage was particularly evident in the abdominal cohorts, where Astra achieved FORTE scores of 0.226 on AMOS-MM, 0.233 on Inhouse-Abdomen-1 and 0.455 on Inhouse-Abdomen-2, compared with 0.116, 0.134, and 0.150, respectively, for the strongest competing HuluMed variants. These findings demonstrate Astra’s robust generalization across public external benchmarks and independent real-world cohorts.

	Astra further demonstrated robustness to non-native 3D CT inputs. For evaluation on CTRG-Chest, we reassembled individual frames into pseudo-3D volumes. Relative to native 3D CT, these inputs lacked true inter-slice information and original Hounsfield unit values, preserving only fixed-sampling frames and pixel intensities mapped to 0–255. Despite this marked information loss, Astra achieved the best performance across all three metrics, with a RadBERT score of 0.432 versus 0.392 for HuluMed-32B, a Rate-Score of 0.243 versus 0.238 for Lingshu-32B, and a FORTE score of 0.231 versus 0.199 for HuluMed-32B. There result suggests that Astra is resilient to substantial degradation in volumetric fidelity and intensity standardization.
	
	\subsection{Astra improves clinical efficiency and diagnostic quality in human–AI collaboration}
	\begin{figure}[!htbp]
		\centering
		\includegraphics[width=0.85\textwidth]{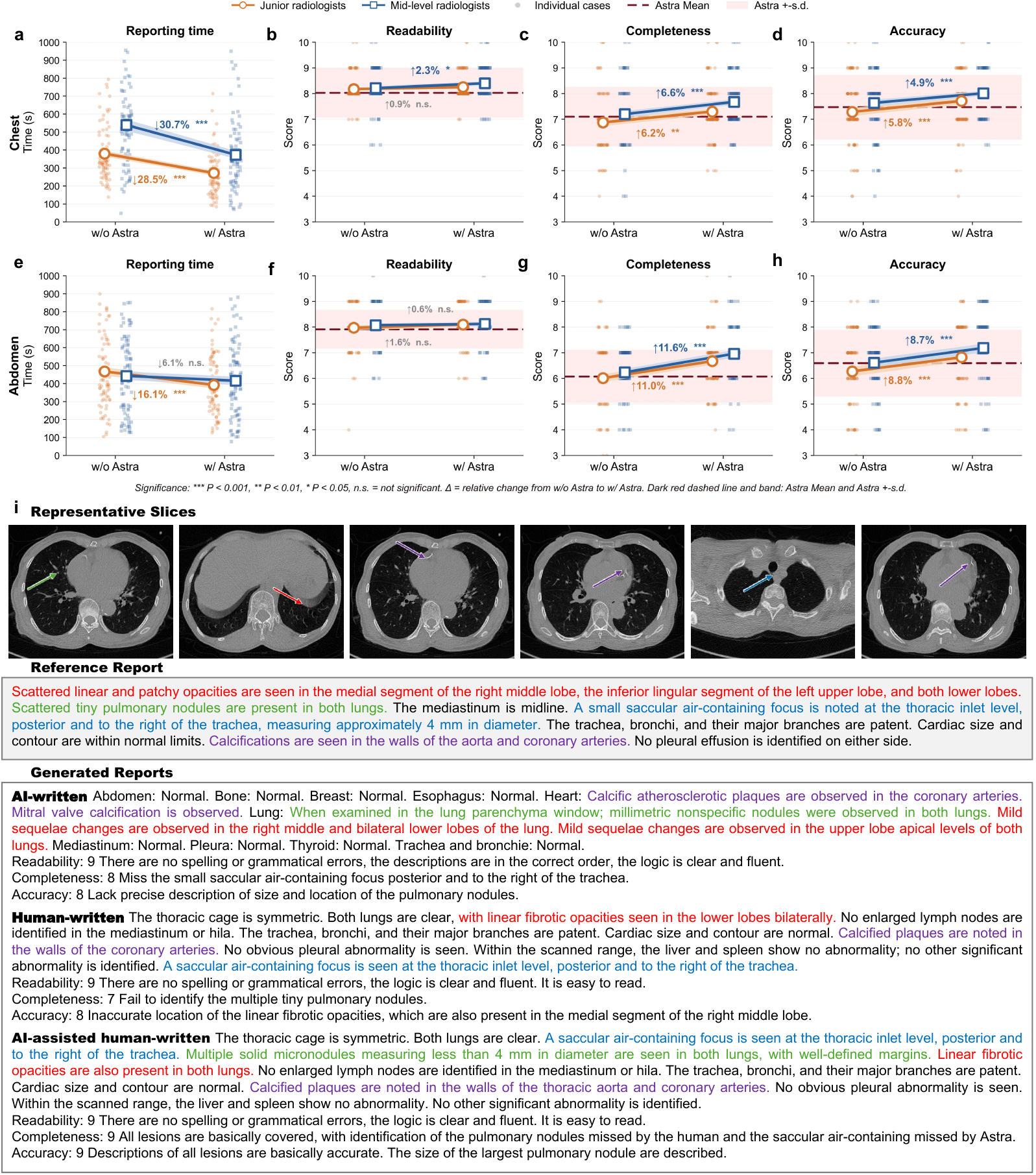}
		\captionof{figure}{\textbf{Human-AI collaboration in radiology report generation.} \textbf{a--d}, Comparison of reporting efficiency and report quality for chest CT interpretation with and without AI assistance. P values was calculated using a two-sided Wilcoxon signed-rank test. \textbf{e--h}, Comparison of reporting efficiency and report quality for abdomen CT interpretation with and without AI assistance. Orange bars indicate results from junior radiologists, blue bars indicate results from mid-level radiologists, and red dashed lines denote the mean performance of Astra. \textbf{i}, Case study of human–AI collaboration in report generation. Abnormality-specific textual descriptions and visual regions are highlighted in matched colours. Astra missed the tracheal diverticulum, whereas the radiologist missed a tiny pulmonary nodule. By contrast, the collaborative report covered nearly all abnormalities with detailed descriptions.}
		\label{fig:ensemble}
	\end{figure}
	After establishing Astra’s generalizability across diverse clinical cohorts, we next examined whether this performance translated into improved clinical reporting. We conducted a human--AI collaborative study at an external institution, providing a real-world evaluation beyond in-distribution benchmarks. Three junior radiologists, licensed within the past 5 years, and three mid-level radiologists, licensed 5--10 years previously, drafted chest and abdominal CT reports with and without Astra assistance. For each report, we recorded drafting time and then asked two senior radiologists with more than 10 years of post-licensure experience to score readability, completeness, and accuracy in a blinded manner using predefined criteria (supplementary Tab.~B5).

	Fig.~4 summarizes the human–AI collaboration results for chest CT. Astra improved reporting efficiency in both experience groups, reducing mean drafting time from 379.5 s to 271.5 s for junior radiologists and from 539 s to 373 s for mid-level radiologists, corresponding to speed gains of 28.5\% and 30.7\%, respectively. 
	Improvements in readability were modest, likely because Astra mainly provided positive abnormalities, whereas radiologists still finalized reports using their own templates. By contrast, Astra improved report completeness and accuracy in both groups. By rapidly localizing abnormal organs and anatomical regions, Astra allowed radiologists to focus on verification and refinement. Compared with unaided reporting, this collaborative workflow reduced missed abnormalities, as supported by the quantitative analysis in Fig.~4i and supplementary Fig.~A9.

	In abdominal CT, Astra showed a different pattern from chest CT. Reporting efficiency improved modestly, with mean drafting time decreasing from 467.0 s to 391.5 s in junior radiologists (16.1\%) and from 442.5 s to 415.0 s in mid-level radiologists (6.1\%). In contrast, report quality improved more substantially. Astra increased completeness and accuracy by 11.0\% and 8.8\% in junior radiologists, and by 11.6\% and 8.7\% in mid-level radiologists, respectively. 
	This pattern likely reflects the greater anatomical and pathological complexity of abdominal CT, which spans multiple organ systems and tissue compartments. 
	Subtle findings, such as small renal cysts, osseous lesions and abdominal aortic atherosclerosis, are more easily missed during unaided interpretation. By systematically highlighting abnormal organs and regions, Astra helped radiologists identify and incorporate these findings, improving completeness and accuracy. However, verifying and refining additional Astra-suggested findings required extra review time, attenuating the efficiency gains. This trade-off may explain the larger quality improvements but smaller efficiency gains observed for abdominal CT. Representative cases are shown in supplementary Fig.~A10,A11.

	Together, these results demonstrate Astra’s clinical value in report generation, improving efficiency in relatively simple, limited-organ interpretation tasks while enhancing diagnostic completeness in anatomically complex cases with a higher risk of missed findings. Rather than merely accelerating report drafting, Astra reconfigured the reporting workflow by allowing radiologists to shift part of their effort from initial abnormality search to targeted verification and refinement. The distinct benefit profiles observed in chest and abdominal CT further suggest that effective human--AI collaboration should be tailored to anatomical context, reader experience and error patterns. Future work should focus on disease categories prone to AI false positives or omissions, as well as on training radiologists to interpret and use AI-generated outputs critically and efficiently.

	\subsection{Astra facilitates efficient development of diagnostic models} 
	\begin{figure}[!htbp]
		\centering
		\includegraphics[width=0.8\textwidth]{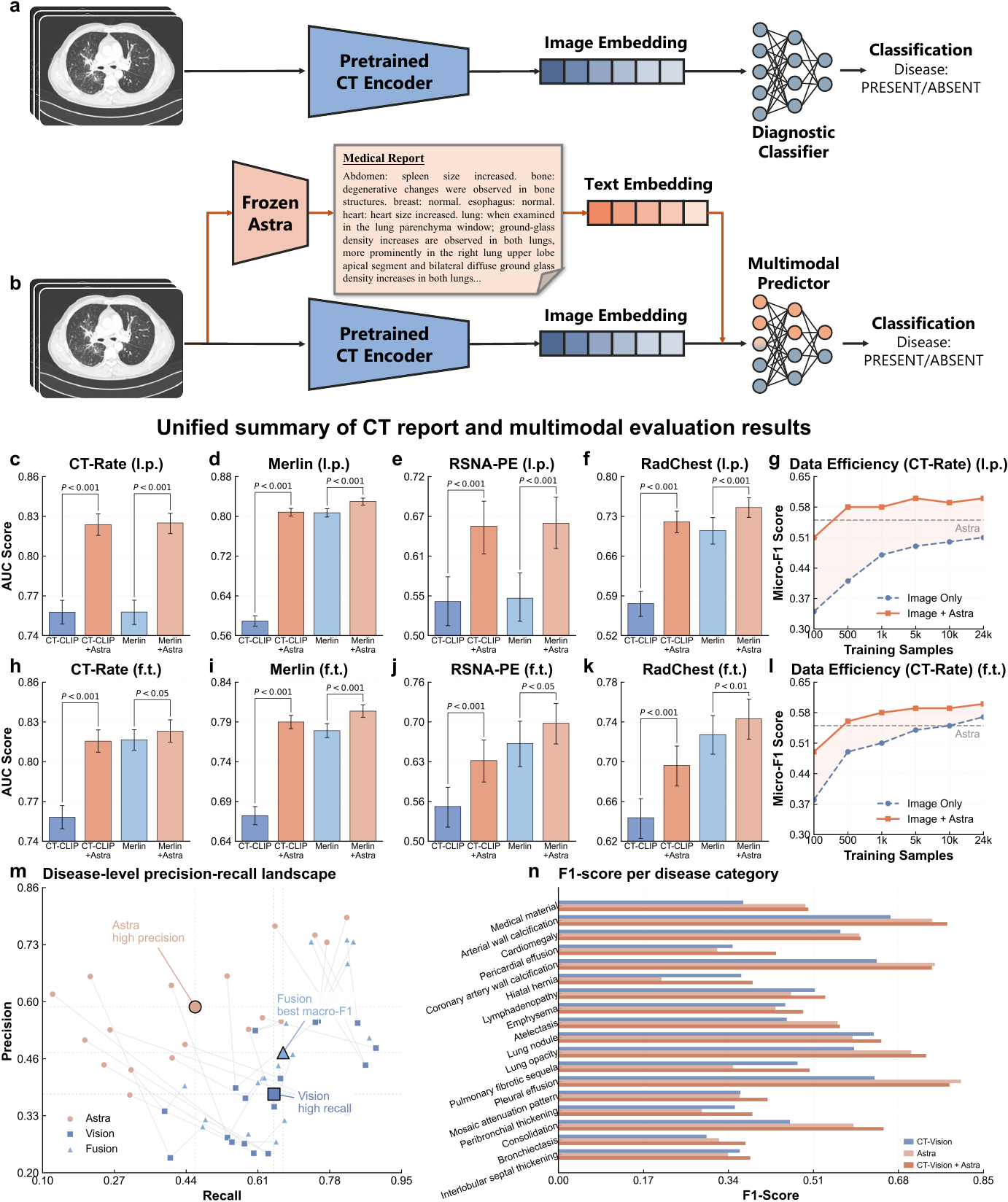}
		\captionof{figure}{\textbf{Astra facilitates efficient development of diagnostic models} \textbf{a}, Current disease diagnosis pipelines rely on fine-tuning pretrained vision foundation models. \textbf{b}, Astra boosts the performance of pretrained CT encoders through an ensemble strategy. \textbf{c--f}, Performance comparison between the ensemble strategy and the pretrained encoder alone under linear probing settings across four datasets (l.p. means linear probing, f.t. means full tune). Error bars represent 95\% confidence intervals. Statistical significance was assessed using 2,000 bootstrap resamples. \textbf{g}, Data efficiency analysis using Merlin encoder under linear probing on CT-Rate. \textbf{h--k}, Performance comparison and data efficiency analysis under full tune. \textbf{l}, Data efficiency analysis using Merlin encoder under full tune on CT-Rate. \textbf{m,n}, Precision-recall landscape and disease-specific F1-score breakdown across 18 categories, evaluated using CT-CLIP as the image encoder under full tune setting. }
		\label{fig:ensemble}
	\end{figure}
	We introduce a text-guided multimodal diagnostic architecture that uses Astra-generated reports to facilitate downstream multi-disease classification. Current disease classification pipelines mainly rely on fine-tuning pretrained vision foundation models (Fig.~5a). Motivated by the improvement observed in AI-assisted clinical interpretation (Fig.~4), we use Astra-generated diagnostic content as preliminary clinical priors for disease recognition. For each CT scan, Astra first generates a detailed diagnostic report, which is encoded into a high-dimensional semantic representation by a frozen text embedding model. In parallel, a vision foundation model extracts visual features from the raw scan. The linguistic and visual representations are then concatenated and passed to a multimodal classification head for downstream prediction. To preserve the specialized semantic information provided by Astra and avoid catastrophic forgetting or representational collapse, both Astra and the text embedding model remain frozen during training (Fig.~5b).

	We evaluated the proposed ensemble strategy on four thoracoabdominal CT classification datasets under linear probing and full tune. Across datasets and training settings, the strategy consistently improved performance over baseline (Fig.~5). These gains were model-agnostic. Integrating Astra-generated report priors improved the Merlin encoder from an AUC of 0.705 to 0.764 under linear probing and from 0.746 to 0.766 under full tune, while the CT-CLIP encoder improved from 0.618 to 0.751 and from 0.658 to 0.733, respectively. Gains were largest when baseline performance was lower. On the Merlin dataset under linear probing, the CT-CLIP baseline achieved an AUC of 0.589 (95\% CI, 0.578–0.600), which increased to 0.808 (95\% CI, 0.800–0.816) after integration with the ensemble strategy. 
	The ensemble strategy also improved data efficiency. In a representative CT-Rate experiment using the Merlin encoder, the ensemble model approached full-data baseline performance with substantially fewer labelled samples (Fig.~5g,l). Under full tune, 1,000 samples yielded an AUC of 0.809 (95\% CI, 0.801–0.817), close to the full-data baseline of 0.816 (95\% CI, 0.808–0.824). Under linear probing, 100 samples achieved an AUC of 0.766 (95\% CI, 0.759–0.771), comparable to the full-data baseline of 0.758 (95\% CI, 0.748–0.766).

	To investigate the mechanism underlying the ensemble strategy, we performed an error analysis across 18 disease categories on CT-Rate, using the CT-CLIP encoder with full tune. Astra-generated reports were mapped to disease labels using RadBERT, enabling comparison of class-wise precision and recall (Fig.~5m,n). This analysis revealed complementary error profiles. Astra showed a more conservative decision boundary, with higher precision but lower recall, whereas the CT-CLIP encoder showed higher recall but lower precision. The ensemble strategy reconciled these profiles, improving F1 scores by increasing precision while preserving the high recall of CT-CLIP.

	\subsection{Astra scales CT vision-language pretraining} 
	\begin{figure}[!htbp]
		\centering
		\includegraphics[width=0.85\textwidth]{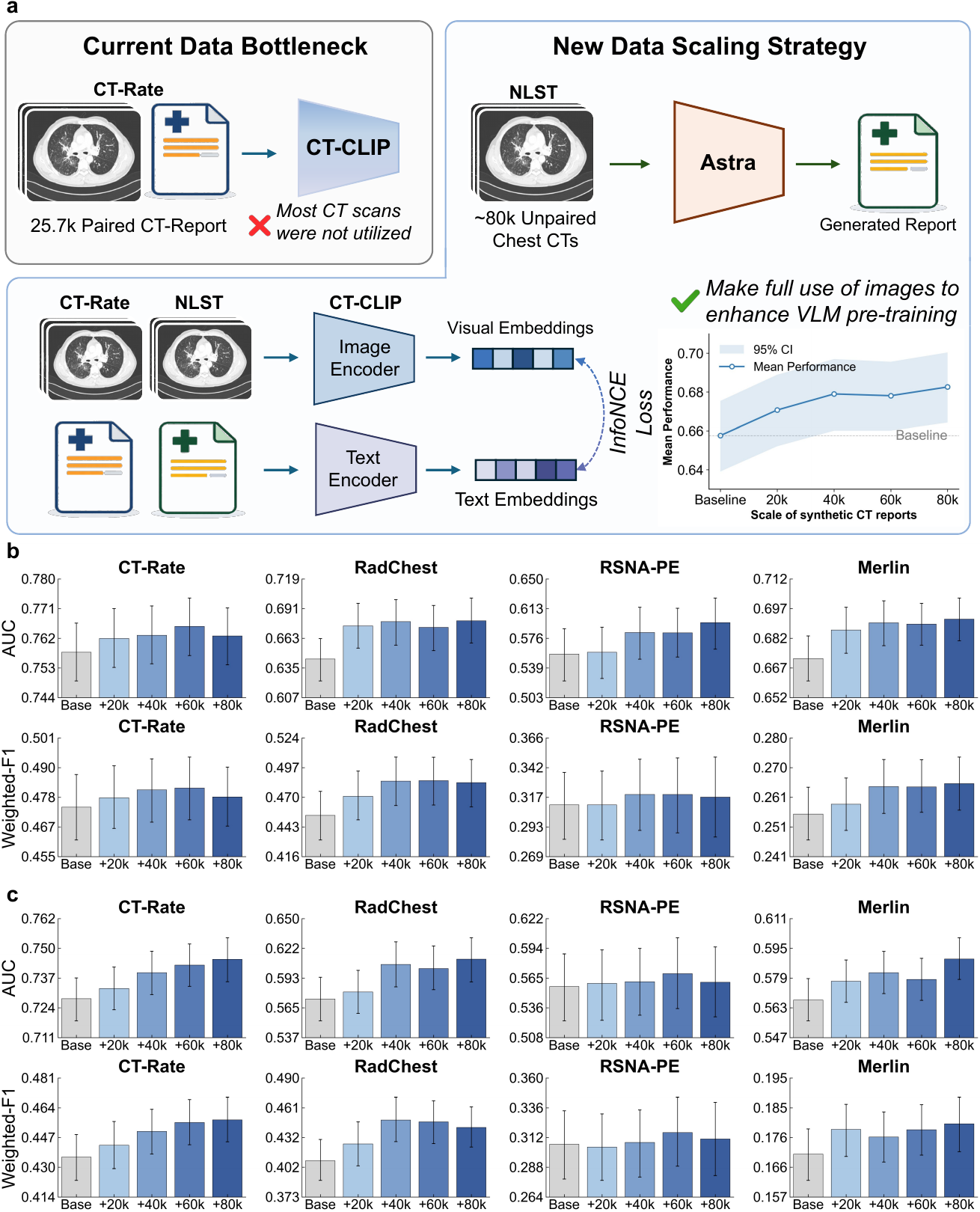}
		\captionof{figure}{\textbf{Astra scales CT vision-language pretraining} \textbf{a}, Scaling pretraining by integrating unlabeled NLST data (augmented with Astra-generated pseudo-reports) with the CT-Rate dataset yields consistent performance gains over pretraining on CT-Rate alone. \textbf{b,c}, Downstream classification performance of the visual encoder using 100\% (\textbf{b}) and 20\% (\textbf{c}) of labeled training samples. Astra-enhanced joint pretraining significantly boosts diagnostic accuracy and label efficiency across multiple benchmarks. Error bars represent 95\% confidence intervals.}
		\label{fig:scaling}
	\end{figure}
	The scarcity of paired CT--report datasets remains a major bottleneck for developing CT vision-language foundation models~\cite{ctrate}. We therefore tested whether Astra-generated reports could scale vision-language pretraining for CLIP-based models. Using CT-CLIP, originally pre-trained on CT-Rate, as the baseline, Astra generated comprehensive diagnostic reports for 84,839 unreported low-dose CT scans from NLST~\cite{nlst}. These synthetic image–report pairs were then progressively incorporated into the original CT-Rate pre-training corpus at varying mixing ratios. By repeating pretraining under these settings, we systematically evaluated how the volume of synthetic data influenced the representation quality of the resulting foundation model.
	
	For downstream classification, we evaluated model performance under full tune setting across three chest CT datasets, CT-Rate, RadChest~\cite{radchest} and RSNA-PE~\cite{colak2021rsna}, as well as the Merlin dataset, which represents a cross-anatomical abdominal application. We observed that as the volume of Astra-generated NLST reports increased, the mean AUC across these four benchmarks exhibited a consistent upward trend (Fig.~6a). Notably, the performance gains were more pronounced on out-of-distribution datasets compared to the in-domain CT-Rate dataset. Specifically, datasets such as Merlin and RadChest demonstrated the most substantial performance gains (Fig.~6b), suggesting that Astra-augmented pre-training significantly enhances the model's capacity for cross-dataset and cross-anatomical generalization.
	
	We next assessed the data efficiency of Astra-augmented pretraining model in downstream classification using only 20\% of the training samples for full tune (Fig.~6c). The resulting performance gains were broadly consistent with those observed under full-data training. Notably, on the in-domain CT-Rate dataset, the benefit of Astra-augmented pre-training was modest when the full training set was available but became more pronounced in the low-data setting. Gains were smaller on more challenging tasks, such as RSNA-PE, likely owing to the greater complexity and more specialized diagnostic criteria required for pulmonary embolism detection; nevertheless, the Astra-augmented model nevertheless remained consistently superior to the baseline. Together, these results indicate that the benefits of synthetic report scaling extend to data-limited downstream adaptation.
	
	\subsection{Reinforcement Learning enhances fine-grained caption ability} 
	\begin{figure}[!htbp]
		\centering
		\includegraphics[width=0.85\textwidth]{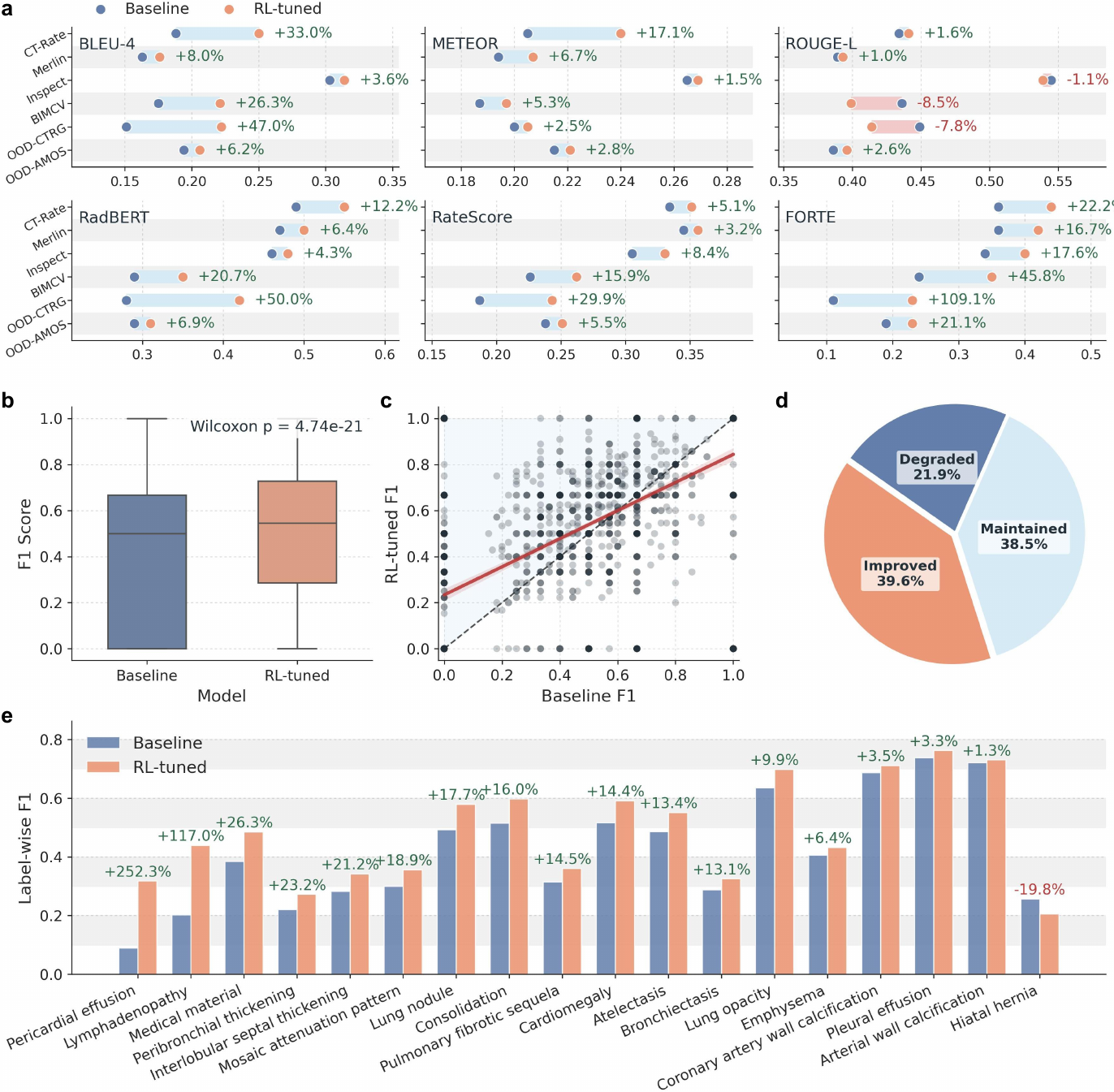}
		\captionof{figure}{\textbf{Ablation studies of harmonization strategy and reinforcement learning.} \textbf{a}, Performance gains from the reinforcement learning across multiple benchmarks. Baseline means the supervised fine-tuning model trained on CTRgDB, whereas RL-tuned means the model further optimized with reinforcement learning. Additional results are provided in supplementary Tab.~B3,4. \textbf{b--d}, Case-level F1-score analysis on CT-Rate. The box plot (\textbf{b}) shows median and quartiles, with P values calculated using a two-sided Wilcoxon signed-rank test. The scatter plot (\textbf{c}) with regression fit (red line) and pie chart (\textbf{d}) illustrate diagnostic shifts and the proportion of improved cases after RL-tuned. \textbf{e}, Label-wise F1-score breakdown for 18 disease categories, highlighting substantial improvements in complex findings.}
		\label{fig:rl_ablation}
	\end{figure}
	We subsequently investigated the contribution of reinforcement learning. Across CTRgDB and six out-of-distribution datasets, GRPO consistently improved clinically oriented report generation metrics, with larger gains under distribution shift (Fig.~7a, supplementary Tab.~B1--B4). Mean relative improvements in RadBERT micro-F1, Rate-Score and FORTE were 10.92\%, 7.70\% and 24.68\% in the in-distribution setting respectively, and increased to 32.29\%, 13.37\% and 45.13\% in the out-of-distribution setting, respectively. These results suggest that supervised fine-tuning captures the linguistic structure of clinical reports but does not fully optimize diagnostic fidelity, whereas GRPO directly rewards clinically relevant report quality, thereby improving both disease-level accuracy and fine-grained abnormality characterization. Improvements on Atlas3.0 were limited, likely owing to its synthetically generated reports and restricted focus on liver, pancreatic and renal lesions. The limited diagnostic diversity and clinical richness of this corpus may have produced sparse reward signals during reinforcement learning, hindering policy convergence. This observation highlights the dependence of reinforcement learning on high-quality and diverse training narratives, suggesting that the proposed RL framework is most effective when applied to clinically rich and complex report data.

	To further investigate the mechanism of reinforcement learning, we performed a fine-grained case-level analysis (Fig.~7c--f). RL shifted diagnostic performance toward higher reliability: 39.6\% of cases showed substantial accuracy gains, whereas 21.9\% declined, consistent with an overall precision improvement. The gains were most evident in challenging cases with near-zero baseline F1 scores. Label-wise analysis also showed marked improvements for subtle pathologies, including pericardial effusion and lymphadenopathy. These findings suggest that RL preferentially improves difficult cases and reduces long-tail errors that are less effectively addressed by supervised training.

	The performance of the reinforcement learning framework was strongly influenced by reward design. We compared four reward functions: FORTE, RateScore, RadBERT and a composite natural language generation (NLG) metric averaging BLEU-4~\cite{papineni2002bleu}, ROUGE-L~\cite{lin2004rouge} and METEOR~\cite{banerjee2005meteor} (supplementary Tab.~B6). FORTE provided the best overall trade-off between clinical efficiency and linguistic fidelity. Two patterns emerged. First, rule-based rewards were more stable than model-based alternatives, the substantial decline in NLG scores under Rate-Score suggested reward hacking. Second, dense rewards were more effective than sparse diagnostic rewards. Unlike RadBERT, which primarily rewards disease-level accuracy, FORTE provides fine-grained caption-level supervision, improving diagnostic performance while preserving descriptive quality.
	
	\section{Discussion}
	
	In this study, we present Astra, a highly generalizable foundation model engineered for site-agnostic thoracoabdominal CT report generation. Astra’s robust capabilities are underpinned by training on a massive multi-center cohort of 90,678 volumetric CT–report pairs. Rigorous systematic evaluation, particularly on out-of-distribution data, across report generation quality, diagnostic accuracy, and fine-grained lesion captioning demonstrates that Astra consistently outperforms existing MLLMs, whether general-purpose, medical-domain, or radiology-specialized. Crucially, whereas current 3D-capable MLLMs lag behind dataset-specific expert models in radiology reporting, Astra bridges this developmental gap as a unified architecture that transcends both. By delivering high-quality reports across arbitrary CT images, Astra establishes a plug-and-play foundation for clinical deployment, streamlining diagnostic workflows and unlocking the immense potential of large-scale CT research.

	Astra represents an early effort to develop a generalizable, multi-region and multi-center foundation model for CT report generation. Previous 3D CT report generation studies~\cite{braingpt,ctagrg,reg2rg,deng2025mvketr,hamamci2024ct2rep,diallama} have largely followed the single-center paradigm established in chest X-ray research~\cite{bannur2024maira,hyland2023maira,r2gen,r2gencmn,wang2023r2gengpt}, with models typically trained and evaluated on isolated benchmarks. A major barrier is the substantial variation in reporting style~\cite{delbrouck2025automated}, conventions and descriptive granularity across institutions, which makes naive multi-center training difficult and can impair generalization. Astra addresses this challenge by combining the largest harmonized multi-center CT report generation dataset to date with reinforcement-learning-based post-training. Region-wise and abnormality-focused preprocessing harmonizes heterogeneous reports into a unified and comprehensive template, while reinforcement learning encourages the model to prioritize positive imaging findings rather than merely imitate physician writing style. Comprehensive evaluations across natural language generation quality, disease-level diagnostic accuracy and fine-grained captioning capability consistently favored Astra. These results move CT report generation beyond the traditional single-center paradigm and support broader clinical translation of generalizable 3D CT reporting models.

	The fine-grained identification and precise modeling of diagnostic information form the basis of Astra. Ablation studies showed that report-style unification enriched diagnostically informative supervision and improved diagnostic performance. This suggests that raw CT reports may not provide the optimal supervision target for report generation, and that careful report processing warrants greater attention in future work. We further showed that Group Relative Policy Optimization substantially improved CT report generation, enhancing both disease-level diagnosis and fine-grained captioning. GRPO also reduced completely erroneous predictions and improved performance on difficult and external cases, which is particularly relevant for clinical translation. These findings indicate that supervised fine-tuning mainly teaches models to reproduce radiologist-like report text, whereas reinforcement learning encourages more accurate diagnostic reasoning and more precise abnormality characterization. These results support reinforcement learning based post-training as a promising strategy for improving the clinical reliability of future CT report generation systems.

	Astra facilitates a seamless integration into clinical workflows, fostering a human-AI collaboration that enhances both the efficiency and quality of report generation for junior and mid-level radiologists. As report generation models move toward clinical translation, collaborative workflows are increasingly favoured over fully autonomous systems~\cite{medmpt,tanno2025collaboration,octrg,dong2025keyword}. However, most previous studies have evaluated such workflows on internal datasets, leaving their utility on external cohorts insufficiently explored. For generalizable CT report generation models, external validation is essential for assessing clinical applicability. In human--AI collaborative testing on Inhouse-Abdomen-3 and Inhouse-Chest-1, Astra primarily improved reporting efficiency for chest CT, whereas its quality benefits were more pronounced for abdominal CT. These findings suggest that Astra can streamline reporting in anatomically simpler and lower-complexity cases, while in multi-organ abdominal cases it helps reduce omitted findings through targeted human verification. Future work should better define the diagnostic strengths and failure modes of AI models and provide radiologists with training in critical and efficient use of AI-generated outputs.

	Beyond report generation, Astra enhances the diagnostic performance of pretrained visual foundation models. Although these models can accelerate downstream medical AI development, they remain constrained by substantial annotation requirements and residual accuracy gaps. We explored a report-guided ensemble strategy where Astra first generates a detailed preliminary report, which is then used as structured semantic guidance for disease classification by vision-pretrained models. Across in-distribution and out-of-distribution datasets, anatomical regions and fine-tuning data scales, this strategy consistently improved classification performance, highlighting the value of language-rich clinical priors for refining visual decision boundaries. It also improved data efficiency, with comparable performance achieved using less than 10\% of the original fine-tuning data. 
	
	Astra unlocks the latent value of unannotated CT archives by autonomously generating high-fidelity diagnostic reports. CLIP-style foundation models rely on large-scale paired image-text pretraining, yet curating such datasets in medicine is constrained by prohibitive costs, manual effort, and stringent privacy regulations. As a result, large public repositories of unpaired CT scans remain underused. Although generative data scaling has been explored in medical imaging, most efforts focus on pixel-level tasks, such as missing-modality synthesis~\cite{koetzier2024generating} or virtual lesion generation~\cite{ma2026generative}. In contrast, Astra extends this concept to volumetric report generation. Leveraging its cross-center generalizability, we generated reports for 84,839 unannotated low-dose chest CT scans from NLST and used the resulting image-report pairs to augment CT-CLIP pretraining. Progressive integration of these synthetic pairs into CT-Rate consistently improved downstream classification performance over pretraining with real CT-Rate reports alone, with particularly clear gains on external validation cohorts and cross-anatomy datasets. 
	These findings suggest that Astra provides a scalable route for converting unannotated CT archives into vision-language training resources, thereby alleviating the annotation bottleneck for next-generation medical foundation models.
	
	Our study also has limitations.  
	First, owing to the limited availability of patient history, treatment history and physiological information, the current model is largely limited to identifying abnormalities that are directly visible on imaging. Integrating richer clinical metadata may enable more precise diagnosis and, eventually, more actionable treatment-related recommendations. 
	Second, Astra does not yet provide explicit lesion localization or segmentation, instead describing lesion location only in text. Addressing this limitation will probably require specialized decoder designs, and future work may explore unified models for joint segmentation and report generation or multi-agent systems that coordinate these complementary tasks.
	Third, we mainly validated Astra as a tool for scaling vision–language pretraining using the NLST and CT-Rate datasets. Future work could expand this direction by aggregating a broader collection of publicly available paired CT–report data together with large-scale report-free CT scans to train stronger pretrained models. 
	
	In conclusion, Astra represents a paradigm shift in 3D CT report generation, serving as a versatile foundation model that transcends the inherent constraints of single-region and single-cohort systems. Our extensive evaluation across multiple dimensions, spanning both in-distribution benchmarks and zero-shot external hospital datasets, confirms its reliability in real-world clinical environments. Crucially, through human-AI collaborative trials conducted within active hospital workflows, we validated that Astra effectively assists radiologists by enhancing both report-drafting efficiency and diagnostic completeness. Furthermore, Astra’s exceptional foundational extensibility is epitomized by its ability to synergistically integrate with pretrained vision models to boost downstream diagnostic performance, while simultaneously unlocking the latent value of massive unannotated archives by providing pseudo-labels through automated report generation. By bridging these capabilities, Astra positions itself as an effective clinical assistant and a pivotal infrastructure for the next generation of multimodal CT artificial intelligence.
	
	\section{Methods}
	
	\subsection{Datasets collection}
	
	To develop Astra, we collected, to the best of our knowledge, all publicly available thoracoabdominal CT--report paired datasets, including CT-Rate, Inspect, BIMCV, Atlas3.0, Merlin, CTRG-Chest and AMOS-MM. CT-Rate, Inspect, BIMCV, Atlas3.0 and Merlin were systematically harmonized to form CTRgDB, which served as the training and internal evaluation resource for Astra. CTRG-Chest and AMOS-MM were held out exclusively for external validation. Detailed train, validation and test splits are provided in supplementary Tab.~B7. Because Inspect does not contain a findings section, its impression section was used; for all other datasets, the findings section was used. Thoracic CT data mainly comprised non-contrast chest CT and CT pulmonary angiography, whereas abdominal CT data were predominantly contrast-enhanced CT.

	For evaluations involving Gemini-3, large-scale inference was limited by computational and API costs. We therefore used a fixed random subset of 1{,}000 cases from the Merlin and Inspect test sets. This sample size was comparable to other benchmark test sets, including CT-Rate ($n = 1{,}564$) and Atlas3.0 ($n = 883$). All other ablation experiments were performed on the full corresponding test sets. To confirm that subsampling did not materially affect Astra evaluation, we compared its performance on the full Merlin test set and the selected 1{,}000-case subset. Performance was highly consistent between the full set and subset. FORTE scores were 0.4021 (95\% CI, 0.3983--0.4058) versus 0.4076 (95\% CI, 0.3992--0.4157), RaTE-Scores were 0.3544 (95\% CI, 0.3520--0.3569) versus 0.3564 (95\% CI, 0.3509--0.3618), and RadBERT scores were 0.4936 (95\% CI, 0.4863--0.5008) versus 0.4973 (95\% CI, 0.4810--0.5137), respectively.
	
	For real-world external evaluation, we further collected one thoracic CT cohort and three abdominal CT cohorts from three external institutions. Guizhou Provincial People’s Hospital provided 400 non-contrast chest CT scans and 400 contrast-enhanced abdominal CT scans. The First Affiliated Hospital, Zhejiang University School of Medicine, provided 400 contrast-enhanced abdominal CT scans. Shanghai Sixth People's Hospital Affiliated to Shanghai Jiao Tong University School of Medicine, also provided 400 contrast-enhanced abdominal CT scans. For these external datasets, the findings sections were used, and all reports were translated into English using the DeepSeek API. Due to institutional data privacy requirements, all evaluations on the in-house datasets were performed using locally deployed models within the respective hospital infrastructures.
	
	We evaluated the ensemble strategy between Astra and pretrained models on four CT classification datasets. Specifically, CT-Rate and RadChest were used to assess classification performance on common thoracic abnormalities. CT-Rate covers 18 thoracic abnormality categories, whereas for RadChest we followed the CT-CLIP setting and evaluated the 16 major thoracic abnormalities. We used the Merlin dataset to evaluate classification performance on common abdominal abnormalities, covering 30 categories. We further used the RSNA-PE dataset to assess pulmonary embolism detection, including left-sided, right-sided, central and chronic pulmonary embolism. For Merlin, CT-Rate and RadChest, we adopted the official dataset splits. For RSNA-PE, as the dataset was originally released for a challenge setting and its test set was not publicly available, we partitioned the original training set into training and test subsets using an 8:2 ratio. The original 2D DICOM series were reconstructed into 3D volumes, resulting in 7,279 cases used for training and evaluation.
	
	To investigate whether Astra can facilitate the scaling of CT vision–language pretraining, we used CT-Rate as the baseline paired image–report dataset and NLST as a large-scale report-free dataset. NLST contains 84,839 low-dose chest CT scans, which were evenly divided into four subsets and progressively incorporated into the 24,128 training cases of CT-Rate for CLIP-style pretraining. The resulting pretrained models were evaluated on four downstream CT classification datasets. Thoracic abnormality recognition was assessed on CT-Rate, external generalization was evaluated on RadChest and RSNA-PE, and cross-anatomy generalization was assessed on Merlin.
	
	\subsection{Astra model development}
	\subsubsection{Report standardization}
	Given the substantial variation across institutions in reporting style, diagnostic ordering and report templates, standardization of radiology reports was necessary for the development of a multi-center foundation model for CT report generation. We therefore first designed unified reporting templates for both thoracic and abdominal CT reports. For thoracic CT, we adopted a template inspired by Reg2RG, in which findings were organized sequentially into ten anatomical regions: abdomen, bones, breasts, esophagus, heart, lungs, mediastinum, pleura, thyroid, and trachea and bronchi. For abdominal CT, we used a template based on the anatomical partitioning strategy of Merlin, with findings sequentially organized into thirteen regions: lower thorax, liver and biliary tree, gallbladder, spleen, pancreas, adrenal glands, kidneys and ureters, gastrointestinal tract, peritoneum, pelvis, vasculature, lymph nodes, and musculoskeletal system.

	Large language models have shown strong performance in medical text processing~\cite{kirchler2026large_llmprocess1,sandmann2025benchmark_llmprocess2,tordjman2025comparative_llmprocess3,hu2024zero_llmprocess4,lopez2025clinical_llmprocess5}. We therefore used DeepSeek to standardize the original radiology reports. Detailed prompts are provided in supplementary Tab.~B8,9, with representative cases shown in supplementary Tab.~B10,11. Non-diagnostic text, including clinician-to-clinician communications, summary codes and procedural notes, was first removed because it is not directly informative for image-based report generation and may hinder training. We then extracted diagnostically relevant statements for each predefined anatomical region, removed negative mentions, rewrote comparative statements into self-contained descriptions and preserved positive findings. This procedure yielded a unified multi-center dataset of diagnostic reports. For CT-Rate, the region-splitting step was omitted because region-level annotations were already available from RadGenome-ChestCT~\cite{radgenome}.
	
	\subsubsection{Model architecture}
	The overall architecture of Astra is shown in Fig.~8. Merlin, a 3D CT pretrained model, was used as the visual encoder. We extracted its final convolutional feature map, with spatial dimensions of $7 \times 7 \times 10$ and 2,048 channels, and reshaped it into 490 visual tokens of dimension 2,048. These tokens were then compressed using a six-layer Perceiver with 32 learnable latent tokens, yielding 32 aggregated visual tokens of dimension 2,048. The compressed visual tokens were projected to the hidden dimension of the large language model through a linear layer. As the language backbone, we used the language component of Qwen2.5-VL, while removing its original 2D visual encoder. For instruction tuning, we designed three prompt templates. The abdomen template and chest template were used for abdominal and thoracic CT report generation, respectively, following the corresponding standardized report templates. We further introduced a focus template for the Atlas3.0 dataset, which contains annotations for only three anatomical regions: liver and biliary tree, pancreas, and kidneys and ureters. The detailed instruction templates are provided in supplementary Tab.~B12.
	
	\subsubsection{Training strategy}
	\begin{figure}[!htbp]
		\centering
		\includegraphics[width=0.99\textwidth]{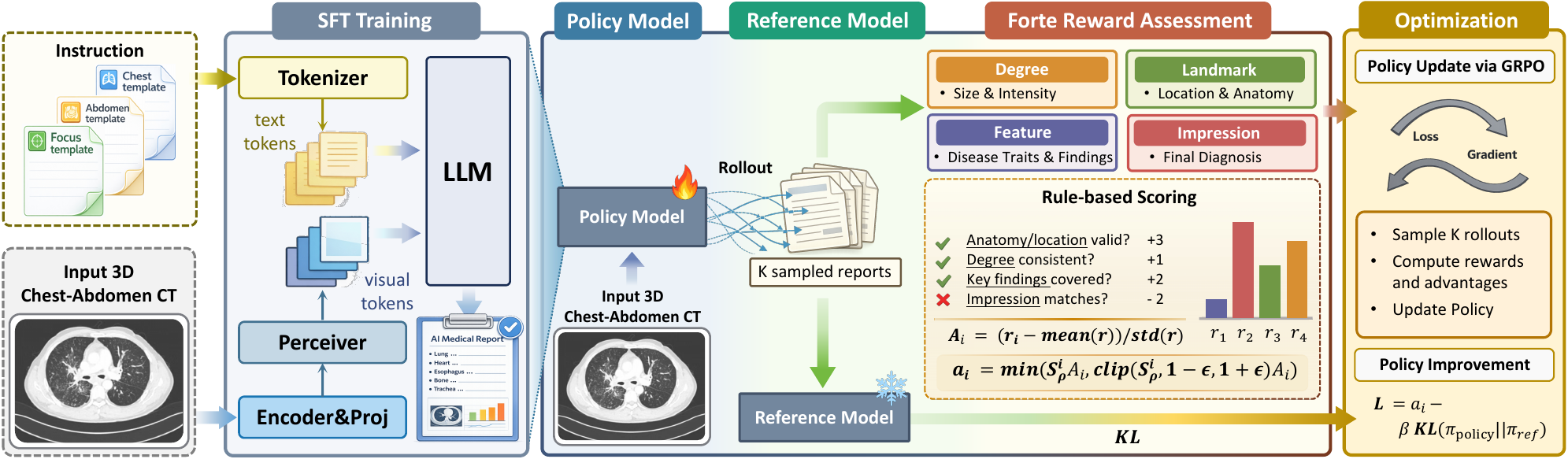}
		\captionof{figure}{\textbf{Architecture and training strategy of Astra.} Astra adopts a LLaVA-style architecture composed of a 3D visual encoder, a Perceiver token compressor and a large language model decoder. Volumetric CT inputs are first encoded into visual features, compressed into a small set of latent tokens, and then projected into the language model for report generation. Training is performed in two stages: supervised fine-tuning to establish 3D CT understanding and report generation ability, followed by GRPO with FORTE-based reward optimization to increase the likelihood of fine-grained diagnostic descriptions and improve detailed captioning performance.}
		\label{fig:rl_ablation}
	\end{figure}
	Astra was trained in two stages, and the overall training pipeline is illustrated in Fig.~8. In the first stage, we performed supervised fine-tuning using the 73,127 training samples constructed from the standardized multi-center dataset. Training was conducted with an autoregressive language modeling objective, with the maximum report length truncated to 1,000 tokens. 
	In the second stage, we further optimized Astra using Group Relative Policy Optimization (GRPO) to improve its global diagnostic ability and fine-grained captioning performance. We initialized both a policy model and a reference model from the supervised fine-tuned (SFT) checkpoint. The policy model was fully trainable, whereas the reference model was kept frozen throughout reinforcement learning.
	
	For each training step, we sampled an input $v \sim P(V)$ from the training distribution, where $v$ consists of an image $I$ and a question $q$. Given $v$, we sampled a group of $G$ candidate completions $\{o_i\}_{i=1}^{G}$ from the old policy model,
	\begin{equation}
		\{o_i\}_{i=1}^{G} \sim \pi_{\theta_{\mathrm{old}}}(\cdot \mid v).
	\end{equation}
	
	Each completion $o_i$ was assigned a scalar reward $r_i$ by the reward function,
	\begin{equation}
		r_i = R(v, o_i, gt).
	\end{equation}
	
	We then computed a standard group-relative advantage for each sampled completion:
	\begin{equation}
		A_i = \frac{r_i - \mathrm{mean}\!\left(\{r_j\}_{j=1}^{G}\right)}
		{\mathrm{std}\!\left(\{r_j\}_{j=1}^{G}\right)}.
	\end{equation}
	This removes the need for a learned value function while preserving a stable relative learning signal within the group.
	
	Following GRPO, the policy $\pi_{\theta}$ was updated using a PPO-style clipped objective with an additional KL regularization term to a fixed reference policy $\pi_{\mathrm{ref}}$:
	\begin{equation}
		J_{\mathrm{GRPO}}(\theta)
		=
		\mathbb{E}_{v, \{o_i\}}
		\left[
		\frac{1}{G}\sum_{i=1}^{G}
		\min\left(
		\rho_i(\theta) A_i,\,
		\mathrm{clip}\!\left(\rho_i(\theta), 1-\epsilon, 1+\epsilon\right) A_i
		\right)
		-\beta\, D_{\mathrm{KL}}\!\left(\pi_{\theta} \,\|\, \pi_{\mathrm{ref}}\right)
		\right],
		\label{eq:grpo}
	\end{equation}
	where the importance ratio is defined as
	\begin{equation}
		\rho_i(\theta)
		=
		\frac{\pi_{\theta}(o_i \mid v)}
		{\pi_{\theta_{\mathrm{old}}}(o_i \mid v)},
	\end{equation}
	and $\epsilon \geq 0$ and $\beta \geq 0$ control the clipping range and regularization strength, respectively.
	
	Reward design is critical for GRPO. Inspired by the FORTE metric, we aimed to construct a rule-based reward that evaluates generated reports from four clinically relevant dimensions: Degree (size and intensity), Landmark (location and anatomy), Feature (disease traits and primary findings), and Impression (final diagnosis). This design is attractive for CT report generation because it can simultaneously reward both global disease diagnosis and fine-grained abnormality description.
	
	However, the original FORTE metric has two limitations. First, its keyword set is relatively limited and does not adequately cover the broad spectrum of thoracic and abdominal abnormalities encountered in multi-center datasets. Second, it relies heavily on sentence-pair matching between the reference and generated reports before computing the FORTE score. In radiology report generation, this can lead to unreasonable alignments when a finding present in the reference is absent in the prediction, while another unmatched finding appears only in the prediction. For example, a liver cyst in the reference may be incorrectly paired with a kidney cyst in the generated report, which introduces substantial noise into reward estimation.
	
	To address these two limitations, we first updated the FORTE keyword inventory and synonym groups. To cover as broad a spectrum of abnormality types and reporting expressions as possible, we began by extracting candidate keywords for the four FORTE dimensions from the 90,678 reports collected in this study. This step was performed with the assistance of DeepSeek using the prompt shown in supplementary Tab.~B13, with the goal of capturing diverse abnormality descriptions and linguistic variations used in real-world radiology reports. We then worked with two clinicians to curate the extracted terms, using ICD-10 as a reference to filter irrelevant entries and merge synonymous expressions. This process resulted in the final FORTE keyword and synonym lexicon used in our study. As shown in supplementary Tab.~B6, GRPO post-training with the updated lexicon substantially improved model performance across NLG, RadBERT, FORTE and Rate-Score metrics, whereas the original keyword set failed to produce comparable gains. This finding further suggests that the revised lexicon provides more comprehensive and accurate coverage of thoracoabdominal CT abnormalities. To address the dependence of the original FORTE metric on sentence-pair matching, we leveraged the region-wise structure of our standardized reports and directly extracted the diagnostic findings for each anatomical region from both the reference and generated reports. This design substantially reduced mismatched alignment of abnormalities across organs and mitigated reward hacking. For a report containing $G$ anatomical regions, we first computed the four FORTE component scores for each region $g$:
	\begin{equation}
		r_{\mathrm{degree}}^{(g)}, \quad
		r_{\mathrm{landmark}}^{(g)}, \quad
		r_{\mathrm{feature}}^{(g)}, \quad
		r_{\mathrm{impression}}^{(g)}.
	\end{equation}
	
	The regional FORTE score was then defined as the average of the four component scores:
	\begin{equation}
		r_{\mathrm{FORTE}}^{(g)} =
		\frac{
			r_{\mathrm{degree}}^{(g)} +
			r_{\mathrm{landmark}}^{(g)} +
			r_{\mathrm{feature}}^{(g)} +
			r_{\mathrm{impression}}^{(g)}
		}{4}.
	\end{equation}
	
	The overall FORTE reward for the report was obtained by averaging across all regions:
	\begin{equation}
		r_{\mathrm{forte}} =
		\frac{1}{G}\sum_{g=1}^{G} r_{\mathrm{FORTE}}^{(g)}.
	\end{equation}
	
	To further reduce the repetition problem of the Qwen-based language model, we introduced a repetition penalty. The final reward was defined as
	\begin{equation}
		r =
		\begin{cases}
			r_{\mathrm{forte}}, & \text{if no repetition is detected},\\
			0, & \text{otherwise}.
		\end{cases}
	\end{equation}

	\subsubsection{Evaluation protocol}
	We evaluated report generation performance from three complementary perspectives: natural language generation (NLG) quality, global disease diagnosis, and fine-grained captioning. For NLG evaluation, we used the conventional captioning metrics BLEU, METEOR, and ROUGE-L. BLEU-$n$ measures the precision of matched $n$-grams between the generated report and the reference report:
	\begin{equation}
		\mathrm{BLEU}\text{-}n
		=
		\exp\left(
		\sum_{i=1}^{n}\frac{1}{n}\log p_i
		\right)\times \mathrm{BP},
	\end{equation}
	where $p_i$ denotes the precision of matched $i$-grams and $\mathrm{BP}$ is the brevity penalty.
	
	METEOR evaluates report quality by considering exact word matches as well as stem, synonym and word-order correspondences:
	\begin{equation}
		\mathrm{METEOR}
		=
		F_{\mathrm{mean}} \times (1-\mathrm{penalty}),
	\end{equation}
	where $F_{\mathrm{mean}}$ is the harmonic mean of precision and recall, and $\mathrm{penalty}$ reflects the fragmentation of matched unigrams.
	
	ROUGE-L measures similarity based on the longest common subsequence between the generated and reference reports:
	\begin{equation}
		\mathrm{ROUGE}\text{-}\mathrm{L}
		=
		\frac{(1+\beta^2)\,R\,P}{R+\beta^2 P},
	\end{equation}
	where $R_{\mathrm{LCS}}$ and $P_{\mathrm{LCS}}$ denote the recall and precision of the longest common subsequence, respectively, and $\beta$ was set to 1.
	
	For global disease diagnosis, we used RadBERT-style classifiers to extract disease categories from CT reports and then computed the corresponding classification F1 score between generated and reference reports. For chest CT reports, we directly adopted the RadBERT evaluator introduced in the CT-Rate study and used the micro F1 score across 18 thoracic abnormality categories as the RadBERT score. To avoid ambiguity, the precision and recall used here are classification metrics and are distinct from the $R$ and $P$ used in ROUGE-L. The F1 score was defined as
	\begin{equation}
		\mathrm{F1}
		=
		\frac{2 \cdot \mathrm{Precision}_{\mathrm{cls}} \cdot \mathrm{Recall}_{\mathrm{cls}}}
		{\mathrm{Precision}_{\mathrm{cls}} + \mathrm{Recall}_{\mathrm{cls}}},
	\end{equation}
	where $\mathrm{Precision}_{\mathrm{cls}}$ and $\mathrm{Recall}_{\mathrm{cls}}$ denote classification precision and recall, respectively. For abdominal CT reports, we trained an additional classifier based on the same architecture for 30 disease classification and used the resulting micro F1 score for evaluation. We used DeepSeek to annotate 4,765 reports by combining the Findings and Impression sections. This process generated a multi-label dataset in which each report was annotated for 30 abnormalities, based on the Merlin label schema. In addition, we manually annotated 447 reports as an independent test set to evaluate the performance of RadBERT. The model architecture was based on the RadBERT-RoBERTa-4m model, with classification performance reported in Supplementary Tab.~B14.

	For fine-grained caption evaluation, we used the FORTE metric and the Rate-Score metric. The FORTE score was computed as described in Section 4.2.3. Unless otherwise stated, all FORTE scores reported in this paper denote the overall FORTE score, aggregated across the degree, landmark, feature and impression dimensions. For Rate-Score, we calculated the score separately for each anatomical region and excluded regions that were negative in both the generated and reference reports. This design avoids inflating the overall score because of the large proportion of trivially negative regions and allows Rate-Score to more accurately reflect the model’s ability to describe positive abnormalities at a fine-grained level.
	
	For baseline comparison, we included four categories of models: (i) general-purpose MLLMs, represented by the state-of-the-art closed-source model Gemini-3 and the open-source model Qwen3-VL; (ii) medical generalist MLLMs, including HuluMed, Lingshu, and Med-Gemma; (iii) radiology-specialized MLLMs, including RadFM and M3D; and (iv) dataset-specific expert models, including 3D-enhanced R2GenGPT and 3D-enhanced LLaVA. For all 2D MLLMs, we followed the strategy used in HuluMed and converted each CT volume into a multi-image input by uniformly sampling 12 slices. For FORTE and Rate-Score evaluation, we used the DeepSeek API to extract region-specific diagnostic content from the generated reports; the detailed prompt is provided in the supplementary Tab.~B8,9. For 3D-enhanced R2GenGPT and 3D-enhanced LLaVA, we used Merlin as the 3D visual encoder and Qwen2.5-VL as the language decoder, and trained each model separately on each standardized dataset for 10 epochs.
	
	\subsection{Ensemble strategy with pretrained foundation models}
	To investigate whether Astra can be integrated with other pretrained foundation models to further improve the performance and efficiency of downstream classification tasks, we designed an ensemble strategy. The core idea is to use the report generated by Astra as preliminary diagnostic guidance for a pretrained vision model. Given an input CT scan, Astra first generates a diagnostic report, which is then encoded into a high-dimensional representation using the Qwen3-Embedding model. This text representation is subsequently projected into the feature space of the visual encoder through a linear layer. In parallel, the pretrained vision model extracts visual-semantic features from the CT image. The projected report features and visual features are then concatenated and fed into a multimodal predictor, implemented as a linear layer, to obtain the final classification output. For the baseline setting, the visual-semantic features extracted by the pretrained vision model were directly passed to a predictor consisting of a linear layer, without incorporating the Astra-generated report.

	\subsection{Scaling CT vision--language pretraining with Astra-generated reports}
	
	The scale of CT vision--language pretraining is limited by the scarcity of paired CT--report datasets, despite the abundance of unpaired CT scans. We therefore tested whether Astra-generated reports could expand the pretraining corpus for CT vision--language models. Astra generated detailed diagnostic reports for 84{,}839 low-dose chest CT scans from NLST, and the resulting synthetic CT--report pairs were divided into four equal subsets. These subsets were progressively mixed with the 24{,}128 paired CT-Rate training samples at ratios of 25\%, 50\%, 75\% and 100\%. Contrastive pretraining was performed using CT-CLIP, with CT-ViT encoding volumetric CT scans into compact visual tokens and CXR-BERT encoding radiology reports. The visual and textual features were projected into a shared 512-dimensional embedding space for contrastive alignment.
	
	To assess the capability of different pretrained models, we focused primarily on downstream classification tasks. We considered two adaptation settings: linear probing and full tuning. In both cases, a linear classification head was appended to the CT-CLIP vision encoder for disease prediction. For linear probing, the vision encoder was kept frozen and only the linear head was optimized, such that performance mainly reflected the quality of the pretrained visual representations. For full tuning, the entire vision encoder was optimized end-to-end to more fully leverage the medical knowledge acquired during pretraining.
	
	\subsection{Implementation details}
	\subsubsection{Training settings}
	For Astra development, the SFT stage used AdamW with an initial learning rate of $3 \times 10^{-5}$, $\beta = (0.9, 0.999)$, and a cosine learning-rate scheduler. Training was run for 10 epochs in bfloat16 mixed precision on four 96-GB NVIDIA H20 GPUs with a batch size of 16. The subsequent GRPO stage followed the R1-V framework with 8 rollouts per sample and a KL regularization coefficient of 0.004. AdamW was again used for optimization, with a learning rate of $1 \times 10^{-6}$. GRPO training was performed for 2 epochs with DeepSpeed ZeRO-3 on eight 96-GB NVIDIA H20 GPUs, using a batch size of 8 and gradient accumulation over 2 steps.

	For pretraining, we used AdamW with an initial learning rate of $1.25 \times 10^{-6}$ and a cosine learning-rate scheduler. Training was performed on four 96-GB NVIDIA H20 GPUs with a per-device batch size of 8. For downstream classification and ensemble experiments, linear probing used a learning rate of $2 \times 10^{-5}$ and full-tune used $2 \times 10^{-6}$. Both settings were run on a single 96-GB NVIDIA H20 GPU with a batch size of 16.
	
	\subsubsection{Preprocessing of CT}
	For CT preprocessing, the Merlin encoder and the CT-CLIP encoder followed different pipelines to remain as consistent as possible with their respective pretraining settings. All preprocessing steps were implemented using the MONAI library. For chest CT images in \texttt{.nii.gz} format processed by the Merlin encoder, images were first reoriented to the RAS coordinate system. A foreground cropping operation was then applied, using a threshold of $-1000$ Hounsfield units (HU) to extract the body region, enlarge the anatomical field of view and reduce background interference. Each scan was resampled to 160 slices, and each slice was resized to $224 \times 224$. Intensity normalization was performed using a window of $[-1000, 200]$ HU. For CTRG-Chest data, the original multi-panel JPG files were converted back into image matrices by extracting pixel values and stitching them into volumetric form. As these values had already been projected to the $0$--$255$ range, the foreground threshold was set to 42 and the normalization range was set to $[42, 246]$, based on estimated HU conversion from the metadata annotations. An additional circular mask was applied because patient information annotations were present in the four corners of the CTRG images.
	
	For abdominal CT images processed by the Merlin encoder, scans were first reoriented to RAS and then resampled to a uniform spatial resolution of $1.5 \times 1.5 \times 3.0$ mm along the $x$, $y$ and $z$ axes. A centered region of interest of size $224 \times 224 \times 160$ was cropped as the model input. Intensity values were normalized using a window of $[-1000, 1000]$ HU.
	
	For the CT-CLIP encoder, a unified preprocessing pipeline was applied to both chest and abdominal CT scans. Images were first reoriented to the RAS coordinate system and then resampled to a uniform spatial resolution of $0.75 \times 0.75 \times 1.5$ mm along the $x$, $y$ and $z$ axes. Intensity values were windowed to $[-1000, 1000]$ HU and normalized to the range $[-1, 1]$. Finally, a centered region of interest of size $480 \times 480 \times 240$ was cropped from each scan as the visual input.

	\subsection{Human--AI collaboration study}
	In this study, we performed a human–AI collaboration experiment to assess whether Astra could improve clinical workflow. We randomly sampled 40 abdominal CT image–report pairs from the inhouse-abdomen-3 dataset and 40 chest CT image–report pairs from the inhouse-chest-1 dataset. We recruited three junior radiologists who had obtained medical licensure within the preceding 5 years and three mid-level radiologists who had obtained medical licensure 5–10 years earlier. For each subspecialty, two junior radiologists and two mid-level radiologists participated in the experiment.
	
	In the first stage, we evaluated whether Astra-assisted reporting reduced drafting time and improved reporting efficiency. A crossover design was used to ensure a fair comparison. For each modality, the 40 CT image–report pairs were randomly divided into two subsets. In the first round, each radiologist drafted reports for one subset with Astra assistance and for the other subset without assistance. Following a 2-week washout period, the assistance conditions were reversed, such that each subset was assessed once under each condition. In the second stage, all reports generated with and without Astra assistance were independently reviewed and scored in a blinded manner by two additional senior radiologists, each with more than 10 years of post-licensure clinical experience, with one responsible for chest CT evaluation and the other for abdominal CT evaluation.
	
	\clearpage

	\bibliographystyle{unsrtnat}
	\bibliography{sn-bibliography}

\begin{thebibliography}{64}
\providecommand{\natexlab}[1]{#1}
\providecommand{\url}[1]{\texttt{#1}}
\expandafter\ifx\csname urlstyle\endcsname\relax
  \providecommand{\doi}[1]{doi: #1}\else
  \providecommand{\doi}{doi: \begingroup \urlstyle{rm}\Url}\fi

\bibitem[Otoni et~al.(2017)Otoni, Noschang, Okamoto, Vieira, Petry,
  de~Araujo~Ramos, Barbosa, Bitencourt, and Chojniak]{otoni2017role}
Jessyca~Couto Otoni, Julia Noschang, Th{\'a}bata~Yaedu Okamoto, Diego~Rosseman
  Vieira, Michel Souto~Mayor Petry, Lucas de~Araujo~Ramos, Paula Nicole
  Vieira~Pinto Barbosa, Almir Galv{\~a}o~Vieira Bitencourt, and Rubens
  Chojniak.
\newblock Role of computed tomography at a cancer center emergency department.
\newblock \emph{Emergency radiology}, 24\penalty0 (2):\penalty0 113--117, 2017.

\bibitem[Sch{\"o}ckel et~al.(2020)Sch{\"o}ckel, Jost, Seidensticker, Lengsfeld,
  Palkowitsch, and Pietsch]{schockel2020developments}
Laura Sch{\"o}ckel, Gregor Jost, Peter Seidensticker, Philipp Lengsfeld, Petra
  Palkowitsch, and Hubertus Pietsch.
\newblock Developments in x-ray contrast media and the potential impact on
  computed tomography.
\newblock \emph{Investigative radiology}, 55\penalty0 (9):\penalty0 592--597,
  2020.

\bibitem[Kanal et~al.(2017)Kanal, Butler, Sengupta, Bhargavan-Chatfield,
  Coombs, and Morin]{kanal2017us}
Kalpana~M Kanal, Priscilla~F Butler, Debapriya Sengupta, Mythreyi
  Bhargavan-Chatfield, Laura~P Coombs, and Richard~L Morin.
\newblock Us diagnostic reference levels and achievable doses for 10 adult ct
  examinations.
\newblock \emph{Radiology}, 284\penalty0 (1):\penalty0 120--133, 2017.

\bibitem[Udare et~al.(2022)Udare, Agarwal, Dhindsa, Alaref, Patlas, Alabousi,
  Kagoma, and van~der Pol]{udare2022radiologist}
Amar Udare, Minu Agarwal, Kiret Dhindsa, Amer Alaref, Michael Patlas, Abdullah
  Alabousi, Yoan~K Kagoma, and Christian~B van~der Pol.
\newblock Radiologist productivity analytics: factors impacting abdominal
  pelvic ct exam reporting times.
\newblock \emph{Journal of Digital Imaging}, 35\penalty0 (2):\penalty0 87--97,
  2022.

\bibitem[Meng et~al.(2023)Meng, Zhan, Liu, and Zhang]{meng2023growing}
Fanyang Meng, Lan Zhan, Shiyuan Liu, and Huimao Zhang.
\newblock The growing problem of radiologist shortage: China’s perspective.
\newblock \emph{Korean Journal of Radiology}, 24\penalty0 (11):\penalty0 1046,
  2023.

\bibitem[Zhang et~al.(2020)Zhang, Liu, Shen, Li, Sang, Wu, Zha, Liang, Wang,
  Wang, et~al.]{zhang2020clinically}
Kang Zhang, Xiaohong Liu, Jun Shen, Zhihuan Li, Ye~Sang, Xingwang Wu, Yunfei
  Zha, Wenhua Liang, Chengdi Wang, Ke~Wang, et~al.
\newblock Clinically applicable ai system for accurate diagnosis, quantitative
  measurements, and prognosis of covid-19 pneumonia using computed tomography.
\newblock \emph{Cell}, 181\penalty0 (6):\penalty0 1423--1433, 2020.

\bibitem[Hu et~al.(2025{\natexlab{a}})Hu, Xia, Zheng, Cao, Zheng, Chen, Sun,
  Chen, Zheng, Pan, et~al.]{hu2025ai}
Can Hu, Yingda Xia, Zhilin Zheng, Mengxuan Cao, Guoliang Zheng, Shangqi Chen,
  Jiancheng Sun, Wujie Chen, Qi~Zheng, Siwei Pan, et~al.
\newblock Ai-based large-scale screening of gastric cancer from noncontrast ct
  imaging.
\newblock \emph{Nature Medicine}, 31\penalty0 (9):\penalty0 3011--3019,
  2025{\natexlab{a}}.

\bibitem[Hu et~al.(2025{\natexlab{b}})Hu, Xiang, Zhou, He, Lang, Yang, Du, Den,
  Xu, Wang, et~al.]{hu2025aa}
Yujian Hu, Yilang Xiang, Yan-Jie Zhou, Yangyan He, Dehai Lang, Shifeng Yang,
  Xiaolong Du, Chunlan Den, Youyao Xu, Gaofeng Wang, et~al.
\newblock Ai-based diagnosis of acute aortic syndrome from noncontrast ct.
\newblock \emph{Nature Medicine}, 31\penalty0 (11):\penalty0 3832--3844,
  2025{\natexlab{b}}.

\bibitem[Cao et~al.(2023)Cao, Xia, Yao, Han, Lambert, Zhang, Tang, Jin, Jiang,
  Fang, et~al.]{cao2023large}
Kai Cao, Yingda Xia, Jiawen Yao, Xu~Han, Lukas Lambert, Tingting Zhang, Wei
  Tang, Gang Jin, Hui Jiang, Xu~Fang, et~al.
\newblock Large-scale pancreatic cancer detection via non-contrast ct and deep
  learning.
\newblock \emph{Nature medicine}, 29\penalty0 (12):\penalty0 3033--3043, 2023.

\bibitem[Milam and Koo(2023)]{milam2023current}
ME~Milam and CW~Koo.
\newblock The current status and future of fda-approved artificial intelligence
  tools in chest radiology in the united states.
\newblock \emph{Clinical Radiology}, 78\penalty0 (2):\penalty0 115--122, 2023.

\bibitem[Li et~al.(2025)Li, Chang, Yang, Wu, Chen, Bansal, Chen, Yang, Chen,
  Chen, et~al.]{braingpt}
Cheng-Yi Li, Kao-Jung Chang, Cheng-Fu Yang, Hsin-Yu Wu, Wenting Chen, Hritik
  Bansal, Ling Chen, Yi-Ping Yang, Yu-Chun Chen, Shih-Pin Chen, et~al.
\newblock Towards a holistic framework for multimodal llm in 3d brain ct
  radiology report generation.
\newblock \emph{Nature Communications}, 16\penalty0 (1):\penalty0 2258, 2025.

\bibitem[Chen et~al.(2025{\natexlab{a}})Chen, Bie, Jin, and Chen]{reg2rg}
Zhixuan Chen, Yequan Bie, Haibo Jin, and Hao Chen.
\newblock Large language model with region-guided referring and grounding for
  ct report generation.
\newblock \emph{IEEE Transactions on Medical Imaging}, 2025{\natexlab{a}}.

\bibitem[Hamamci et~al.(2024{\natexlab{a}})Hamamci, Er, and
  Menze]{hamamci2024ct2rep}
Ibrahim~Ethem Hamamci, Sezgin Er, and Bjoern Menze.
\newblock Ct2rep: Automated radiology report generation for 3d medical imaging.
\newblock In \emph{International Conference on Medical Image Computing and
  Computer-Assisted Intervention}, pages 476--486. Springer,
  2024{\natexlab{a}}.

\bibitem[Chen et~al.(2025{\natexlab{b}})Chen, Luo, Bie, and Chen]{diallama}
Zhixuan Chen, Luyang Luo, Yequan Bie, and Hao Chen.
\newblock Dia-llama: Towards large language model-driven ct report generation.
\newblock In \emph{International Conference on Medical Image Computing and
  Computer-Assisted Intervention}, pages 141--151. Springer,
  2025{\natexlab{b}}.

\bibitem[Deng et~al.(2025)Deng, He, Bao, Zhou, Cai, Cai, and
  Chen]{deng2025mvketr}
Xiwei Deng, Xianchun He, Jianfeng Bao, Yudan Zhou, Shuhui Cai, Congbo Cai, and
  Zhong Chen.
\newblock Mvketr: chest ct report generation with multi-view perception and
  knowledge enhancement.
\newblock \emph{IEEE Journal of Biomedical and Health Informatics}, 2025.

\bibitem[Di~Piazza et~al.(2025)Di~Piazza, Lazarus, Nempont, and
  Boussel]{ctagrg}
Theo Di~Piazza, Carole Lazarus, Olivier Nempont, and Loic Boussel.
\newblock Ct-agrg: Automated abnormality-guided report generation from 3d chest
  ct volumes.
\newblock In \emph{2025 IEEE 22nd International Symposium on Biomedical Imaging
  (ISBI)}, pages 01--05. IEEE, 2025.

\bibitem[Kalisch et~al.(2025)Kalisch, H{\"o}rst, Kleesiek, Herrmann, and
  Seibold]{ctgraph}
Hamza Kalisch, Fabian H{\"o}rst, Jens Kleesiek, Ken Herrmann, and Constantin
  Seibold.
\newblock Ct-graph: Hierarchical graph attention network for anatomy-guided ct
  report generation.
\newblock In \emph{Proceedings of the IEEE/CVF International Conference on
  Computer Vision}, pages 6775--6784, 2025.

\bibitem[Hamamci et~al.(2026)Hamamci, Er, Shit, Reynaud, Yang, Guo, Edgar, Xu,
  Kainz, and Menze]{BTB3D}
Ibrahim~Ethem Hamamci, Sezgin Er, Suprosanna Shit, Hadrien Reynaud, Dong Yang,
  Pengfei Guo, Marc Edgar, Daguang Xu, Bernhard Kainz, and Bjoern Menze.
\newblock Better tokens for better 3d: Advancing vision-language modeling in 3d
  medical imaging.
\newblock \emph{Advances in Neural Information Processing Systems},
  38:\penalty0 135074--135102, 2026.

\bibitem[Chen et~al.(2024{\natexlab{a}})Chen, Zhao, Li, Zhong, Wang, Shang,
  Guo, Han, Liu, Liu, et~al.]{3dctgpt}
Hao Chen, Wei Zhao, Yingli Li, Tianyang Zhong, Yisong Wang, Youlan Shang, Lei
  Guo, Junwei Han, Tianming Liu, Jun Liu, et~al.
\newblock 3d-ct-gpt: Generating 3d radiology reports through integration of
  large vision-language models.
\newblock \emph{arXiv preprint arXiv:2409.19330}, 2024{\natexlab{a}}.

\bibitem[Delbrouck et~al.(2025)Delbrouck, Xu, Moll, Thomas, Chen, Ostmeier,
  Azhar, Li, Johnston, Bluethgen, et~al.]{delbrouck2025automated}
Jean-Benoit Delbrouck, Justin Xu, Johannes Moll, Alois Thomas, Zhihong Chen,
  Sophie Ostmeier, Asfandyar Azhar, Kelvin~Zhenghao Li, Andrew Johnston,
  Christian Bluethgen, et~al.
\newblock Automated structured radiology report generation.
\newblock In \emph{Proceedings of the 63rd Annual Meeting of the Association
  for Computational Linguistics (Volume 1: Long Papers)}, pages 26813--26829,
  2025.

\bibitem[Achiam et~al.(2023)Achiam, Adler, Agarwal, Ahmad, Akkaya, Aleman,
  Almeida, Altenschmidt, Altman, Anadkat, et~al.]{gpt4}
Josh Achiam, Steven Adler, Sandhini Agarwal, Lama Ahmad, Ilge Akkaya,
  Florencia~Leoni Aleman, Diogo Almeida, Janko Altenschmidt, Sam Altman,
  Shyamal Anadkat, et~al.
\newblock Gpt-4 technical report.
\newblock \emph{arXiv preprint arXiv:2303.08774}, 2023.

\bibitem[Team et~al.(2023)Team, Anil, Borgeaud, Alayrac, Yu, Soricut,
  Schalkwyk, Dai, Hauth, Millican, et~al.]{team2023gemini}
Gemini Team, Rohan Anil, Sebastian Borgeaud, Jean-Baptiste Alayrac, Jiahui Yu,
  Radu Soricut, Johan Schalkwyk, Andrew~M Dai, Anja Hauth, Katie Millican,
  et~al.
\newblock Gemini: a family of highly capable multimodal models.
\newblock \emph{arXiv preprint arXiv:2312.11805}, 2023.

\bibitem[Wu et~al.(2025)Wu, Zhang, Zhang, Hui, Wang, and Xie]{radfm}
Chaoyi Wu, Xiaoman Zhang, Ya~Zhang, Hui Hui, Yanfeng Wang, and Weidi Xie.
\newblock Towards generalist foundation model for radiology by leveraging
  web-scale 2d\&3d medical data.
\newblock \emph{Nature Communications}, 16\penalty0 (1):\penalty0 7866, 2025.

\bibitem[Bai et~al.(2024)Bai, Du, Huang, Meng, and Zhao]{bai2024m3d}
Fan Bai, Yuxin Du, Tiejun Huang, Max Q-H Meng, and Bo~Zhao.
\newblock M3d: Advancing 3d medical image analysis with multi-modal large
  language models.
\newblock \emph{arXiv preprint arXiv:2404.00578}, 2024.

\bibitem[Jiang et~al.(2025)Jiang, Wang, Song, Hu, Zhou, Pu, Zhang, Yang, Feng,
  Zhou, et~al.]{jiang2025hulu}
Songtao Jiang, Yuan Wang, Sibo Song, Tianxiang Hu, Chenyi Zhou, Bin Pu, Yan
  Zhang, Zhibo Yang, Yang Feng, Joey~Tianyi Zhou, et~al.
\newblock Hulu-med: A transparent generalist model towards holistic medical
  vision-language understanding.
\newblock \emph{arXiv preprint arXiv:2510.08668}, 2025.

\bibitem[Kirchler et~al.(2026)Kirchler, Ferro, Lorenzini, van~de Water, 3,
  Lippert, and Ganna]{kirchler2026large_llmprocess1}
Matthias Kirchler, Matteo Ferro, Veronica Lorenzini, Robin~P van~de Water,
  FinnGen Ganna~Andrea 3, Christoph Lippert, and Andrea Ganna.
\newblock Large language models improve transferability of electronic health
  record-based predictions across countries and coding systems.
\newblock \emph{npj Digital Medicine}, 2026.

\bibitem[Sandmann et~al.(2025)Sandmann, Hegselmann, Fujarski, Bickmann, Wild,
  Eils, and Varghese]{sandmann2025benchmark_llmprocess2}
Sarah Sandmann, Stefan Hegselmann, Michael Fujarski, Lucas Bickmann, Benjamin
  Wild, Roland Eils, and Julian Varghese.
\newblock Benchmark evaluation of deepseek large language models in clinical
  decision-making.
\newblock \emph{Nature medicine}, 31\penalty0 (8):\penalty0 2546--2549, 2025.

\bibitem[Tordjman et~al.(2025)Tordjman, Liu, Yuce, Fauveau, Mei, Hadjadj,
  Bolger, Almansour, Horst, Parihar,
  et~al.]{tordjman2025comparative_llmprocess3}
Mickael Tordjman, Zelong Liu, Murat Yuce, Valentin Fauveau, Yunhao Mei, Jerome
  Hadjadj, Ian Bolger, Haidara Almansour, Carolyn Horst, Ashwin~Singh Parihar,
  et~al.
\newblock Comparative benchmarking of the deepseek large language model on
  medical tasks and clinical reasoning.
\newblock \emph{Nature medicine}, 31\penalty0 (8):\penalty0 2550--2555, 2025.

\bibitem[Hu et~al.(2024)Hu, Liu, Zhu, Lu, and Wu]{hu2024zero_llmprocess4}
Danqing Hu, Bing Liu, Xiaofeng Zhu, Xudong Lu, and Nan Wu.
\newblock Zero-shot information extraction from radiological reports using
  chatgpt.
\newblock \emph{International Journal of Medical Informatics}, 183:\penalty0
  105321, 2024.

\bibitem[Lopez et~al.(2025)Lopez, Swaminathan, Vedula, Narayanan,
  Nateghi~Haredasht, Ma, Liang, Tate, Maddali, Gallo,
  et~al.]{lopez2025clinical_llmprocess5}
Ivan Lopez, Akshay Swaminathan, Karthik Vedula, Sanjana Narayanan, Fateme
  Nateghi~Haredasht, Stephen~P Ma, April~S Liang, Steven Tate, Manoj Maddali,
  Robert~Joseph Gallo, et~al.
\newblock Clinical entity augmented retrieval for clinical information
  extraction.
\newblock \emph{NPJ digital medicine}, 8\penalty0 (1):\penalty0 45, 2025.

\bibitem[Blankemeier et~al.(2024)Blankemeier, Cohen, Kumar, Van~Veen, Gardezi,
  Paschali, Chen, Delbrouck, Reis, Truyts, et~al.]{merlin}
Louis Blankemeier, Joseph~Paul Cohen, Ashwin Kumar, Dave Van~Veen, Syed
  Jamal~Safdar Gardezi, Magdalini Paschali, Zhihong Chen, Jean-Benoit
  Delbrouck, Eduardo Reis, Cesar Truyts, et~al.
\newblock Merlin: A vision language foundation model for 3d computed
  tomography.
\newblock \emph{Research Square}, pages rs--3, 2024.

\bibitem[Nguyen et~al.(2023)Nguyen, Chen, He, and Tan]{nguyen2023pragmatic}
Dang Nguyen, Chacha Chen, He~He, and Chenhao Tan.
\newblock Pragmatic radiology report generation.
\newblock In \emph{Machine Learning for Health (ML4H)}, pages 385--402. PMLR,
  2023.

\bibitem[Guo et~al.(2025)Guo, Yang, Zhang, Song, Wang, Zhu, Xu, Zhang, Ma, Bi,
  et~al.]{guo2025deepseek}
Daya Guo, Dejian Yang, Haowei Zhang, Junxiao Song, Peiyi Wang, Qihao Zhu,
  Runxin Xu, Ruoyu Zhang, Shirong Ma, Xiao Bi, et~al.
\newblock Deepseek-r1 incentivizes reasoning in llms through reinforcement
  learning.
\newblock \emph{Nature}, 645\penalty0 (8081):\penalty0 633--638, 2025.

\bibitem[Bassi et~al.(2025)Bassi, Yavuz, Hamamci, Er, Chen, Li, Menze,
  Decherchi, Cavalli, Wang, et~al.]{atlas}
Pedro~RAS Bassi, Mehmet~Can Yavuz, Ibrahim~Ethem Hamamci, Sezgin Er, Xiaoxi
  Chen, Wenxuan Li, Bjoern Menze, Sergio Decherchi, Andrea Cavalli, Kang Wang,
  et~al.
\newblock Radgpt: Constructing 3d image-text tumor datasets.
\newblock In \emph{Proceedings of the IEEE/CVF International Conference on
  Computer Vision}, pages 23720--23730, 2025.

\bibitem[Hamamci et~al.(2024{\natexlab{b}})Hamamci, Er, Almas, Simsek, Esirgun,
  Dogan, Dasdelen, Wittmann, Simsar, Simsar, et~al.]{ctrate}
Ibrahim~Ethem Hamamci, Sezgin Er, Furkan Almas, Ayse~Gulnihan Simsek,
  Sevval~Nil Esirgun, Irem Dogan, Muhammed~Furkan Dasdelen, Bastian Wittmann,
  Enis Simsar, Mehmet Simsar, et~al.
\newblock A foundation model utilizing chest ct volumes and radiology reports
  for supervised-level zero-shot detection of abnormalities.
\newblock \emph{CoRR}, 2024{\natexlab{b}}.

\bibitem[Huang et~al.(2023)Huang, Huo, Steinberg, Chiang, Lungren, Langlotz,
  Yeung, Shah, and Fries]{huang2023inspect}
Shih-Cheng Huang, Zepeng Huo, Ethan Steinberg, Chia-Chun Chiang, Matthew~P
  Lungren, Curtis~P Langlotz, Serena Yeung, Nigam~H Shah, and Jason~A Fries.
\newblock Inspect: a multimodal dataset for pulmonary embolism diagnosis and
  prognosis.
\newblock \emph{arXiv preprint arXiv:2311.10798}, 2023.

\bibitem[Chen et~al.(2024{\natexlab{b}})Chen, Liu, Liu, Arcucci, and
  Xiong]{chen2024bimcv}
Yinda Chen, Che Liu, Xiaoyu Liu, Rossella Arcucci, and Zhiwei Xiong.
\newblock Bimcv-r: A landmark dataset for 3d ct text-image retrieval.
\newblock In \emph{International Conference on Medical Image Computing and
  Computer-Assisted Intervention}, pages 124--134. Springer,
  2024{\natexlab{b}}.

\bibitem[Zhao et~al.(2024)Zhao, Wu, Zhang, Zhang, Wang, and
  Xie]{zhao2024ratescore}
Weike Zhao, Chaoyi Wu, Xiaoman Zhang, Ya~Zhang, Yanfeng Wang, and Weidi Xie.
\newblock Ratescore: A metric for radiology report generation.
\newblock In \emph{Proceedings of the 2024 Conference on Empirical Methods in
  Natural Language Processing}, pages 15004--15019, 2024.

\bibitem[Zhang et~al.(2025{\natexlab{a}})Zhang, Li, Long, Zhang, Lin, Yang,
  Xie, Yang, Liu, Lin, et~al.]{zhang2025qwen3_embed}
Yanzhao Zhang, Mingxin Li, Dingkun Long, Xin Zhang, Huan Lin, Baosong Yang,
  Pengjun Xie, An~Yang, Dayiheng Liu, Junyang Lin, et~al.
\newblock Qwen3 embedding: Advancing text embedding and reranking through
  foundation models.
\newblock \emph{arXiv preprint arXiv:2506.05176}, 2025{\natexlab{a}}.

\bibitem[Van~der Maaten and Hinton(2008)]{tsne}
Laurens Van~der Maaten and Geoffrey Hinton.
\newblock Visualizing data using t-sne.
\newblock \emph{Journal of machine learning research}, 9\penalty0 (11), 2008.

\bibitem[Yang et~al.(2025)Yang, Li, Yang, Zhang, Hui, Zheng, Yu, Gao, Huang,
  Lv, et~al.]{yang2025qwen3}
An~Yang, Anfeng Li, Baosong Yang, Beichen Zhang, Binyuan Hui, Bo~Zheng, Bowen
  Yu, Chang Gao, Chengen Huang, Chenxu Lv, et~al.
\newblock Qwen3 technical report.
\newblock \emph{arXiv preprint arXiv:2505.09388}, 2025.

\bibitem[Xu et~al.(2025)Xu, Chan, Li, Aljunied, Yuan, Wang, Xiao, Chen, Liu,
  Li, et~al.]{xu2025lingshu}
Weiwen Xu, Hou~Pong Chan, Long Li, Mahani Aljunied, Ruifeng Yuan, Jianyu Wang,
  Chenghao Xiao, Guizhen Chen, Chaoqun Liu, Zhaodonghui Li, et~al.
\newblock Lingshu: A generalist foundation model for unified multimodal medical
  understanding and reasoning.
\newblock \emph{arXiv preprint arXiv:2506.07044}, 2025.

\bibitem[Sellergren et~al.(2025)Sellergren, Kazemzadeh, Jaroensri, Kiraly,
  Traverse, Kohlberger, Xu, Jamil, Hughes, Lau, et~al.]{sellergren2025medgemma}
Andrew Sellergren, Sahar Kazemzadeh, Tiam Jaroensri, Atilla Kiraly, Madeleine
  Traverse, Timo Kohlberger, Shawn Xu, Fayaz Jamil, C{\'\i}an Hughes, Charles
  Lau, et~al.
\newblock Medgemma technical report.
\newblock \emph{arXiv preprint arXiv:2507.05201}, 2025.

\bibitem[Wang et~al.(2023)Wang, Liu, Wang, and Zhou]{wang2023r2gengpt}
Zhanyu Wang, Lingqiao Liu, Lei Wang, and Luping Zhou.
\newblock R2gengpt: Radiology report generation with frozen llms.
\newblock \emph{Meta-Radiology}, 1\penalty0 (3):\penalty0 100033, 2023.

\bibitem[Zambrano~Chaves et~al.(2025)Zambrano~Chaves, Huang, Xu, Xu, Usuyama,
  Zhang, Wang, Xie, Khademi, Yang, et~al.]{llavarad}
Juan~Manuel Zambrano~Chaves, Shih-Cheng Huang, Yanbo Xu, Hanwen Xu, Naoto
  Usuyama, Sheng Zhang, Fei Wang, Yujia Xie, Mahmoud Khademi, Ziyi Yang, et~al.
\newblock A clinically accessible small multimodal radiology model and
  evaluation metric for chest x-ray findings.
\newblock \emph{Nature Communications}, 16\penalty0 (1):\penalty0 3108, 2025.

\bibitem[Gai et~al.(2025)Gai, Liu, Li, Meng, Wu, and Liu]{amos}
Xiaotang Gai, Jiaxiang Liu, Yichen Li, Zijie Meng, Jian Wu, and Zuozhu Liu.
\newblock 3d-rad: A comprehensive 3d radiology med-vqa dataset with
  multi-temporal analysis and diverse diagnostic tasks.
\newblock \emph{arXiv preprint arXiv:2506.11147}, 2025.

\bibitem[Tang et~al.(2024)Tang, Yang, Zhang, and Yuan]{ctrg}
Yuhao Tang, Haichen Yang, Liyan Zhang, and Ye~Yuan.
\newblock Work like a doctor: Unifying scan localizer and dynamic generator for
  automated computed tomography report generation.
\newblock \emph{Expert Systems with Applications}, 237:\penalty0 121442, 2024.

\bibitem[Kramer et~al.(2011)Kramer, Berg, Aberle, and Prorok]{nlst}
Barnett~S Kramer, Christine~D Berg, Denise~R Aberle, and Philip~C Prorok.
\newblock Lung cancer screening with low-dose helical ct: results from the
  national lung screening trial (nlst), 2011.

\bibitem[Draelos et~al.(2021)Draelos, Dov, Mazurowski, Lo, Henao, Rubin, and
  Carin]{radchest}
Rachel~Lea Draelos, David Dov, Maciej~A Mazurowski, Joseph~Y Lo, Ricardo Henao,
  Geoffrey~D Rubin, and Lawrence Carin.
\newblock Machine-learning-based multiple abnormality prediction with
  large-scale chest computed tomography volumes.
\newblock \emph{Medical image analysis}, 67:\penalty0 101857, 2021.

\bibitem[Colak et~al.(2021)Colak, Kitamura, Hobbs, Wu, Lungren, Prevedello,
  Kalpathy-Cramer, Ball, Shih, Stein, et~al.]{colak2021rsna}
Errol Colak, Felipe~C Kitamura, Stephen~B Hobbs, Carol~C Wu, Matthew~P Lungren,
  Luciano~M Prevedello, Jayashree Kalpathy-Cramer, Robyn~L Ball, George Shih,
  Anouk Stein, et~al.
\newblock The rsna pulmonary embolism ct dataset.
\newblock \emph{Radiology: Artificial Intelligence}, 3\penalty0 (2):\penalty0
  e200254, 2021.

\bibitem[Papineni et~al.(2002)Papineni, Roukos, Ward, and
  Zhu]{papineni2002bleu}
Kishore Papineni, Salim Roukos, Todd Ward, and Wei-Jing Zhu.
\newblock Bleu: a method for automatic evaluation of machine translation.
\newblock In \emph{Proceedings of the 40th annual meeting of the Association
  for Computational Linguistics}, pages 311--318, 2002.

\bibitem[Lin(2004)]{lin2004rouge}
Chin-Yew Lin.
\newblock Rouge: A package for automatic evaluation of summaries.
\newblock In \emph{Text summarization branches out}, pages 74--81, 2004.

\bibitem[Banerjee and Lavie(2005)]{banerjee2005meteor}
Satanjeev Banerjee and Alon Lavie.
\newblock Meteor: An automatic metric for mt evaluation with improved
  correlation with human judgments.
\newblock In \emph{Proceedings of the acl workshop on intrinsic and extrinsic
  evaluation measures for machine translation and/or summarization}, pages
  65--72, 2005.

\bibitem[Bannur et~al.(2024)Bannur, Bouzid, Castro, Schwaighofer, Thieme,
  Bond-Taylor, Ilse, P{\'e}rez-Garc{\'\i}a, Salvatelli, Sharma,
  et~al.]{bannur2024maira}
Shruthi Bannur, Kenza Bouzid, Daniel~C Castro, Anton Schwaighofer, Anja Thieme,
  Sam Bond-Taylor, Maximilian Ilse, Fernando P{\'e}rez-Garc{\'\i}a, Valentina
  Salvatelli, Harshita Sharma, et~al.
\newblock Maira-2: Grounded radiology report generation.
\newblock \emph{arXiv preprint arXiv:2406.04449}, 2024.

\bibitem[Hyland et~al.(2023)Hyland, Bannur, Bouzid, Castro, Ranjit,
  Schwaighofer, P{\'e}rez-Garc{\'\i}a, Salvatelli, Srivastav, Thieme,
  et~al.]{hyland2023maira}
Stephanie~L Hyland, Shruthi Bannur, Kenza Bouzid, Daniel~C Castro, Mercy
  Ranjit, Anton Schwaighofer, Fernando P{\'e}rez-Garc{\'\i}a, Valentina
  Salvatelli, Shaury Srivastav, Anja Thieme, et~al.
\newblock Maira-1: A specialised large multimodal model for radiology report
  generation.
\newblock \emph{arXiv preprint arXiv:2311.13668}, 2023.

\bibitem[Chen et~al.(2020)Chen, Song, Chang, and Wan]{r2gen}
Zhihong Chen, Yan Song, Tsung-Hui Chang, and Xiang Wan.
\newblock Generating radiology reports via memory-driven transformer.
\newblock In \emph{Proceedings of the 2020 conference on empirical methods in
  natural language processing (EMNLP)}, pages 1439--1449, 2020.

\bibitem[Chen et~al.(2021)Chen, Shen, Song, and Wan]{r2gencmn}
Zhihong Chen, Yaling Shen, Yan Song, and Xiang Wan.
\newblock Cross-modal memory networks for radiology report generation.
\newblock In \emph{Proceedings of the 59th annual meeting of the association
  for computational linguistics and the 11th international joint conference on
  natural language processing (volume 1: long papers)}, pages 5904--5914, 2021.

\bibitem[Ma et~al.(2025)Ma, Liang, He, Wang, Yan, Li, Wang, Li, Lizhu, Liu,
  et~al.]{medmpt}
Liangdi Ma, Hengrui Liang, Yuwei He, Wei Wang, Zeping Yan, Wuchao Li, Rongpin
  Wang, Yongyi Li, Yuerong Lizhu, Yaou Liu, et~al.
\newblock A vision--language pretrained transformer for versatile clinical
  respiratory disease applications.
\newblock \emph{Nature Biomedical Engineering}, pages 1--19, 2025.

\bibitem[Tanno et~al.(2025)Tanno, Barrett, Sellergren, Ghaisas, Dathathri, See,
  Welbl, Lau, Tu, Azizi, et~al.]{tanno2025collaboration}
Ryutaro Tanno, David~GT Barrett, Andrew Sellergren, Sumedh Ghaisas, Sumanth
  Dathathri, Abigail See, Johannes Welbl, Charles Lau, Tao Tu, Shekoofeh Azizi,
  et~al.
\newblock Collaboration between clinicians and vision--language models in
  radiology report generation.
\newblock \emph{Nature Medicine}, 31\penalty0 (2):\penalty0 599--608, 2025.

\bibitem[Chen et~al.(2025{\natexlab{c}})Chen, Fu, Wang, Lin, Cheng, Li, Wang,
  Chen, Lin, Zhang, et~al.]{octrg}
Xinjian Chen, Huazhu Fu, Jingtao Wang, Tian Lin, Qian Cheng, Cangxin Li, Meng
  Wang, Zhongyue Chen, Aidi Lin, Anlin Zhang, et~al.
\newblock A deep learning based automatic report generator for retinal optical
  coherence tomography images.
\newblock \emph{npj Digital Medicine}, 8\penalty0 (1):\penalty0 618,
  2025{\natexlab{c}}.

\bibitem[Dong et~al.(2025)Dong, Nie, Chen, Xu, and Li]{dong2025keyword}
Fei Dong, Shouping Nie, Manling Chen, Fangfang Xu, and Qian Li.
\newblock Keyword-based ai assistance in the generation of radiology reports: A
  pilot study.
\newblock \emph{NPJ Digital Medicine}, 8\penalty0 (1):\penalty0 490, 2025.

\bibitem[Koetzier et~al.(2024)Koetzier, Wu, Mastrodicasa, Lutz, Chung, Koszek,
  Pratap, Chaudhari, Rajpurkar, Lungren, et~al.]{koetzier2024generating}
Lennart~R Koetzier, Jie Wu, Domenico Mastrodicasa, Aline Lutz, Matthew Chung,
  W~Adam Koszek, Jayanth Pratap, Akshay~S Chaudhari, Pranav Rajpurkar,
  Matthew~P Lungren, et~al.
\newblock Generating synthetic data for medical imaging.
\newblock \emph{Radiology}, 312\penalty0 (3):\penalty0 e232471, 2024.

\bibitem[Ma et~al.(2026)Ma, Li, Li, Liu, Wu, Zhou, Liang, Chan, Wong, and
  Chen]{ma2026generative}
Jiabo Ma, Wenqiang Li, Jinbang Li, Ziyi Liu, Linshan Wu, Fengtao Zhou,
  Li~Liang, Ronald Cheong~Kin Chan, Terence~TW Wong, and Hao Chen.
\newblock Generative ai for misalignment-resistant virtual staining to
  accelerate histopathology workflows.
\newblock \emph{Nature Communications}, 2026.

\bibitem[Zhang et~al.(2025{\natexlab{b}})Zhang, Wu, Zhao, Lei, Tian, Zhang,
  Xie, and Wang]{radgenome}
Xiaoman Zhang, Chaoyi Wu, Ziheng Zhao, Jiayu Lei, Weiwei Tian, Ya~Zhang, Weidi
  Xie, and Yanfeng Wang.
\newblock Development of a large-scale grounded vision language dataset for
  chest ct analysis.
\newblock \emph{Scientific Data}, 12\penalty0 (1):\penalty0 1636,
  2025{\natexlab{b}}.

\end{thebibliography}
	
	\begin{appendices}
		
		\section{Supplementary Figures}\label{secA1}
		\begin{center}
			\includegraphics[width=0.9\textwidth]{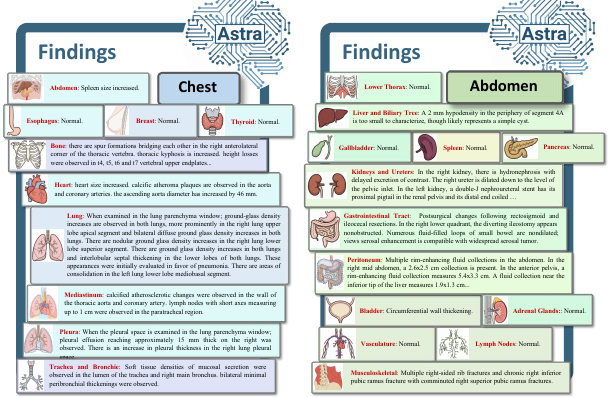}
			\captionof{figure}{\textbf{Representative harmonized chest and abdominal CT report cases.} The original free-text reports are reorganized into predefined anatomical regions, and each region focuses on positive abnormalities while regions without relevant findings are labeled as normal. For chest CT reports, the standardized reporting order is abdomen, bone, breast, esophagus, heart, lung, mediastinum, pleura, thyroid, and trachea and bronchi. For abdominal CT reports, the standardized reporting order is lower thorax, liver and biliary tree, gallbladder, spleen, pancreas, adrenal glands, kidneys and ureters, gastrointestinal tract, peritoneum, pelvis, vasculature, lymph nodes, and musculoskeletal system. For visual clarity, the order of regions shown in the figure may differ from the actual harmonized report order.} 
		\end{center}
		\clearpage
		\begin{center}
			\includegraphics[width=0.9\textwidth]{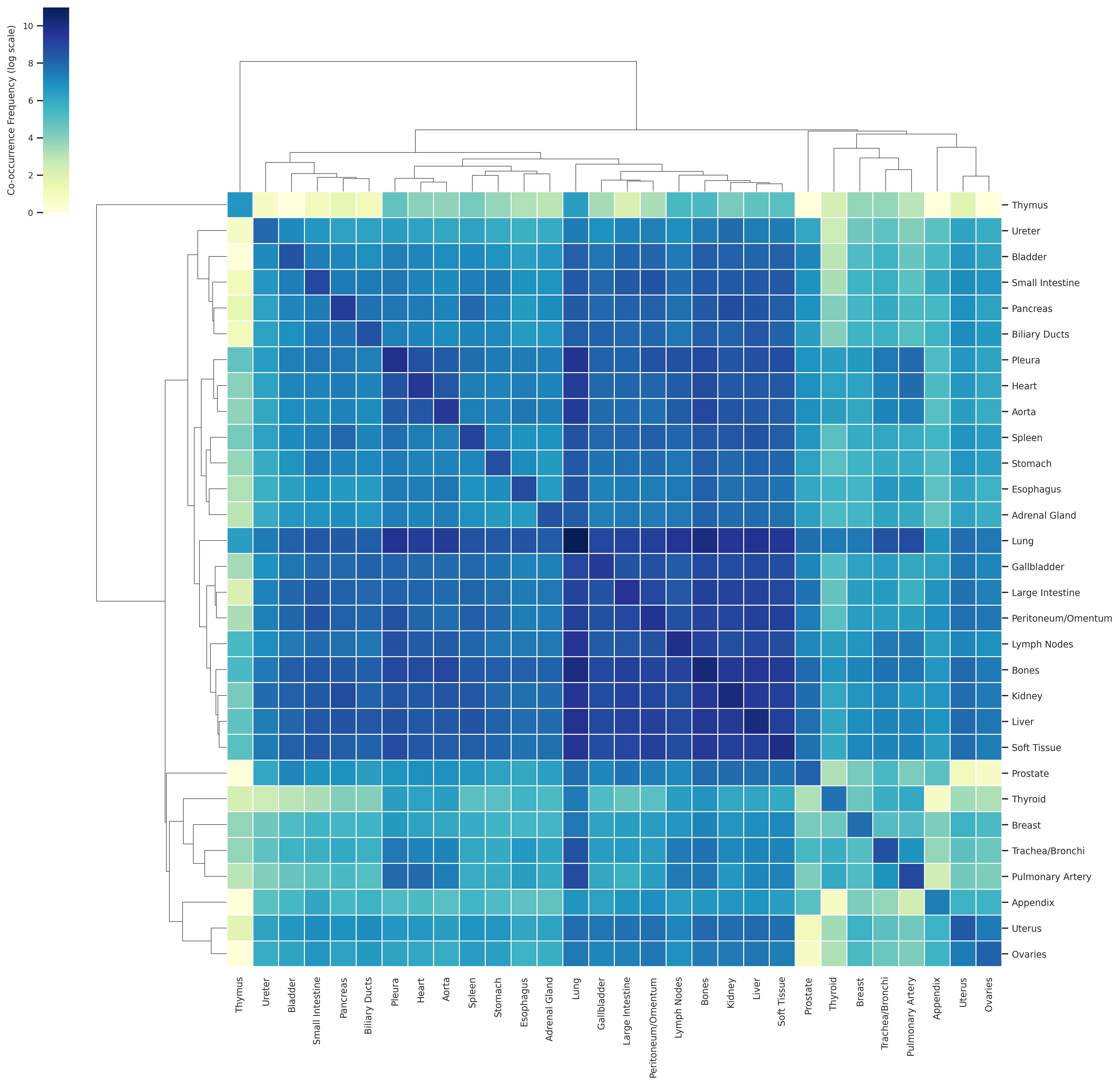}
			\captionof{figure}{\textbf{Organ-level abnormality co-occurrence matrix in the CTRgDB dataset.} The matrix shows multi-organ abnormality co-occurrence across 30 major organs in CTRgDB. In the abdomen, abnormalities of the liver, kidneys, peritoneum and lymph nodes frequently co-occur, consistent with common metastatic spread patterns. In the thorax, lung abnormalities co-occur broadly with abnormalities in multiple other organs. Together, these patterns show that CTRgDB captures complex multi-organ abnormality involvement.} 
		\end{center}
		\clearpage
		
		\begin{center}
			\includegraphics[width=0.9\textwidth]{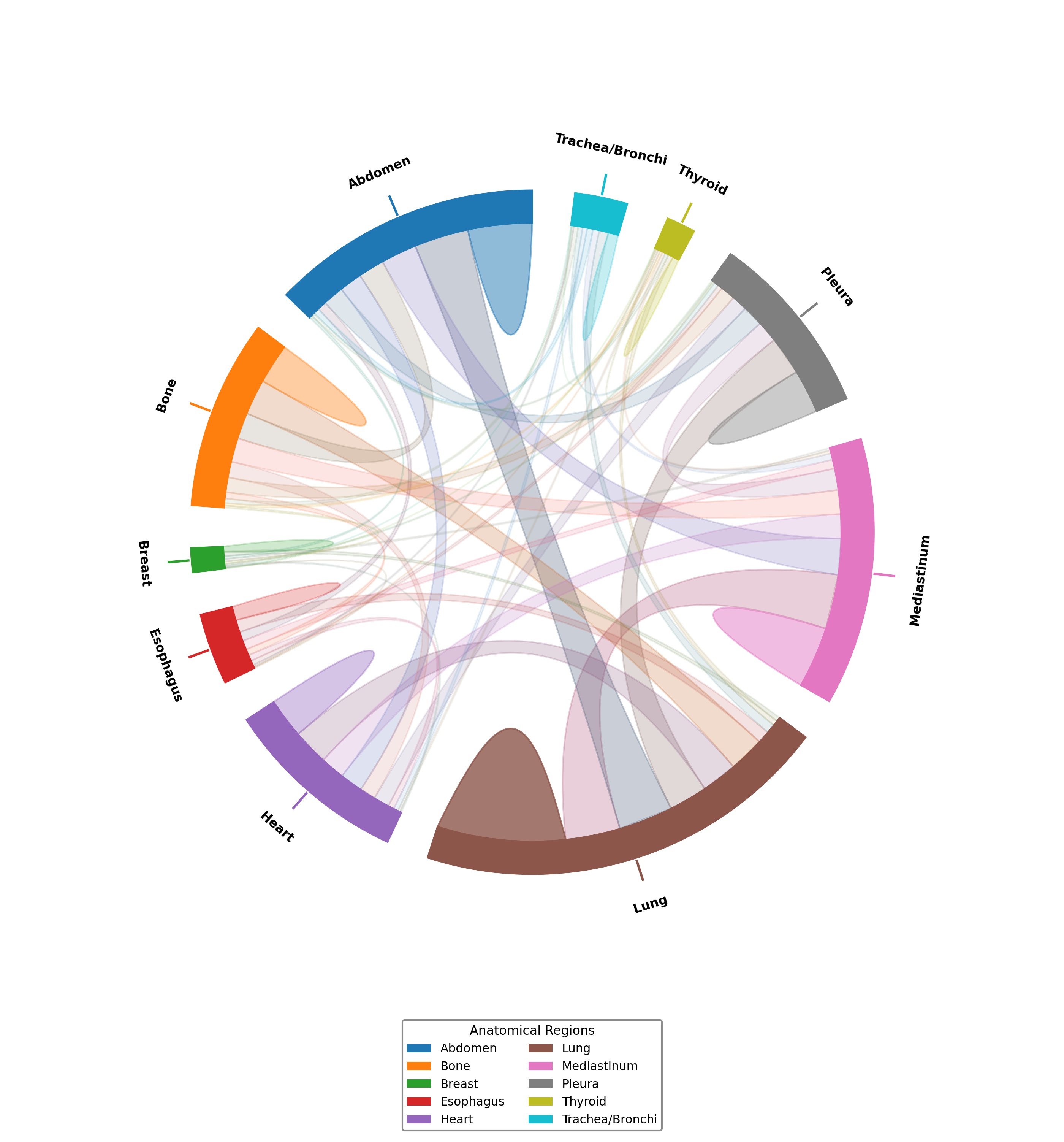}
			\captionof{figure}{\textbf{Region-level abnormality co-occurrence chord diagram for chest CTs in CTRgDB.} The diagram shows the frequency of multi-region abnormality involvement across 10 major thoracic regions in CTRgDB. Abnormalities in the mediastinum frequently co-occur with those in the lungs, consistent with the common spread of neoplastic and inflammatory processes to the mediastinal region. CTRgDB captures complex multi-region abnormality involvement in chest CT examinations.} 
		\end{center}
		\clearpage
		
		\begin{center}
			\includegraphics[width=0.9\textwidth]{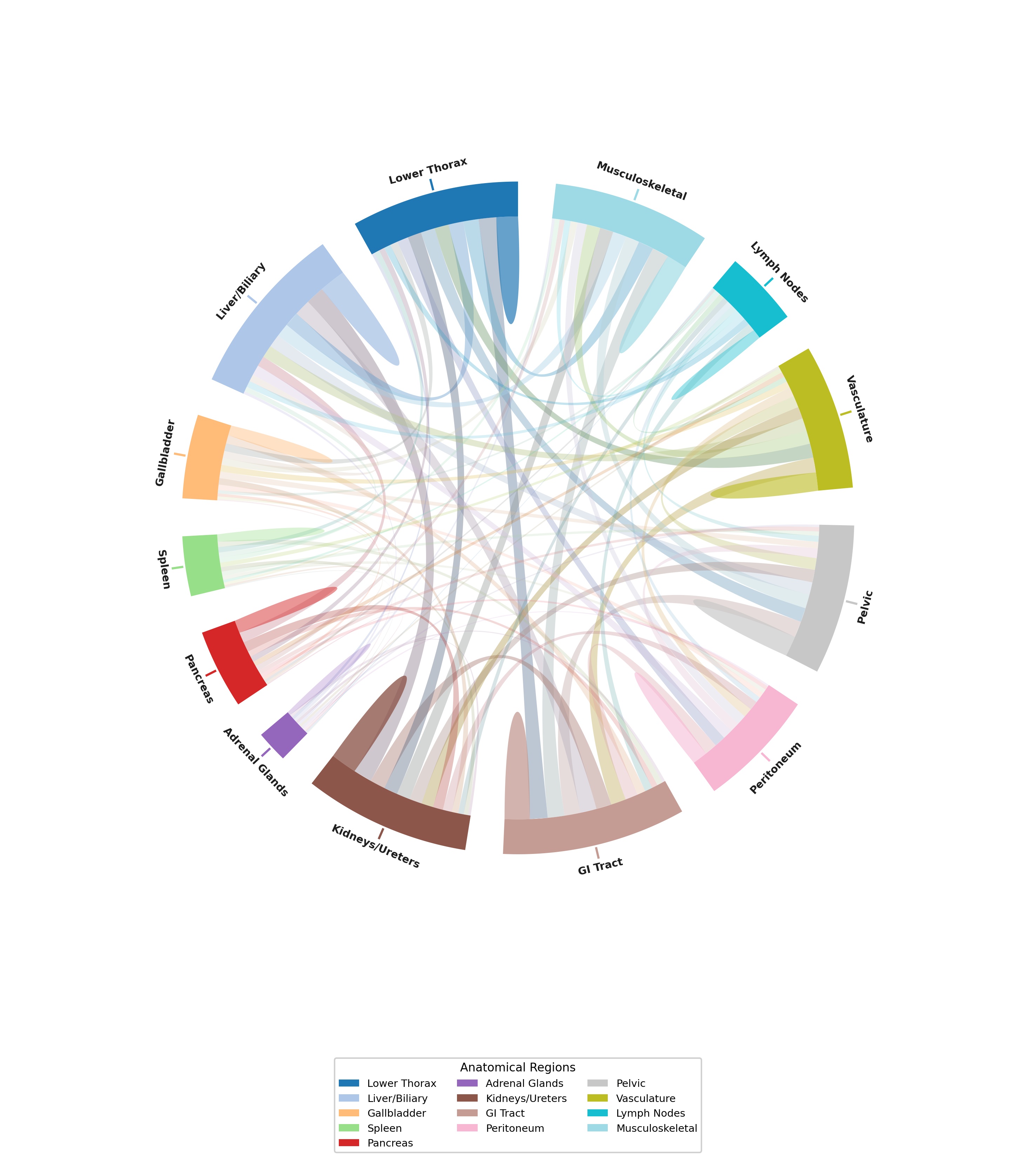}
			\captionof{figure}{\textbf{Region-level abnormality co-occurrence chord diagram for abdominal CTs in CTRgDB.} The diagram shows the frequency of multi-region abnormality involvement across 13 major abdominal regions in CTRgDB. CTRgDB captures complex multi-region abnormality involvement in abdominal CT examinations.} 
		\end{center}
		\clearpage
		
		\begin{center}
			\includegraphics[width=0.9\textwidth]{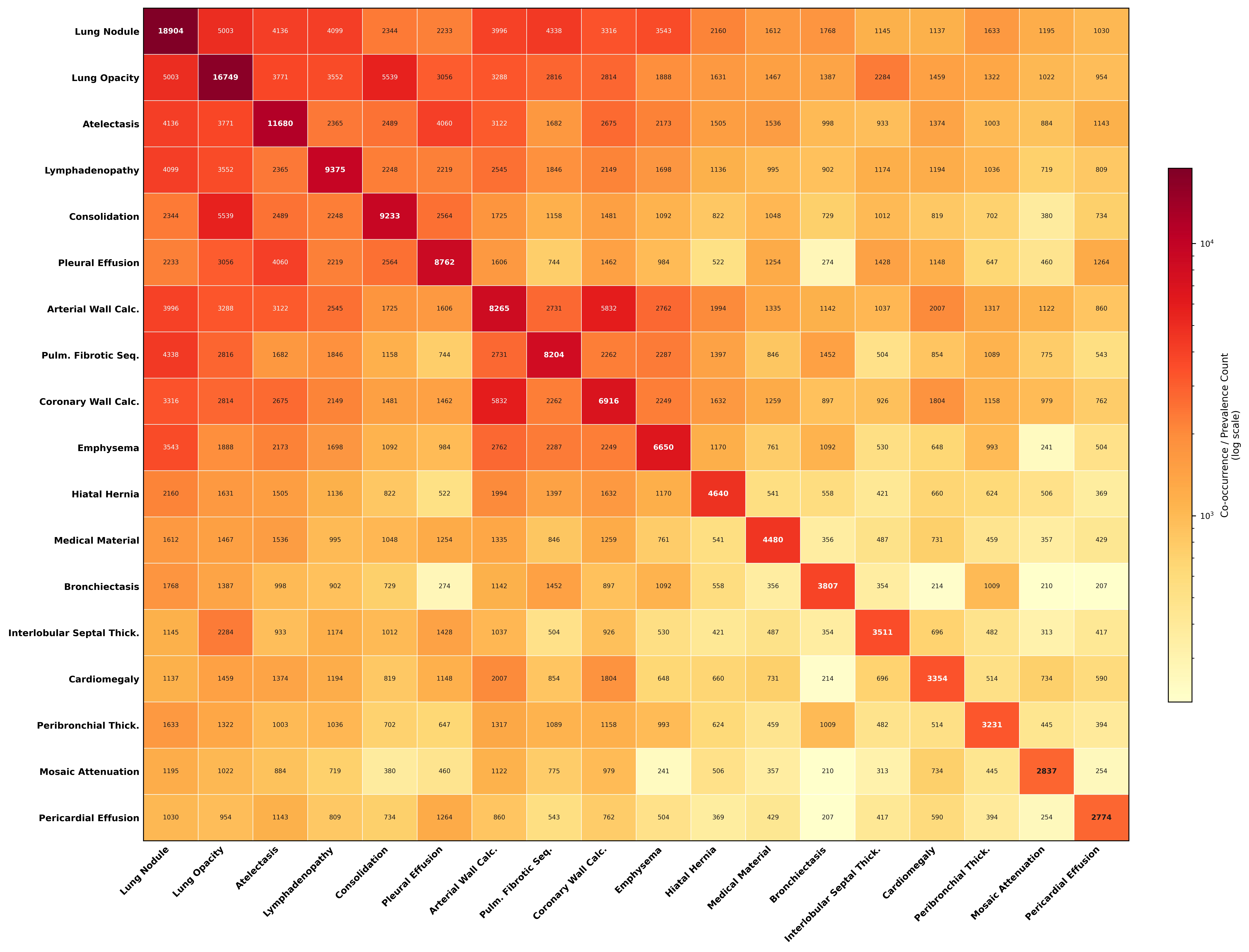}
			\captionof{figure}{\textbf{Disease-level co-occurrence matrix for chest CTs in CTRgDB.} The matrix shows co-occurrence patterns among 18 major diseases defined in CT-Rate, highlighting frequent multi-disease involvement in individual chest CT examinations. These results demonstrate that CTRgDB captures complex disease-level abnormalities in chest CTs.} 
		\end{center}
		\clearpage
		
		\begin{center}
			\includegraphics[width=0.9\textwidth]{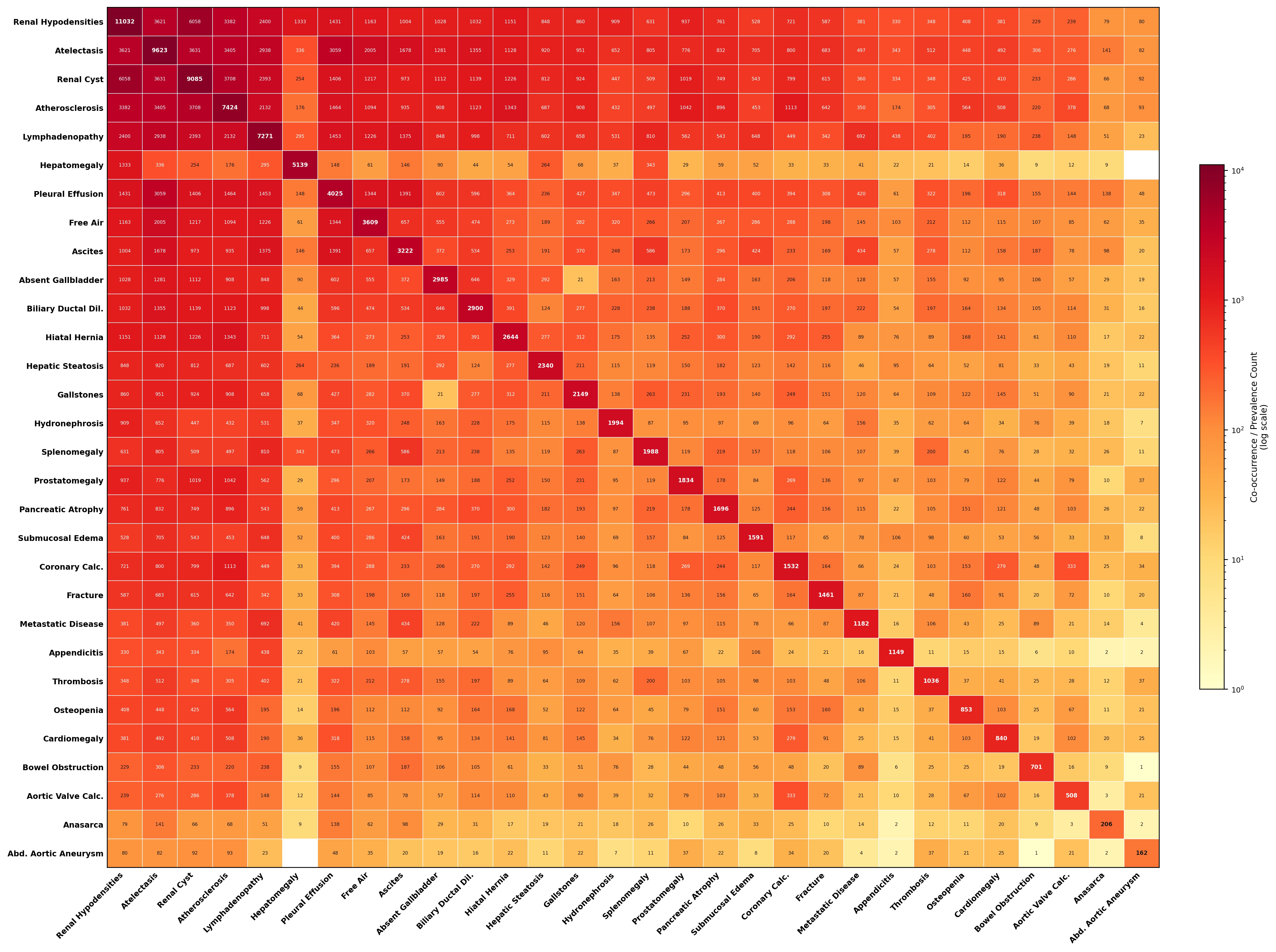}
			\captionof{figure}{\textbf{Disease-level co-occurrence matrix for abdominal CTs in CTRgDB.} The matrix illustrates co-occurrence patterns among 30 major abdominal diseases defined with reference to Merlin, revealing frequent coexisting disease entities within individual abdominal CT examinations. This distribution supports the ability of CTRgDB to represent complex disease-level involvement in abdominal CTs.} 
		\end{center}
		\clearpage
		
		\begin{center}
			\includegraphics[width=0.9\textwidth]{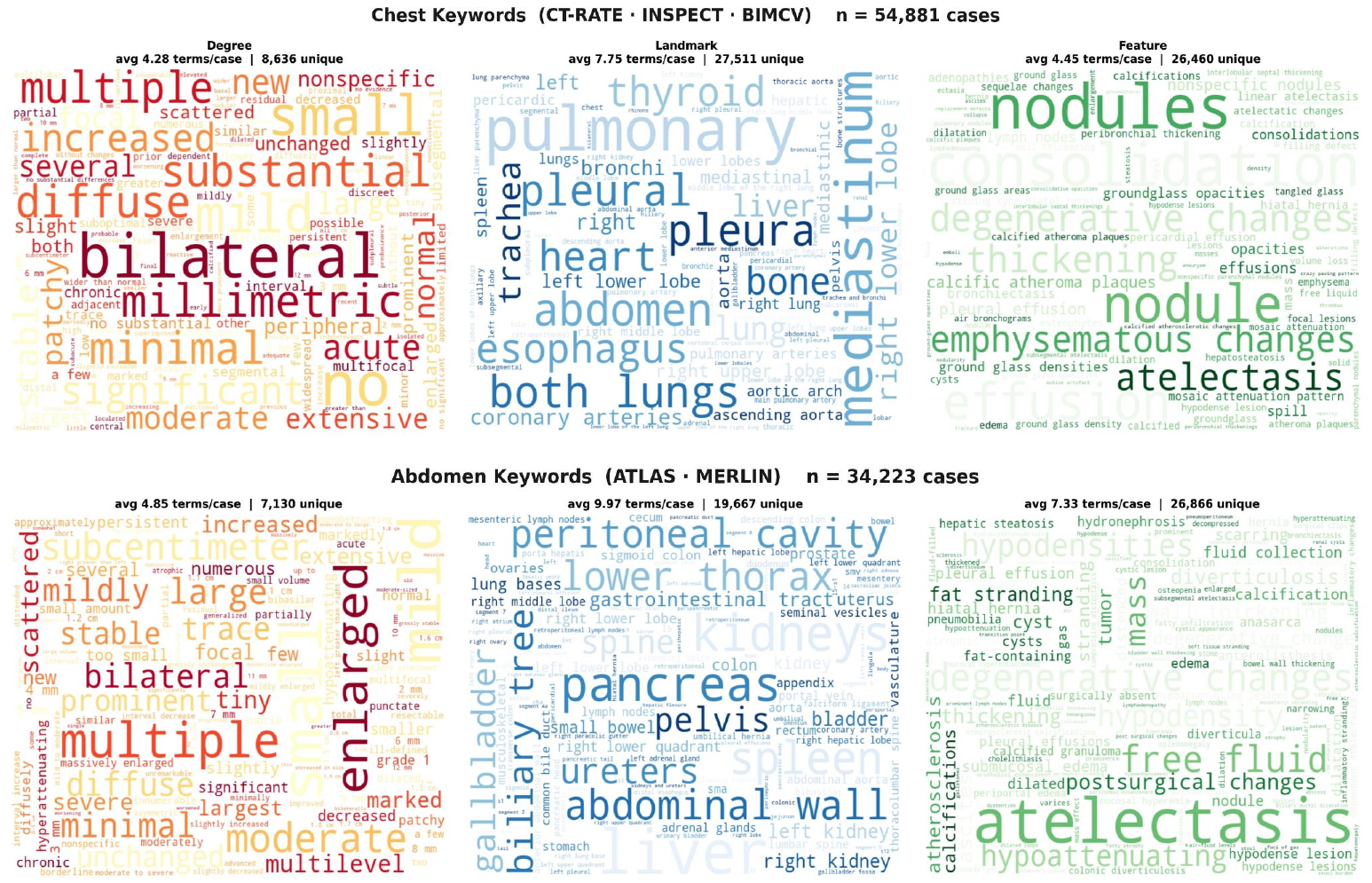}
			\captionof{figure}{\textbf{Word clouds in CTRgDB reports.} The word clouds highlight the rich diagnostic vocabulary used in chest and abdominal CT reports in CTRgDB. Reports frequently describe abnormalities in terms of degree, landmark and feature, enabling precise and clinically informative characterization. These patterns indicate that CTRgDB provides detailed diagnostic descriptions at the finding level.} 
		\end{center}
		\clearpage

		\begin{center}
			\includegraphics[width=0.9\textwidth]{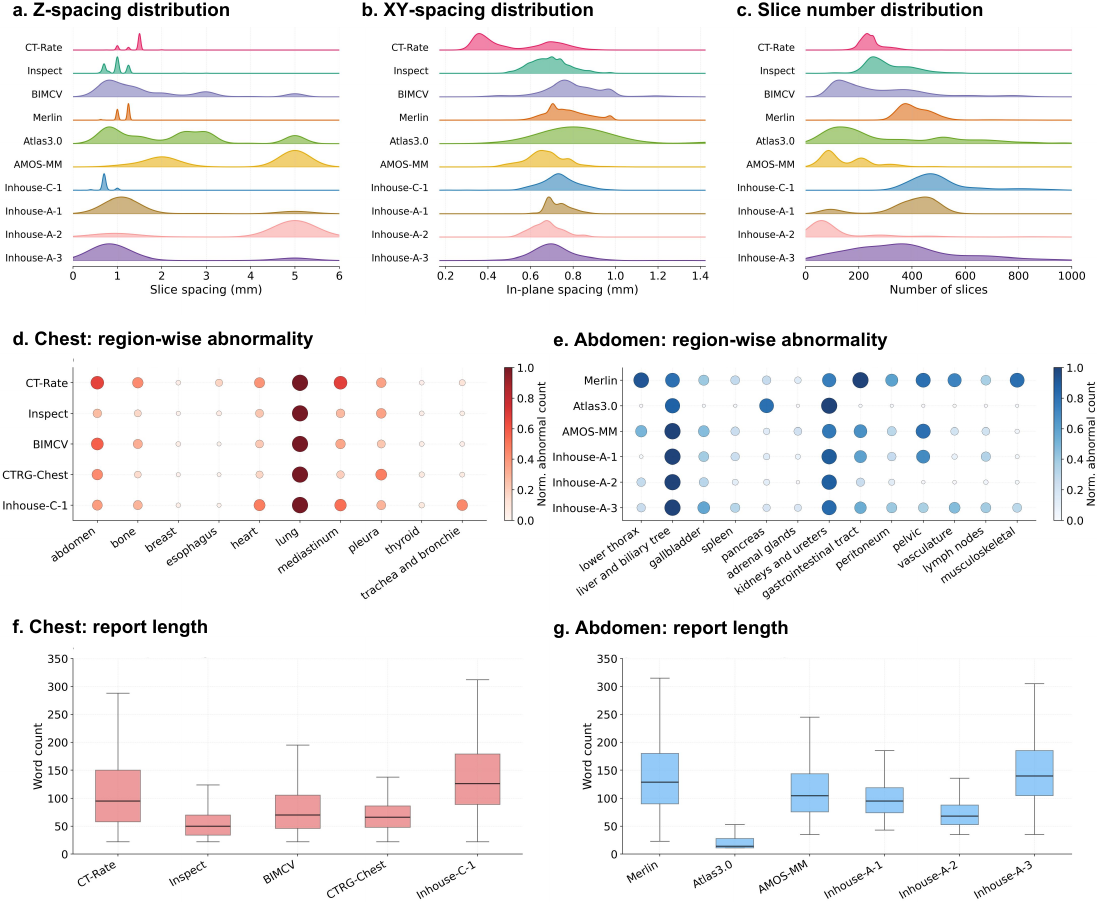}
			\captionof{figure}{\textbf{Heterogeneity of imaging and reporting distributions across CT cohorts.}
				\textbf{a--c}, Distributions of slice spacing, in-plane XY spacing and slice number across cohorts.
				\textbf{d,e}, Region-wise abnormality profiles in chest and abdominal cohorts, with dot size and colour denoting normalized abnormality counts.
				\textbf{f,g}, Report length distributions in chest and abdominal cohorts, measured by word count.
				The cohorts showed marked differences in acquisition parameters, anatomical abnormality distributions and reporting granularity, supporting a comprehensive assessment of cross-cohort generalizability. All statistics were computed from the full datasets, including training, validation and test splits.} 
		\end{center}
		\clearpage
		
		\begin{center}
			\includegraphics[width=0.9\textwidth]{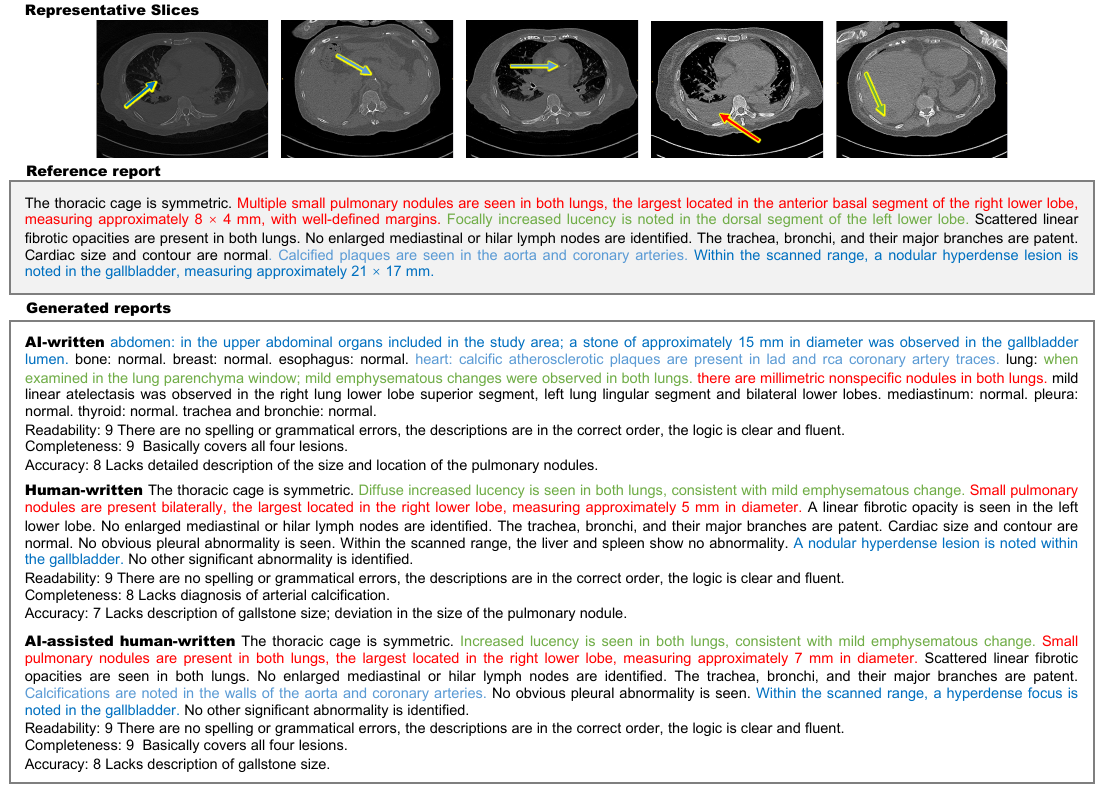}
			\captionof{figure}{\textbf{Case study of human-AI collaboration in chest CT report generation.} Astra identified all abnormalities but provided insufficient characterization of a pulmonary nodule. The radiologist missed arteriosclerosis. Human–AI collaboration accurately detected all abnormalities with precise characterization.} 
		\end{center}
		\clearpage
		
		\begin{center}
			\includegraphics[width=0.9\textwidth]{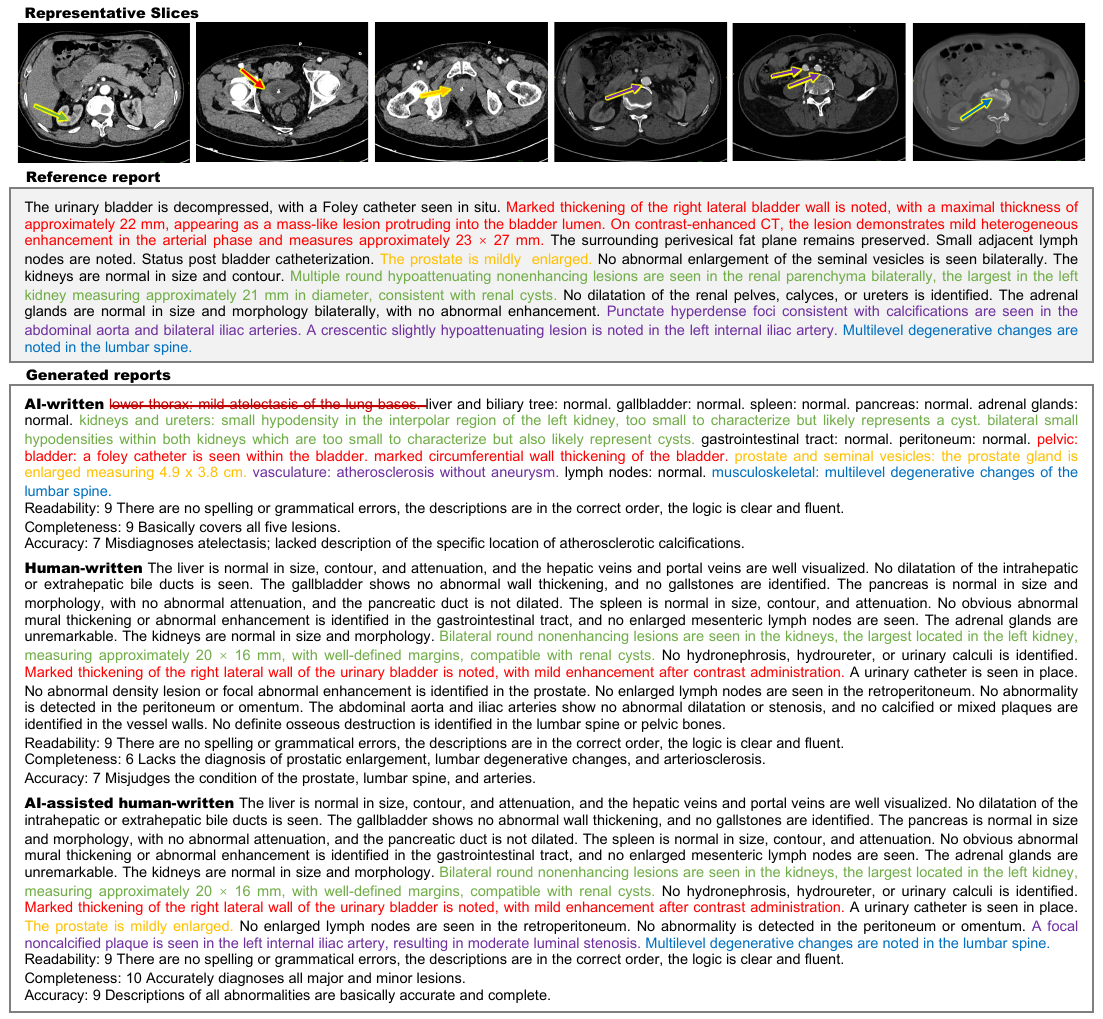}
			\captionof{figure}{\textbf{Case study of human-AI collaboration in abdomen CT report generation. } Astra identified bladder tumour, renal cyst, enlarged prostate, arteriosclerosis, and spinal degeneration, but misdiagnosed atelectasis. The radiologist missed enlarged prostate, arteriosclerosis, and spinal degeneration. Human–AI collaboration correctly detected all abnormalities, provided accurate characterizations, and corrected the initial misdiagnosis.} 
		\end{center}
		\clearpage

		\begin{center}
			\includegraphics[width=0.9\textwidth]{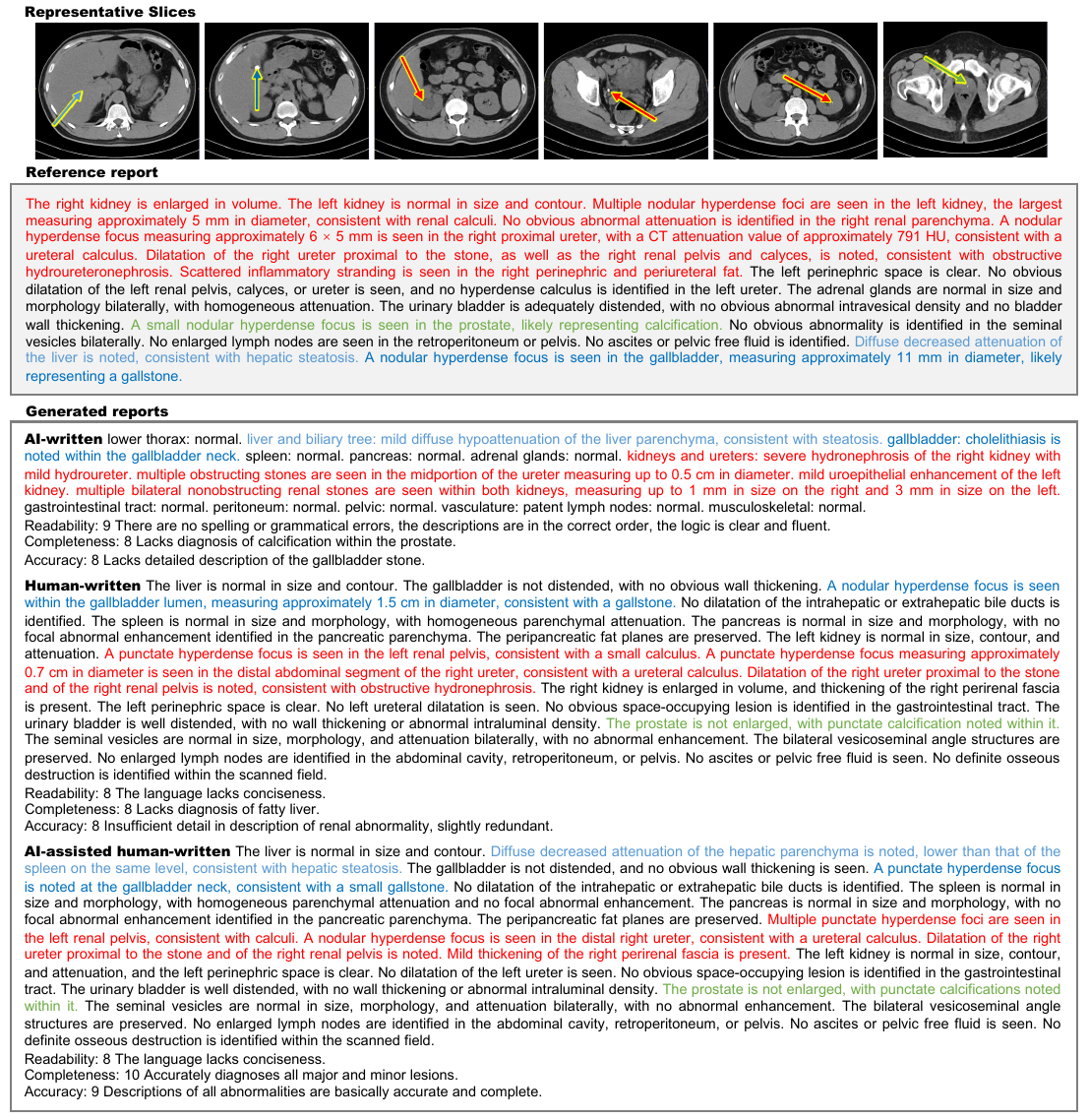}
			\captionof{figure}{\textbf{ Case study of human-AI collaboration in abdomen CT report generation. } Astra missed prostatic calcification, while the radiologist missed hepatic steatosis. Human–AI collaboration accurately identified all abnormalities with precise characterization.} 
		\end{center}
		\clearpage
		
		\section{Supplementary Tables}\label{secB1}
		\begin{table*}[htbp]
			\resizebox{\textwidth}{!}{%
				\begin{tabularx}{\textwidth}{ll >{\centering\arraybackslash}X >{\centering\arraybackslash}X >{\centering\arraybackslash}X}
					\toprule
					\multirow{2}{*}{\textbf{Dataset}}
					& \multirow{2}{*}{\textbf{Model}}
					& \textbf{micro Precision} & \textbf{micro Recall} & \textbf{micro F1} \\
					& & \scriptsize{[95\% CI]} & \scriptsize{[95\% CI]} & \scriptsize{[95\% CI]} \\
					\midrule
					
					\multirow{6}{*}{\textbf{CT-Rate}}
					& \multirow{2}{*}{Astra-Base-Ori}
					& 0.5618 & 0.3694 & 0.4458 \\
					& & \scriptsize{[0.5431, 0.5803]}
					& \scriptsize{[0.3538, 0.3838]}
					& \scriptsize{[0.4302, 0.4596]} \\
					\cmidrule{2-5}
					
					& \multirow{2}{*}{Astra-Base}
					& 0.5613 & 0.4301$^*$ & 0.4870$^*$ \\
					& & \scriptsize{[0.5445, 0.5780]}
					& \scriptsize{[0.4174, 0.4428]}
					& \scriptsize{[0.4748, 0.4994]} \\
					\cmidrule{2-5}
					
					& \multirow{2}{*}{Astra}
					& 0.5789$^{**}$ & 0.5242$^{**}$ & 0.5502$^{**}$ \\
					& & \scriptsize{[0.5632, 0.5948]}
					& \scriptsize{[0.5103, 0.5381]}
					& \scriptsize{[0.5380, 0.5628]} \\
					
					\midrule
					
					\multirow{6}{*}{\textbf{BIMCV}}
					& \multirow{2}{*}{Astra-Base-Ori}
					& 0.3112 & 0.2739 & 0.2914 \\
					& & \scriptsize{[0.2927, 0.3301]}
					& \scriptsize{[0.2568, 0.2917]}
					& \scriptsize{[0.2750, 0.3078]} \\
					\cmidrule{2-5}
					
					& \multirow{2}{*}{Astra-Base}
					& 0.3398$^*$ & 0.2578 & 0.2932 \\
					& & \scriptsize{[0.3186, 0.3601]}
					& \scriptsize{[0.2402, 0.2752]}
					& \scriptsize{[0.2759, 0.3103]} \\
					\cmidrule{2-5}
					
					& \multirow{2}{*}{Astra}
					& 0.3572$^{**}$ & 0.3405$^{**}$ & 0.3486$^{**}$ \\
					& & \scriptsize{[0.3373, 0.3774]}
					& \scriptsize{[0.3222, 0.3578]}
					& \scriptsize{[0.3305, 0.3652]} \\
					
					\midrule
					
					\multirow{6}{*}{\textbf{Inspect}}
					& \multirow{2}{*}{Astra-Base-Ori}
					& 0.4306 & 0.4524 & 0.4412 \\
					& & \scriptsize{[0.4161, 0.4449]}
					& \scriptsize{[0.4385, 0.4667]}
					& \scriptsize{[0.4288, 0.4537]} \\
					\cmidrule{2-5}
					
					& \multirow{2}{*}{Astra-Base}
					& 0.4700$^*$ & 0.4321 & 0.4503 \\
					& & \scriptsize{[0.4542, 0.4851]}
					& \scriptsize{[0.4181, 0.4465]}
					& \scriptsize{[0.4372, 0.4628]} \\
					\cmidrule{2-5}
					
					& \multirow{2}{*}{Astra}
					& 0.4720 & 0.4759$^{**}$ & 0.4739$^{**}$ \\
					& & \scriptsize{[0.4576, 0.4867]}
					& \scriptsize{[0.4625, 0.4892]}
					& \scriptsize{[0.4616, 0.4865]} \\
					
					\midrule
					
					\multirow{6}{*}{\textbf{Merlin}}
					& \multirow{2}{*}{Astra-Base-Ori}
					& 0.4727 & 0.4313 & 0.4510 \\
					& & \scriptsize{[0.4650, 0.4808]}
					& \scriptsize{[0.4237, 0.4392]}
					& \scriptsize{[0.4443, 0.4584]} \\
					\cmidrule{2-5}
					
					& \multirow{2}{*}{Astra-Base}
					& 0.5131$^*$ & 0.4220 & 0.4631$^*$ \\
					& & \scriptsize{[0.5044, 0.5218]}
					& \scriptsize{[0.4146, 0.4299]}
					& \scriptsize{[0.4560, 0.4701]} \\
					\cmidrule{2-5}
					
					& \multirow{2}{*}{Astra}
					& 0.5300$^{**}$ & 0.4619$^{**}$ & 0.4936$^{**}$ \\
					& & \scriptsize{[0.5215, 0.5387]}
					& \scriptsize{[0.4539, 0.4699]}
					& \scriptsize{[0.4863, 0.5008]} \\
					
					\bottomrule
				\end{tabularx}%
			}
			\centering
			\caption{\textbf{Ablation study: RadBERT classification performance (micro Precision, Recall, F1) on in-distribution datasets}
				(95\% confidence intervals estimated via 2{,}000 bootstrap iterations).
				\textit{Astra}: RL tuned model; 
				\textit{Astra-Base}: sft model with preprocessed reports;
				\textit{Astra-Base-Ori}: sft model with original reports. $^*$ represents a significant improvement between Astra-Base and Astra-Base-Ori with $P < 0.05$, $^{**}$represents a significant improvement between Astra and Astra-Base with $P < 0.05$.}
			\label{tab:ablation_radbert_indistribution}
		\end{table*}

		\clearpage
		\begin{table*}[htbp]
			\resizebox{\textwidth}{!}{%
				\begin{tabular}{llcccccc}
					\toprule
					\multirow{2}{*}{\textbf{Dataset}}
					& \multirow{2}{*}{\textbf{Model}}
					& \textbf{RaTE-Score} & \textbf{F.-Degree} & \textbf{F.-Landmark} & \textbf{F.-Feature} & \textbf{F.-Impression} & \textbf{F.-Overall} \\
					& & \scriptsize{[95\% CI]} & \scriptsize{[95\% CI]} & \scriptsize{[95\% CI]} & \scriptsize{[95\% CI]} & \scriptsize{[95\% CI]} & \scriptsize{[95\% CI]} \\
					\midrule
					
					\multirow{6}{*}{\textbf{CT-Rate}}
					& \multirow{2}{*}{Astra-Base-Ori}
					& 0.2776 & 0.2466 & 0.3316 & 0.2531 & 0.2653 & 0.2742 \\
					& & \scriptsize{[0.2720, 0.2830]}
					& \scriptsize{[0.2371, 0.2565]}
					& \scriptsize{[0.3204, 0.3423]}
					& \scriptsize{[0.2428, 0.2635]}
					& \scriptsize{[0.2503, 0.2795]}
					& \scriptsize{[0.2655, 0.2827]} \\
					\cmidrule{2-8}
					
					& \multirow{2}{*}{Astra-Base}
					& 0.3343$^*$ & 0.3079$^*$ & 0.4252$^*$ & 0.3244$^*$ & 0.3755$^*$ & 0.3582$^*$ \\
					& & \scriptsize{[0.3275, 0.3413]}
					& \scriptsize{[0.2971, 0.3176]}
					& \scriptsize{[0.4145, 0.4365]}
					& \scriptsize{[0.3124, 0.3358]}
					& \scriptsize{[0.3602, 0.3900]}
					& \scriptsize{[0.3489, 0.3673]} \\
					\cmidrule{2-8}
					
					& \multirow{2}{*}{Astra}
					& 0.3510$^{**}$ & 0.3844$^{**}$ & 0.5140$^{**}$ & 0.4159$^{**}$ & 0.4457$^{**}$ & 0.4400$^{**}$ \\
					& & \scriptsize{[0.3436, 0.3580]}
					& \scriptsize{[0.3739, 0.3946]}
					& \scriptsize{[0.5033, 0.5252]}
					& \scriptsize{[0.4034, 0.4273]}
					& \scriptsize{[0.4306, 0.4601]}
					& \scriptsize{[0.4300, 0.4493]} \\
					
					\midrule
					
					\multirow{6}{*}{\textbf{BIMCV}}
					& \multirow{2}{*}{Astra-Base-Ori}
					& 0.2226 & 0.2962 & 0.2619 & 0.1911 & 0.2414 & 0.2477 \\
					& & \scriptsize{[0.2179, 0.2274]}
					& \scriptsize{[0.2856, 0.3072]}
					& \scriptsize{[0.2501, 0.2734]}
					& \scriptsize{[0.1809, 0.2016]}
					& \scriptsize{[0.2270, 0.2562]}
					& \scriptsize{[0.2387, 0.2565]} \\
					\cmidrule{2-8}
					
					& \multirow{2}{*}{Astra-Base}
					& 0.2262 & 0.2612 & 0.2575 & 0.1833 & 0.2590 & 0.2402 \\
					& & \scriptsize{[0.2204, 0.2323]}
					& \scriptsize{[0.2488, 0.2734]}
					& \scriptsize{[0.2443, 0.2698]}
					& \scriptsize{[0.1719, 0.1945]}
					& \scriptsize{[0.2425, 0.2757]}
					& \scriptsize{[0.2299, 0.2503]} \\
					\cmidrule{2-8}
					
					& \multirow{2}{*}{Astra}
					& 0.2624$^{**}$ & 0.4133$^{**}$ & 0.3783$^{**}$ & 0.2755$^{**}$ & 0.3448$^{**}$ & 0.3530$^{**}$ \\
					& & \scriptsize{[0.2567, 0.2684]}
					& \scriptsize{[0.4023, 0.4240]}
					& \scriptsize{[0.3663, 0.3897]}
					& \scriptsize{[0.2649, 0.2863]}
					& \scriptsize{[0.3294, 0.3595]}
					& \scriptsize{[0.3441, 0.3620]} \\
					
					\midrule
					
					\multirow{6}{*}{\textbf{Inspect}}
					& \multirow{2}{*}{Astra-Base-Ori}
					& 0.2904 & 0.2815 & 0.3598 & 0.3236 & 0.3243 & 0.3223 \\
					& & \scriptsize{[0.2855, 0.2953]}
					& \scriptsize{[0.2733, 0.2895]}
					& \scriptsize{[0.3512, 0.3684]}
					& \scriptsize{[0.3145, 0.3327]}
					& \scriptsize{[0.3151, 0.3332]}
					& \scriptsize{[0.3158, 0.3288]} \\
					\cmidrule{2-8}
					
					& \multirow{2}{*}{Astra-Base}
					& 0.3047$^*$ & 0.2962$^*$ & 0.3847$^*$ & 0.3432$^*$ & 0.3502$^*$ & 0.3436$^*$ \\
					& & \scriptsize{[0.2994, 0.3098]}
					& \scriptsize{[0.2880, 0.3045]}
					& \scriptsize{[0.3758, 0.3935]}
					& \scriptsize{[0.3332, 0.3531]}
					& \scriptsize{[0.3400, 0.3598]}
					& \scriptsize{[0.3365, 0.3505]} \\
					\cmidrule{2-8}
					
					& \multirow{2}{*}{Astra}
					& 0.3255$^{**}$ & 0.3564$^{**}$ & 0.4417$^{**}$ & 0.3780$^{**}$ & 0.3958$^{**}$ & 0.3930$^{**}$ \\
					& & \scriptsize{[0.3202, 0.3310]}
					& \scriptsize{[0.3481, 0.3647]}
					& \scriptsize{[0.4327, 0.4500]}
					& \scriptsize{[0.3682, 0.3880]}
					& \scriptsize{[0.3867, 0.4055]}
					& \scriptsize{[0.3854, 0.4000]} \\
					
					\midrule
					
					\multirow{6}{*}{\textbf{Merlin}}
					& \multirow{2}{*}{Astra-Base-Ori}
					& 0.3178 & 0.3000 & 0.3187 & 0.3115 & 0.3902 & 0.3301 \\
					& & \scriptsize{[0.3152, 0.3205]}
					& \scriptsize{[0.2951, 0.3048]}
					& \scriptsize{[0.3139, 0.3233]}
					& \scriptsize{[0.3072, 0.3156]}
					& \scriptsize{[0.3838, 0.3964]}
					& \scriptsize{[0.3261, 0.3337]} \\
					\cmidrule{2-8}
					
					& \multirow{2}{*}{Astra-Base}
					& 0.3442$^*$ & 0.3085 & 0.3402$^*$ & 0.3292$^*$ & 0.4262$^*$ & 0.3510$^*$ \\
					& & \scriptsize{[0.3415, 0.3468]}
					& \scriptsize{[0.3034, 0.3136]}
					& \scriptsize{[0.3356, 0.3448]}
					& \scriptsize{[0.3247, 0.3334]}
					& \scriptsize{[0.4200, 0.4325]}
					& \scriptsize{[0.3472, 0.3548]} \\
					\cmidrule{2-8}
					
					& \multirow{2}{*}{Astra}
					& 0.3544$^{**}$ & 0.3692$^{**}$ & 0.4045$^{**}$ & 0.3706$^{**}$ & 0.4639$^{**}$ & 0.4021$^{**}$ \\
					& & \scriptsize{[0.3520, 0.3569]}
					& \scriptsize{[0.3644, 0.3742]}
					& \scriptsize{[0.4000, 0.4090]}
					& \scriptsize{[0.3664, 0.3747]}
					& \scriptsize{[0.4578, 0.4699]}
					& \scriptsize{[0.3983, 0.4058]} \\
					
					\bottomrule
				\end{tabular}%
			}
			\centering
			\caption{\textbf{Ablation study: fine-grained caption metrics (RaTE-Score, FORTE) on in-distribution datasets}
				(95\% confidence intervals estimated via 2{,}000 bootstrap iterations). F.\ means FORTE metric.
				\textit{Astra}: RL tuned model; 
				\textit{Astra-Base}: sft model with preprocessed reports;
				\textit{Astra-Base-Ori}: sft model with original reports. $^*$ represents a significant improvement between Astra-Base and Astra-Base-Ori with $P < 0.05$, $^{**}$represents a significant improvement between Astra and Astra-Base with $P < 0.05$.}
			\label{tab:ablation_rate_forte_indistribution}
		\end{table*}

		\clearpage
		\begin{table*}[htbp]
			\centering
			{\renewcommand{\arraystretch}{0.82}
				\begin{adjustbox}{max width=\textwidth, max totalheight=0.88\textheight, keepaspectratio}
					\begin{tabularx}{\textwidth}{ll >{\centering\arraybackslash}X >{\centering\arraybackslash}X >{\centering\arraybackslash}X}
						\toprule
						\multirow{2}{*}{\textbf{Dataset}}
						& \multirow{2}{*}{\textbf{Model}}
						& \textbf{micro P.} & \textbf{micro R.} & \textbf{micro F1} \\
						& & \scriptsize{[95\% CI]} & \scriptsize{[95\% CI]} & \scriptsize{[95\% CI]} \\
						\midrule
						
						\multirow{6}{*}{\textbf{CTRG-Chest}}
						& \multirow{2}{*}{Astra-Base-Ori}
						& 0.5556 & 0.1228 & 0.2011 \\
						& & \scriptsize{[0.4342, 0.6812]}
						& \scriptsize{[0.0877, 0.1617]}
						& \scriptsize{[0.1479, 0.2571]} \\
						\cmidrule{2-5}
						
						& \multirow{2}{*}{Astra-Base}
						& 0.6901$^*$ & 0.1744$^*$ & 0.2784$^*$ \\
						& & \scriptsize{[0.5873, 0.7888]}
						& \scriptsize{[0.1329, 0.2188]}
						& \scriptsize{[0.2196, 0.3386]} \\
						\cmidrule{2-5}
						
						& \multirow{2}{*}{Astra}
						& 0.6618 & 0.3203$^{**}$ & 0.4317$^{**}$ \\
						& & \scriptsize{[0.5781, 0.7379]}
						& \scriptsize{[0.2684, 0.3742]}
						& \scriptsize{[0.3718, 0.4907]} \\
						
						\midrule
						
						\multirow{6}{*}{\textbf{AMOS-MM}}
						& \multirow{2}{*}{Astra-Base-Ori}
						& 0.3662 & 0.1368 & 0.1992 \\
						& & \scriptsize{[0.2914, 0.4476]}
						& \scriptsize{[0.1043, 0.1731]}
						& \scriptsize{[0.1565, 0.2461]} \\
						\cmidrule{2-5}
						
						& \multirow{2}{*}{Astra-Base}
						& 0.4151$^*$ & 0.2895$^*$ & 0.3411$^*$ \\
						& & \scriptsize{[0.3542, 0.4762]}
						& \scriptsize{[0.2400, 0.3412]}
						& \scriptsize{[0.2885, 0.3921]} \\
						\cmidrule{2-5}
						
						& \multirow{2}{*}{Astra}
						& 0.4051$^{**}$ & 0.4211$^{**}$ & 0.4129$^{**}$ \\
						& & \scriptsize{[0.3590, 0.4534]}
						& \scriptsize{[0.3658, 0.4744]}
						& \scriptsize{[0.3694, 0.4569]} \\
						
						\midrule
						
						\multirow{6}{*}{\textbf{Inhouse-Chest-1}}
						& \multirow{2}{*}{Astra-Base-Ori}
						& 0.4162 & 0.3288 & 0.3674 \\
						& & \scriptsize{[0.3882, 0.4442]}
						& \scriptsize{[0.3007, 0.3578]}
						& \scriptsize{[0.3422, 0.3913]} \\
						\cmidrule{2-5}
						
						& \multirow{2}{*}{Astra-Base}
						& 0.5158$^*$ & 0.3408$^*$ & 0.4104$^*$ \\
						& & \scriptsize{[0.4818, 0.5498]}
						& \scriptsize{[0.3155, 0.3675]}
						& \scriptsize{[0.3851, 0.4365]} \\
						\cmidrule{2-5}
						
						& \multirow{2}{*}{Astra}
						& 0.5675$^{**}$ & 0.4355$^{**}$ & 0.4928$^{**}$ \\
						& & \scriptsize{[0.5361, 0.6002]}
						& \scriptsize{[0.4094, 0.4615]}
						& \scriptsize{[0.4679, 0.5176]} \\
						
						\midrule
						
						\multirow{6}{*}{\textbf{Inhouse-Abdomen-1}}
						& \multirow{2}{*}{Astra-Base-Ori}
						& 0.1789 & 0.2092 & 0.1928 \\
						& & \scriptsize{[0.1538, 0.2050]}
						& \scriptsize{[0.1757, 0.2439]}
						& \scriptsize{[0.1653, 0.2198]} \\
						\cmidrule{2-5}
						
						& \multirow{2}{*}{Astra-Base}
						& 0.1988$^*$ & 0.3825$^*$ & 0.2616$^*$ \\
						& & \scriptsize{[0.1765, 0.2202]}
						& \scriptsize{[0.3374, 0.4261]}
						& \scriptsize{[0.2338, 0.2879]} \\
						\cmidrule{2-5}
						
						& \multirow{2}{*}{Astra}
						& 0.2453$^{**}$ & 0.5239$^{**}$ & 0.3342$^{**}$ \\
						& & \scriptsize{[0.2263, 0.2654]}
						& \scriptsize{[0.4840, 0.5626]}
						& \scriptsize{[0.3109, 0.3575]} \\
						
						\midrule
						
						\multirow{6}{*}{\textbf{Inhouse-Abdomen-2}}
						& \multirow{2}{*}{Astra-Base-Ori}
						& 0.3516 & 0.3147 & 0.3321 \\
						& & \scriptsize{[0.3044, 0.4022]}
						& \scriptsize{[0.2657, 0.3643]}
						& \scriptsize{[0.2853, 0.3789]} \\
						\cmidrule{2-5}
						
						& \multirow{2}{*}{Astra-Base}
						& 0.2818 & 0.5268$^*$ & 0.3672$^*$ \\
						& & \scriptsize{[0.2528, 0.3102]}
						& \scriptsize{[0.4760, 0.5743]}
						& \scriptsize{[0.3321, 0.4000]} \\
						\cmidrule{2-5}
						
						& \multirow{2}{*}{Astra}
						& 0.3659$^{**}$ & 0.7599$^{**}$ & 0.4939$^{**}$ \\
						& & \scriptsize{[0.3437, 0.3885]}
						& \scriptsize{[0.7204, 0.7995]}
						& \scriptsize{[0.4706, 0.5167]} \\
						
						\midrule
						
						\multirow{6}{*}{\textbf{Inhouse-Abdomen-3}}
						& \multirow{2}{*}{Astra-Base-Ori}
						& 0.3521 & 0.3055 & 0.3271 \\
						& & \scriptsize{[0.3191, 0.3884]}
						& \scriptsize{[0.2752, 0.3358]}
						& \scriptsize{[0.2989, 0.3571]} \\
						\cmidrule{2-5}
						
						& \multirow{2}{*}{Astra-Base}
						& 0.3054 & 0.3979$^*$ & 0.3456$^*$ \\
						& & \scriptsize{[0.2802, 0.3318]}
						& \scriptsize{[0.3617, 0.4335]}
						& \scriptsize{[0.3180, 0.3729]} \\
						\cmidrule{2-5}
						
						& \multirow{2}{*}{Astra}
						& 0.3811$^{**}$ & 0.6046$^{**}$ & 0.4675$^{**}$ \\
						& & \scriptsize{[0.3576, 0.4041]}
						& \scriptsize{[0.5736, 0.6377]}
						& \scriptsize{[0.4443, 0.4898]} \\
						
						\bottomrule
					\end{tabularx}
			\end{adjustbox}}
			\caption{\textbf{Ablation study: RadBERT classification performance (micro Precision, Recall, F1) on external validation datasets}
				(95\% confidence intervals estimated via 2{,}000 bootstrap iterations). P. means precision. R. means recall.
				\textit{Astra}: RL tuned model;
				\textit{Astra-Base}: SFT model with preprocessed reports;
				\textit{Astra-Base-Ori}: SFT model with original reports.
				$^*$ represents a significant improvement between Astra-Base and Astra-Base-Ori with $P < 0.05$;
				$^{**}$ represents a significant improvement between Astra and Astra-Base with $P < 0.05$.}
			\label{tab:ablation_radbert_external}
		\end{table*}

		\clearpage
		\begin{table*}[htbp]
			{\renewcommand{\arraystretch}{0.82}
				\begin{adjustbox}{max width=\textwidth, max totalheight=0.88\textheight, keepaspectratio}
					\begin{tabular}{llcccccc}
						\toprule
						\multirow{2}{*}{\textbf{Dataset}}
						& \multirow{2}{*}{\textbf{Model}}
						& \textbf{RaTE-Score} & \textbf{F.-Degree} & \textbf{F.-Landmark} & \textbf{F.-Feature} & \textbf{F.-Impression} & \textbf{F.-Overall} \\
						& & \scriptsize{[95\% CI]} & \scriptsize{[95\% CI]} & \scriptsize{[95\% CI]} & \scriptsize{[95\% CI]} & \scriptsize{[95\% CI]} & \scriptsize{[95\% CI]} \\
						\midrule
						
						\multirow{6}{*}{\textbf{CTRG-Chest}}
						& \multirow{2}{*}{Astra-Base-Ori}
						& 0.1991 & 0.1273 & 0.2160 & 0.0769 & 0.0685 & 0.1222 \\
						& & \scriptsize{[0.1922, 0.2065]}
						& \scriptsize{[0.1082, 0.1473]}
						& \scriptsize{[0.1931, 0.2394]}
						& \scriptsize{[0.0577, 0.0965]}
						& \scriptsize{[0.0438, 0.0936]}
						& \scriptsize{[0.1061, 0.1389]} \\
						\cmidrule{2-8}
						
						& \multirow{2}{*}{Astra-Base}
						& 0.1872 & 0.1169 & 0.1614 & 0.0608 & 0.0939$^*$ & 0.1083 \\
						& & \scriptsize{[0.1791, 0.1956]}
						& \scriptsize{[0.0933, 0.1407]}
						& \scriptsize{[0.1320, 0.1906]}
						& \scriptsize{[0.0414, 0.0803]}
						& \scriptsize{[0.0586, 0.1300]}
						& \scriptsize{[0.0879, 0.1294]} \\
						\cmidrule{2-8}
						
						& \multirow{2}{*}{Astra}
						& 0.2434$^{**}$ & 0.2550$^{**}$ & 0.3918$^{**}$ & 0.1408$^{**}$ & 0.1346$^{**}$ & 0.2306$^{**}$ \\
						& & \scriptsize{[0.2348, 0.2525]}
						& \scriptsize{[0.2324, 0.2783]}
						& \scriptsize{[0.3690, 0.4131]}
						& \scriptsize{[0.1195, 0.1634]}
						& \scriptsize{[0.1063, 0.1649]}
						& \scriptsize{[0.2151, 0.2467]} \\
						
						\midrule
						
						\multirow{6}{*}{\textbf{AMOS-MM}}
						& \multirow{2}{*}{Astra-Base-Ori}
						& 0.2025 & 0.1220 & 0.1507 & 0.1196 & 0.0516 & 0.1110 \\
						& & \scriptsize{[0.1964, 0.2087]}
						& \scriptsize{[0.1030, 0.1406]}
						& \scriptsize{[0.1331, 0.1684]}
						& \scriptsize{[0.1030, 0.1375]}
						& \scriptsize{[0.0326, 0.0729]}
						& \scriptsize{[0.0979, 0.1238]} \\
						\cmidrule{2-8}
						
						& \multirow{2}{*}{Astra-Base}
						& 0.2389$^*$ & 0.2480$^*$ & 0.2602$^*$ & 0.1710$^*$ & 0.0913$^*$ & 0.1926$^*$ \\
						& & \scriptsize{[0.2335, 0.2443]}
						& \scriptsize{[0.2266, 0.2704]}
						& \scriptsize{[0.2410, 0.2786]}
						& \scriptsize{[0.1547, 0.1883]}
						& \scriptsize{[0.0730, 0.1107]}
						& \scriptsize{[0.1798, 0.2060]} \\
						\cmidrule{2-8}
						
						& \multirow{2}{*}{Astra}
						& 0.2507$^{**}$ & 0.2880$^{**}$ & 0.3084$^{**}$ & 0.1973$^{**}$ & 0.1036$^{**}$ & 0.2243$^{**}$ \\
						& & \scriptsize{[0.2449, 0.2561]}
						& \scriptsize{[0.2684, 0.3070]}
						& \scriptsize{[0.2904, 0.3258]}
						& \scriptsize{[0.1807, 0.2138]}
						& \scriptsize{[0.0826, 0.1258]}
						& \scriptsize{[0.2117, 0.2370]} \\
						
						\midrule
						
						\multirow{6}{*}{\textbf{Inhouse-Chest-1}}
						& \multirow{2}{*}{Astra-Base-Ori}
						& 0.2484 & 0.2143 & 0.3545 & 0.1938 & 0.1888 & 0.2379 \\
						& & \scriptsize{[0.2410, 0.2564]}
						& \scriptsize{[0.1992, 0.2295]}
						& \scriptsize{[0.3338, 0.3752]}
						& \scriptsize{[0.1786, 0.2089]}
						& \scriptsize{[0.1611, 0.2162]}
						& \scriptsize{[0.2233, 0.2521]} \\
						\cmidrule{2-8}
						
						& \multirow{2}{*}{Astra-Base}
						& 0.2533$^*$ & 0.2029 & 0.3168 & 0.2136$^*$ & 0.2704$^*$ & 0.2509$^*$ \\
						& & \scriptsize{[0.2459, 0.2609]}
						& \scriptsize{[0.1869, 0.2191]}
						& \scriptsize{[0.2951, 0.3373]}
						& \scriptsize{[0.1957, 0.2308]}
						& \scriptsize{[0.2365, 0.3039]}
						& \scriptsize{[0.2359, 0.2663]} \\
						\cmidrule{2-8}
						
						& \multirow{2}{*}{Astra}
						& 0.2874$^{**}$ & 0.2745$^{**}$ & 0.4805$^{**}$ & 0.2641 & 0.3227$^{**}$ & 0.3355$^{**}$ \\
						& & \scriptsize{[0.2792, 0.2957]}
						& \scriptsize{[0.2605, 0.2890]}
						& \scriptsize{[0.4647, 0.4957]}
						& \scriptsize{[0.2489, 0.2788]}
						& \scriptsize{[0.2888, 0.3562]}
						& \scriptsize{[0.3231, 0.3473]} \\
						
						\midrule
						
						\multirow{6}{*}{\textbf{Inhouse-Abd.-1}}
						& \multirow{2}{*}{Astra-Base-Ori}
						& 0.1929 & 0.1588 & 0.1796 & 0.1160 & 0.0396 & 0.1235 \\
						& & \scriptsize{[0.1871, 0.1991]}
						& \scriptsize{[0.1408, 0.1762]}
						& \scriptsize{[0.1620, 0.1990]}
						& \scriptsize{[0.0993, 0.1325]}
						& \scriptsize{[0.0252, 0.0557]}
						& \scriptsize{[0.1121, 0.1353]} \\
						\cmidrule{2-8}
						
						& \multirow{2}{*}{Astra-Base}
						& 0.2343$^*$ & 0.2420$^*$ & 0.2609$^*$ & 0.1639$^*$ & 0.0534$^*$ & 0.1800$^*$ \\
						& & \scriptsize{[0.2290, 0.2401]}
						& \scriptsize{[0.2240, 0.2602]}
						& \scriptsize{[0.2427, 0.2799]}
						& \scriptsize{[0.1488, 0.1798]}
						& \scriptsize{[0.0382, 0.0697]}
						& \scriptsize{[0.1696, 0.1906]} \\
						\cmidrule{2-8}
						
						& \multirow{2}{*}{Astra}
						& 0.2509$^{**}$ & 0.3029$^{**}$ & 0.3500$^{**}$ & 0.2169$^{**}$ & 0.0631$^{**}$ & 0.2332$^{**}$ \\
						& & \scriptsize{[0.2452, 0.2564]}
						& \scriptsize{[0.2883, 0.3185]}
						& \scriptsize{[0.3345, 0.3664]}
						& \scriptsize{[0.2020, 0.2308]}
						& \scriptsize{[0.0471, 0.0807]}
						& \scriptsize{[0.2243, 0.2422]} \\
						
						\midrule
						
						\multirow{6}{*}{\textbf{Inhouse-Abd.-2}}
						& \multirow{2}{*}{Astra-Base-Ori}
						& 0.2437 & 0.1563 & 0.2876 & 0.2172 & 0.1894 & 0.2126 \\
						& & \scriptsize{[0.2339, 0.2540]}
						& \scriptsize{[0.1329, 0.1780]}
						& \scriptsize{[0.2634, 0.3114]}
						& \scriptsize{[0.1911, 0.2418]}
						& \scriptsize{[0.1480, 0.2309]}
						& \scriptsize{[0.1909, 0.2341]} \\
						\cmidrule{2-8}
						
						& \multirow{2}{*}{Astra-Base}
						& 0.2765$^*$ & 0.3309$^*$ & 0.3890$^*$ & 0.2979$^*$ & 0.2850$^*$ & 0.3257$^*$ \\
						& & \scriptsize{[0.2700, 0.2832]}
						& \scriptsize{[0.3099, 0.3513]}
						& \scriptsize{[0.3706, 0.4085]}
						& \scriptsize{[0.2789, 0.3176]}
						& \scriptsize{[0.2582, 0.3136]}
						& \scriptsize{[0.3126, 0.3389]} \\
						\cmidrule{2-8}
						
						& \multirow{2}{*}{Astra}
						& 0.3157$^{**}$ & 0.4701$^{**}$ & 0.5270$^{**}$ & 0.4234$^{**}$ & 0.3871$^{**}$ & 0.4519$^{**}$ \\
						& & \scriptsize{[0.3097, 0.3216]}
						& \scriptsize{[0.4538, 0.4869]}
						& \scriptsize{[0.5125, 0.5415]}
						& \scriptsize{[0.4078, 0.4403]}
						& \scriptsize{[0.3562, 0.4175]}
						& \scriptsize{[0.4426, 0.4613]} \\
						
						\midrule
						
						\multirow{6}{*}{\textbf{Inhouse-Abd.-3}}
						& \multirow{2}{*}{Astra-Base-Ori}
						& 0.2364 & 0.1755 & 0.2263 & 0.1939 & 0.1883 & 0.1960 \\
						& & \scriptsize{[0.2298, 0.2428]}
						& \scriptsize{[0.1559, 0.1949]}
						& \scriptsize{[0.2093, 0.2431]}
						& \scriptsize{[0.1766, 0.2099]}
						& \scriptsize{[0.1585, 0.2176]}
						& \scriptsize{[0.1814, 0.2101]} \\
						\cmidrule{2-8}
						
						& \multirow{2}{*}{Astra-Base}
						& 0.2843$^*$ & 0.2801$^*$ & 0.3107$^*$ & 0.2305$^*$ & 0.2133$^*$ & 0.2587$^*$ \\
						& & \scriptsize{[0.2769, 0.2915]}
						& \scriptsize{[0.2647, 0.2956]}
						& \scriptsize{[0.2946, 0.3250]}
						& \scriptsize{[0.2157, 0.2446]}
						& \scriptsize{[0.1861, 0.2381]}
						& \scriptsize{[0.2468, 0.2698]} \\
						\cmidrule{2-8}
						
						& \multirow{2}{*}{Astra}
						& 0.3143$^{**}$ & 0.4053$^{**}$ & 0.4512$^{**}$ & 0.3215$^{**}$ & 0.2640$^{**}$ & 0.3605$^{**}$ \\
						& & \scriptsize{[0.3077, 0.3216]}
						& \scriptsize{[0.3927, 0.4182]}
						& \scriptsize{[0.4376, 0.4650]}
						& \scriptsize{[0.3069, 0.3353]}
						& \scriptsize{[0.2352, 0.2916]}
						& \scriptsize{[0.3506, 0.3704]} \\
						
						\bottomrule
					\end{tabular}
			\end{adjustbox}}
			\centering
			\caption{\textbf{Ablation study: fine-grained caption metrics (RaTE-Score, FORTE) on external validation datasets}
				(95\% confidence intervals estimated via 2{,}000 bootstrap iterations). F.\ means FORTE metric.
				\textit{Astra}: RL tuned model; 
				\textit{Astra-Base}: sft model with preprocessed reports;
				\textit{Astra-Base-Ori}: sft model with original reports. $^*$ represents a significant improvement between Astra-Base and Astra-Base-Ori with $P < 0.05$, $^{**}$represents a significant improvement between Astra and Astra-Base with $P < 0.05$.}
			\label{tab:ablation_rate_forte_indistribution}
		\end{table*}

		\newpage
		\begin{table*}[t]
			\centering
			\footnotesize
			\begin{tabular}{p{1.0\textwidth}}
				\toprule
				\textbf{Human evaluation criteria for radiology report quality} \\
				\\
				For each report, the evaluator should assign a score for each of the three dimensions below and provide a brief rationale explaining the assigned score. \\
				\\
				\textbf{Readability:} Writing quality of the report in terms of grammatical correctness, logical coherence and clarity of terminology. \\
				\textbf{10--9:} \textit{Highly fluent.} No spelling or grammatical errors; appropriate ordering of descriptions; clear and coherent logic; easy to read. \\
				\textbf{8--7:} \textit{Generally fluent.} No more than one spelling or grammatical error; minor issues in the order of description may be present, but readability is not affected. \\
				\textbf{6--5:} \textit{Moderately fluent.} No more than three spelling or grammatical errors; some issues in the order of description may be present, but the report remains understandable. \\
				\textbf{4--3:} \textit{Poorly fluent.} More than three spelling or grammatical errors; obvious logical problems that make the report difficult to read. \\
				\textbf{2--1:} \textit{Extremely poor fluency.} More than five spelling or grammatical errors; major logical problems that severely compromise readability. \\
				\\
				\textbf{Completeness:} Whether the report contains all necessary information and observations. \\
				\textbf{10--9:} \textit{Highly complete.} All lesions and clinically relevant findings are included. \\
				\textbf{8--7:} \textit{Mostly complete.} All major lesions are included; no more than two minor lesions are omitted; the missing information does not affect clinical diagnosis or decision-making. \\
				\textbf{6--5:} \textit{Moderately complete.} All major lesions are included; more than two minor lesions or negative findings are omitted; the missing information may reduce diagnostic thoroughness but does not fundamentally change the clinical interpretation. \\
				\textbf{4--3:} \textit{Incomplete.} Important lesions are omitted; some minor lesions or negative findings may still be identified; the missing findings are clinically meaningful and may affect diagnosis or treatment decisions. \\
				\textbf{2--1:} \textit{Severely incomplete.} The report fails to identify key abnormalities and provides only general negative statements or highly nonspecific descriptions; omission of critical findings may lead to serious diagnostic error. \\
				\\
				\textbf{Accuracy:} The extent to which the report content is consistent with the patient’s actual condition as shown on CT imaging. \\
				\textbf{10--9:} \textit{Highly accurate.} The location, morphology and imaging characteristics of all lesions are described correctly. \\
				\textbf{8--7:} \textit{Generally accurate.} Major lesions are described accurately; minor lesions may contain slight inaccuracies, but these do not affect clinical diagnosis. \\
				\textbf{6--5:} \textit{Moderately accurate.} Descriptions of major lesions contain minor inaccuracies, but these do not affect the preliminary clinical diagnosis. \\
				\textbf{4--3:} \textit{Inaccurate.} Significant inaccuracies are present in the description of minor lesions; these errors may be clinically relevant. \\
				\textbf{2--1:} \textit{Completely inaccurate.} Significant inaccuracies are present in the description of major lesions; these errors are clinically relevant and may lead to misleading or incorrect clinical interpretation. \\
				\\
				\textbf{Scoring instruction:} For each dimension, the evaluator must provide both a numerical score and a brief justification explaining why the report received that score. \\
				\bottomrule
			\end{tabular}
			\caption{Scoring criteria used for human evaluation of radiology report quality in the human--AI collaboration study.}
			\label{tab:human_eval_criteria}
		\end{table*}
		
		\clearpage
		\begin{table*}[htbp]
			\centering
			{\renewcommand{\arraystretch}{0.78}
				\begin{adjustbox}{max width=\textwidth, max totalheight=0.88\textheight, keepaspectratio}
					\begin{tabularx}{\textwidth}{l >{\centering\arraybackslash}X >{\centering\arraybackslash}X >{\centering\arraybackslash}X >{\centering\arraybackslash}X >{\centering\arraybackslash}X >{\centering\arraybackslash}X}
						\toprule
						& \multicolumn{6}{c}{\textbf{Reward Function}} \\
						\cmidrule{2-7}
						\textbf{Metric}
						& \textbf{Base} & \textbf{NLG} & \textbf{RadBERT} & \textbf{Rate-S.} & \textbf{\mbox{FORTE-O.}} & \textbf{FORTE} \\
						\midrule
						
						\multirow{2}{*}{BLEU-1}
						& 0.4058 & 0.4993 & 0.4559 & 0.3314 & 0.4036 & 0.4923 \\
						& \scriptsize{[0.3936, 0.4179]} & \scriptsize{[0.4894, 0.5089]} & \scriptsize{[0.4462, 0.4646]} & \scriptsize{[0.3231, 0.3392]} & \scriptsize{[0.3913, 0.4154]} & \scriptsize{[0.4838, 0.5002]} \\
						\thinhline
						
						\multirow{2}{*}{BLEU-2}
						& 0.3140 & 0.3949 & 0.3450 & 0.2505 & 0.3177 & 0.3793 \\
						& \scriptsize{[0.3041, 0.3239]} & \scriptsize{[0.3868, 0.4025]} & \scriptsize{[0.3373, 0.3522]} & \scriptsize{[0.2444, 0.2564]} & \scriptsize{[0.3076, 0.3276]} & \scriptsize{[0.3722, 0.3860]} \\
						\thinhline
						
						\multirow{2}{*}{BLEU-3}
						& 0.2491 & 0.3178 & 0.2691 & 0.1944 & 0.2558 & 0.2965 \\
						& \scriptsize{[0.2406, 0.2575]} & \scriptsize{[0.3106, 0.3245]} & \scriptsize{[0.2626, 0.2755]} & \scriptsize{[0.1896, 0.1990]} & \scriptsize{[0.2472, 0.2642]} & \scriptsize{[0.2900, 0.3028]} \\
						\thinhline
						
						\multirow{2}{*}{BLEU-4}
						& 0.2027 & 0.2604 & 0.2150 & 0.1542 & 0.2102 & 0.2373 \\
						& \scriptsize{[0.1953, 0.2103]} & \scriptsize{[0.2537, 0.2668]} & \scriptsize{[0.2090, 0.2208]} & \scriptsize{[0.1501, 0.1580]} & \scriptsize{[0.2026, 0.2178]} & \scriptsize{[0.2311, 0.2434]} \\
						\thinhline
						
						\multirow{2}{*}{METEOR}
						& 0.2161 & 0.2494 & 0.2200 & 0.2372 & 0.2193 & 0.2373 \\
						& \scriptsize{[0.2121, 0.2204]} & \scriptsize{[0.2459, 0.2529]} & \scriptsize{[0.2162, 0.2237]} & \scriptsize{[0.2344, 0.2401]} & \scriptsize{[0.2156, 0.2231]} & \scriptsize{[0.2338, 0.2407]} \\
						\thinhline
						
						\multirow{2}{*}{ROUGE-L}
						& 0.4156 & 0.4499 & 0.3922 & 0.3550 & 0.4210 & 0.4165 \\
						& \scriptsize{[0.4077, 0.4232]} & \scriptsize{[0.4423, 0.4574]} & \scriptsize{[0.3852, 0.3991]} & \scriptsize{[0.3479, 0.3624]} & \scriptsize{[0.4128, 0.4295]} & \scriptsize{[0.4096, 0.4234]} \\
						
						\midrule
						
						\multirow{2}{*}{RadBERT-P.}
						& 0.5658 & 0.5504 & 0.5565 & 0.5301 & 0.5913 & 0.5570 \\
						& \scriptsize{[0.5498, 0.5835]} & \scriptsize{[0.5351, 0.5660]} & \scriptsize{[0.5410, 0.5718]} & \scriptsize{[0.5155, 0.5452]} & \scriptsize{[0.5752, 0.6088]} & \scriptsize{[0.5423, 0.5725]} \\
						\thinhline
						
						\multirow{2}{*}{RadBERT-R.}
						& 0.4635 & 0.5125 & 0.5377 & 0.5453 & 0.4308 & 0.5397 \\
						& \scriptsize{[0.4502, 0.4781]} & \scriptsize{[0.4991, 0.5275]} & \scriptsize{[0.5237, 0.5524]} & \scriptsize{[0.5310, 0.5597]} & \scriptsize{[0.4184, 0.4436]} & \scriptsize{[0.5252, 0.5544]} \\
						\thinhline
						
						\multirow{2}{*}{RadBERT-F1}
						& 0.5096 & 0.5308 & 0.5470 & 0.5376 & 0.4985 & 0.5482 \\
						& \scriptsize{[0.4971, 0.5222]} & \scriptsize{[0.5186, 0.5428]} & \scriptsize{[0.5353, 0.5591]} & \scriptsize{[0.5253, 0.5501]} & \scriptsize{[0.4865, 0.5107]} & \scriptsize{[0.5360, 0.5611]} \\
						
						\midrule
						
						\multirow{2}{*}{RaTE-Score}
						& 0.3891 & 0.4069 & 0.3868 & 0.4333 & 0.4052 & 0.4121 \\
						& \scriptsize{[0.3825, 0.3960]} & \scriptsize{[0.4000, 0.4138]} & \scriptsize{[0.3800, 0.3934]} & \scriptsize{[0.4260, 0.4404]} & \scriptsize{[0.3980, 0.4122]} & \scriptsize{[0.4052, 0.4190]} \\
						
						\midrule
						
						\multirow{2}{*}{F.-Degree}
						& 0.3898 & 0.4245 & 0.3671 & 0.3988 & 0.4228 & 0.4549 \\
						& \scriptsize{[0.3816, 0.3984]} & \scriptsize{[0.4165, 0.4324]} & \scriptsize{[0.3591, 0.3759]} & \scriptsize{[0.3907, 0.4065]} & \scriptsize{[0.4150, 0.4311]} & \scriptsize{[0.4467, 0.4631]} \\
						\thinhline
						
						\multirow{2}{*}{F.-Landmark}
						& 0.4958 & 0.5434 & 0.4810 & 0.5456 & 0.5226 & 0.5597 \\
						& \scriptsize{[0.4868, 0.5048]} & \scriptsize{[0.5344, 0.5521]} & \scriptsize{[0.4713, 0.4909]} & \scriptsize{[0.5364, 0.5547]} & \scriptsize{[0.5137, 0.5315]} & \scriptsize{[0.5510, 0.5685]} \\
						\thinhline
						
						\multirow{2}{*}{F.-Feature}
						& 0.3403 & 0.4042 & 0.3583 & 0.4062 & 0.3678 & 0.4154 \\
						& \scriptsize{[0.3286, 0.3521]} & \scriptsize{[0.3929, 0.4153]} & \scriptsize{[0.3464, 0.3700]} & \scriptsize{[0.3950, 0.4164]} & \scriptsize{[0.3564, 0.3792]} & \scriptsize{[0.4037, 0.4262]} \\
						\thinhline
						
						\multirow{2}{*}{F.-Impression}
						& 0.3833 & 0.4243 & 0.4013 & 0.4211 & 0.4170 & 0.4400 \\
						& \scriptsize{[0.3687, 0.3986]} & \scriptsize{[0.4108, 0.4390]} & \scriptsize{[0.3871, 0.4156]} & \scriptsize{[0.4075, 0.4343]} & \scriptsize{[0.4021, 0.4314]} & \scriptsize{[0.4258, 0.4544]} \\
						\thinhline
						
						\multirow{2}{*}{F.-Overall}
						& 0.4023 & 0.4491 & 0.4020 & 0.4429 & 0.4326 & 0.4675 \\
						& \scriptsize{[0.3941, 0.4109]} & \scriptsize{[0.4407, 0.4574]} & \scriptsize{[0.3931, 0.4108]} & \scriptsize{[0.4346, 0.4506]} & \scriptsize{[0.4243, 0.4406]} & \scriptsize{[0.4589, 0.4754]} \\
						
						\bottomrule
					\end{tabularx}
			\end{adjustbox}}
			\caption{\textbf{Ablation study on GRPO reward function} ($n = 1{,}564$ cases on CT-Rate; 95\% confidence intervals estimated via 2{,}000 bootstrap iterations).
				\textit{Base}: SFT baseline without RL;
				\textit{NLG}: RL with NLG reward only;
				\textit{RadBERT}: RL with RadBERT reward only;
				\textit{Rate-Score}: RL with RaTE-Score reward only;
				\textit{FORTE-Ori}: RL with original FORTE reward;
				\textit{FORTE}: RL with full FORTE reward (Astra).
				F.\ means FORTE metric.}
			\label{tab:ablation_grpo_reward}
		\end{table*}

		\begin{table}[htbp]
			\begin{tabular}{@{} l c c c l l@{}}
				\toprule
				Dataset & Train & Val & Test & Tasks & Modality \\
				\midrule
				CT-Rate$^*$        & 24128 & -- & 1564 & report generation, classification & Non-contrast chest CT  \\
				Merlin$^*$         & 15211 & 5025 & 5105 & report generation, classification & Contrast-enhanced abdominal CT \\
				Atlas3.0$^*$       & 8019  & -- & 883  & report generation & Multi-phase abdominal CT \\
				Inspect$^*$        & 18925 & 1088 & 3212 & report generation & CT pulmonary angiography \\
				BIMCV$^*$        & 5262  & 751  & 1505 & report generation & Non-contrast chest CT \\
				CTRG-chest     & 1480  & --  & 324  & report generation & Non-contrast chest CT \\
				AMOS-MM        & 1287  & --  & 400  & report generation & Contrast-enhanced abdominal CT \\
				Inhouse-Chest  & --    & --  & 400  & report generation & Non-contrast chest CT \\
				Inhouse-Abdomen-1 & -- & -- & 400 & report generation & Contrast-enhanced abdominal CT\\
				Inhouse-Abdomen-2 & -- & -- & 400 & report generation & Multi-phase abdominal CT\\
				Inhouse-Abdomen-3 & -- & -- & 400 & report generation & Multi-phase abdominal CT \\
				Radchest       & 2287  & 985  & 361  & classification & Non-contrast chest CT \\
				RSNA-PE        & 5823  & --  & 1456 & classification & CT pulmonary angiography\\
				NLST           & 84839 & --  & --  & pretrain & Low-dose chest CT\\
				\bottomrule
			\end{tabular}
			\centering
			\caption{Overview of the datasets used for Astra development, ensemble strategy, and scaling pretraining. $^*$ datasets comprised CTRgDB and were utilized for the training of Astra. Other report generation datasets were used for external validation. The NLST dataset served as an unpaired report dataset for scaling vision-language pretraining. CT-Rate, Merlin, Radchest, and RSNA-PE were used to evaluate the ensemble strategy and scaling pretraining performance.}
		\end{table}

		\newpage
		\begin{table*}[t]
			\centering
			\footnotesize
			\begin{tabular}{p{1.0\textwidth}}
				\toprule
				\textbf{Structured Abdominal CT Report Extraction Prompt:} \\
				\textbf{System Message:} You are an expert radiology report processor specializing in extracting structured information from abdomen CT reports. \\
				\textbf{User Message:} Your task is to analyze a given abdomen CT report, apply a set of specific rules, and output the results as a single, valid JSON object. \\
				\\
				\textbf{Target Anatomical Regions:} \\
				You must extract information for the following 13 specific anatomical regions ONLY: \\
				1. lower thorax: includes lower chest and lung bases \\
				2. liver and biliary tree: includes liver and biliary tree (bile ducts), but NOT gallbladder \\
				3. gallbladder: gallbladder only (separate from liver) \\
				4. spleen: spleen only \\
				5. pancreas: pancreas only \\
				6. adrenal glands: adrenal glands only \\
				7. kidneys and ureters: includes kidneys and ureters \\
				8. gastrointestinal tract: includes stomach, intestines, appendix; be careful not to miss the appendix \\
				9. peritoneum: includes peritoneal space, peritoneal cavity, and abdominal wall \\
				10. pelvic: includes pelvic organs, bladder, prostate and seminal vesicles, uterus and ovaries \\
				11. vasculature: vasculature system \\
				12. lymph nodes: lymph nodes \\
				13. musculoskeletal: includes bones and muscles \\
				\\
				\textbf{Rule 1: Remove Non-Diagnostic Text} \\
				Before any other processing, you must identify and completely remove any text that represents communication between clinicians, summary codes, or procedural notes. This includes but is not limited to phrases like ``FINDINGS GIVEN TO Dr. ...'', ``SUMMARY 4:...'', ``END OF IMPRESSION:'', and ``AT THE CONCLUSION OF THE EXAMINATION...''. \\
				\\
				\textbf{Rule 2: Process Each Target Region} \\
				For each of the 13 target regions listed above, find the corresponding section in the report and apply these sub-rules: \\
				A. \textbf{Delete Negative Findings:} Remove all sentences, clauses, or phrases that describe normal, negative, or absent findings. Examples include ``The liver is normal in size and shape'', ``with coordinated proportion of liver lobes'', ``unremarkable appearance'', and ``without evidence of...''. \\
				B. \textbf{Rewrite Comparative Statements:} If a finding is compared to a prior scan (for example, ``stable'', ``unchanged'', or ``increased/decreased in size''), rewrite the sentence to describe only the current state of the finding. For example, ``The 2 cm nodule is stable'' becomes ``There is a 2 cm nodule.'' If a finding has resolved, it is now normal and should be removed according to Rule 2A. \\
				C. \textbf{Preserve Positive Findings:} For general non-comparative positive findings, the original statements should be retained in their entirety \textbf{without any modifications}. \\
				\\
				\textbf{Rule 3: Generate Structured JSON Output} \\
				The final output MUST be a single, valid JSON object and nothing else. Do not include any introductory text, explanations, or markdown code block markers. \\
				The JSON object must have keys corresponding to ALL 13 predefined anatomical regions. \\
				For each region: \\
				1. If there are any positive findings remaining after processing, the value should be a string containing those findings. \\
				2. If a region had no positive findings (i.e., it was normal, all findings were negative, or the section was not mentioned in the report), the value MUST be the string ``normal''. \\
				\\
				\textbf{Report:} \{report\} \\
				\textbf{Processed Report (JSON):} \dots \\
				\bottomrule
			\end{tabular}
			\caption{Prompt used for structured information extraction from abdominal CT reports.}
			\label{tab:abdomen_ct_prompt}
		\end{table*}
		
		\newpage
		\begin{table*}[t]
			\centering
			\footnotesize
			\begin{tabular}{p{1.0\textwidth}}
				\toprule
				\textbf{Structured Chest CT Report Extraction Prompt:} \\
				\textbf{System Message:} You are an expert radiology report processor specializing in extracting structured information from chest CT reports. \\
				\textbf{User Message:} Your task is to analyze a given chest CT report, apply a set of specific rules, and output the results as a single, valid JSON object. \\
				\\
				\textbf{Target Anatomical Regions:} \\
				You must extract information for the following 10 specific anatomical regions ONLY: \\
				1. abdomen: include all findings related to liver, gallbladder, pancreas, spleen, kidneys, adrenals, gastrointestinal tract, abdominal vessels, and abdominal lymph nodes, etc. \\
				2. bone: bone \\
				3. breast: breast \\
				4. esophagus: esophagus \\
				5. heart: heart \\
				6. lung: lung \\
				7. mediastinum: mediastinum area \\
				8. pleura: pleura \\
				9. thyroid: thyroid \\
				10. trachea and bronchi: trachea and bronchi \\
				\\
				\textbf{Rule 1: Remove Non-Diagnostic Text} \\
				Before any other processing, you must identify and completely remove any text that represents communication between clinicians, summary codes, or procedural notes. This includes but is not limited to phrases like ``FINDINGS GIVEN TO Dr. ...'', ``SUMMARY 4:...'', ``END OF IMPRESSION:'', and ``AT THE CONCLUSION OF THE EXAMINATION...'', etc. \\
				\\
				\textbf{Rule 2: Process Each Primary Target Region} \\
				For each of the 10 primary target regions listed above, find the corresponding section in the report and apply these sub-rules: \\
				A. \textbf{Delete Negative Findings:} Remove all sentences, clauses, or phrases that describe normal, negative, or absent findings. Examples include ``The lungs are clear'', ``No pleural effusion is seen'', ``unremarkable appearance'', and ``without evidence of...''. \\
				B. \textbf{Rewrite Comparative Statements:} If a finding is compared to a prior scan (for example, ``stable'', ``unchanged'', or ``increased/decreased in size''), rewrite the sentence to describe only the current state of the finding. For example, ``The 2 cm nodule is stable'' becomes ``There is a 2 cm nodule.'' If a finding has resolved, it is now normal and should be removed according to Rule 2A. \\
				C. \textbf{Preserve Positive Findings:} For general non-comparative positive findings, the original statements should be retained in their entirety \textbf{without any modifications}. \\
				\\
				\textbf{Rule 5: Generate Structured JSON Output} \\
				The final output MUST be a single, valid JSON object and nothing else. Do not include any introductory text, explanations, or markdown code block markers. \\
				The JSON object must have keys corresponding to ALL 10 primary anatomical regions. If Rule 3 is triggered, an 11th key, ``brain'', must also be included. \\
				For each region: \\
				1. If there are any positive findings remaining after processing, the value should be a string containing those findings. \\
				2. If a region had no positive findings (i.e., it was normal, all findings were negative, or the section was not mentioned in the report), the value MUST be the string ``normal''. \\
				\textbf{Report:} \texttt{\{report\}} \\
				\textbf{Processed Report (JSON):} \dots \\
				\bottomrule
			\end{tabular}
			\caption{Prompt used for structured information extraction from chest CT reports.}
			\label{tab:chest_ct_prompt}
		\end{table*}
		
		\newpage
		\begin{table*}[t]
			\centering
			\footnotesize
			\begin{tabular}{p{1.0\textwidth}}
				\toprule
				\textbf{Abdominal CT Example Case:} \\
				\textbf{Input Report:} ``Abdomen: The liver is normal in size and shape, with coordinated proportion of liver lobes. Multiple mixed high- and low-density lesions are found in the liver, with CT values of approximately 17--186 HU. The largest lesion is in the posterior right lobe of the liver measuring approximately 69mm x 88mm. On the enhancement scan, no significant enhancement is seen in the lesions, with a low-density ring seen surrounding the lesions. Patchy slightly low-density foci are seen in the left lobe of the liver(S4), measuring approximately 14.3mm x 5.8mm, with delayed isodense enhancement. The intrahepatic duct system and common bile duct are not dilated and normal in course. The portal structure is clear. The position and morphology of gallbladder are normal. The spleen is enlarged, approximately 8 rib units, with no abnormal density. A round low-density lesion with no enhanced is found in the left kidney, with a diameter of approximately 4.3mm. Pelvis: The local lumen of the rectum and sigmoid colon is dilated, with increased wall thickness, measuring approximately 29mm x 55mm. On the enhancement scan, the lesions show lower uneven enhancement than normal intestinal walls. Multiple small lymph nodes are seen around the intestine. The myometrium enhancement is uneven.'' \\
				\\
				\textbf{Expected Output:} \\
				\texttt{\{} \\
				\texttt{\ \ "lower thorax": "normal",} \\
				\texttt{\ \ "liver and biliary tree": "Multiple mixed high- and low-density lesions are found in the liver, with CT values of approximately 17--186 HU. The largest lesion is in the posterior right lobe of the liver measuring approximately 69mm x 88mm. On the enhancement scan, no significant enhancement is seen in the lesions, with a low-density ring seen surrounding the lesions. Patchy slightly low-density foci are seen in the left lobe of the liver(S4), measuring approximately 14.3mm x 5.8mm, with delayed isodense enhancement.",} \\
				\texttt{\ \ "gallbladder": "normal",} \\
				\texttt{\ \ "spleen": "The spleen is enlarged, approximately 8 rib units.",} \\
				\texttt{\ \ "pancreas": "normal",} \\
				\texttt{\ \ "adrenal glands": "normal",} \\
				\texttt{\ \ "kidneys and ureters": "A round low-density lesion with no enhanced is found in the left kidney, with a diameter of approximately 4.3mm.",} \\
				\texttt{\ \ "gastrointestinal tract": "The local lumen of the rectum and sigmoid colon is dilated, with increased wall thickness, measuring approximately 29mm x 55mm. On the enhancement scan, the lesions show lower uneven enhancement than normal intestinal walls.",} \\
				\texttt{\ \ "peritoneum": "normal",} \\
				\texttt{\ \ "pelvic": "The myometrium enhancement is uneven.",} \\
				\texttt{\ \ "vasculature": "normal",} \\
				\texttt{\ \ "lymph nodes": "Multiple small lymph nodes are seen around the intestine.",} \\
				\texttt{\ \ "musculoskeletal": "normal"} \\
				\texttt{\}} \\
				\bottomrule
			\end{tabular}
			\caption{Illustrative example of structured information extraction from an abdominal CT report.}
			\label{tab:abdominal_ct_example}
		\end{table*}

		\newpage
		\begin{table*}[t]
			\centering
			\footnotesize
			\begin{tabular}{p{1.0\textwidth}}
				\toprule
				\textbf{Chest CT Example Case:} \\
				\textbf{Input Report:} ``LUNGS: Mild centrilobular emphysema. No focal consolidation. Several small nodules are stable. HEART AND MEDIASTINUM: The cardiomediastinal silhouette is normal in size. No pericardial effusion. No mediastinal lymphadenopathy. PLEURA: No pleural effusion or pneumothorax. BONES: Degenerative changes of the thoracic spine are present. FINDINGS GIVEN TO Dr. ..., AT THE CONCLUSION OF THE EXAMINATION, BY THE ONCALL RADIOLOGY RESIDENT. END OF IMPRESSION: SUMMARY 2: No acute abnormality.'' \\
				\\
				\textbf{Expected Output:} \\
				\texttt{\{} \\
				\texttt{\ \ "abdomen": "normal",} \\
				\texttt{\ \ "bone": "Degenerative changes of the thoracic spine are present.",} \\
				\texttt{\ \ "breast": "normal",} \\
				\texttt{\ \ "esophagus": "normal",} \\
				\texttt{\ \ "heart": "normal",} \\
				\texttt{\ \ "lung": "Mild centrilobular emphysema. There are several small nodules.",} \\
				\texttt{\ \ "mediastinum": "normal",} \\
				\texttt{\ \ "pleura": "normal",} \\
				\texttt{\ \ "thyroid": "normal",} \\
				\texttt{\ \ "trachea and bronchi": "normal"} \\
				\texttt{\}} \\
				\bottomrule
			\end{tabular}
			\caption{Illustrative example of structured information extraction from a chest CT report.}
			\label{tab:chest_ct_example}
		\end{table*}

		\clearpage
		\begin{table*}[t]
			\centering
			\footnotesize
			\begin{tabular}{p{1.0\textwidth}}
				\toprule
				\textbf{Instruction Prompt Templates for Report Generation:} \\
				\\
				\textbf{Chest Template} (used for thoracic CT report generation): \\
				\texttt{Generate a comprehensive and detailed diagnosis report for this chest CT image. Structure the report by describing the following regions in this exact order: abdomen, bone, breast, esophagus, heart, lung, mediastinum, pleura, thyroid, trachea and bronchi. For any region without abnormalities, state `normal.'.} \\
				\\
				\textbf{Abdomen Template} (used for abdominal CT report generation): \\
				\texttt{Generate a comprehensive and detailed diagnosis report for this abdomen CT image. Structure the report by describing the following regions in this exact order: lower thorax, liver and biliary tree, gallbladder, spleen, pancreas, adrenal glands, kidneys and ureters, gastrointestinal tract, peritoneum, pelvic, vasculature, lymph nodes, musculoskeletal. For any region without abnormalities, state `normal.'.} \\
				\\
				\textbf{Focus Template} (used for the ATLAS dataset, covering three anatomical regions only): \\
				\texttt{Please analyze the liver and biliary tree, pancreas, and kidneys and ureters areas from this abdominal CT scan. For any region without abnormalities, state `normal.'.} \\
				\bottomrule
			\end{tabular}
			\caption{Instruction prompt templates used for report generation fine-tuning.
				The \textbf{Chest} and \textbf{Abdomen} templates follow their respective standardized report structures.
				The \textbf{Focus} template is designed for the ATLAS dataset, which contains annotations for only three anatomical regions: liver and biliary tree, pancreas, and kidneys and ureters.}
			\label{tab:prompt_templates}
		\end{table*}

		\clearpage
		\begin{table*}[t]
			\centering
			\footnotesize
			\begin{tabular}{p{1.0\textwidth}}
				\toprule
				\textbf{Prompt Template for Medical Term Extraction (FORTE Pre-processing):} \\
				\\
				You are an expert medical terminologist. Your task is to extract specific medical terms from a given CT report and categorize them into four distinct dimensions. \\
				\\
				The four dimensions are: \\
				\textbf{1. Degree:} Terms describing size, intensity, severity, or quantity (e.g., ``mild'', ``diffuse'', ``multiple'', ``enlarged''). \\
				\textbf{2. Landmark:} Anatomical locations, organs, or spatial descriptors (e.g., ``liver'', ``pancreas'', ``right kidney'', ``pericholecystic''). \\
				\textbf{3. Feature:} Visual characteristics of diseases or primary abnormal findings (e.g., ``fat stranding'', ``thickening'', ``lesion'', ``stone'', ``ascites''). \\
				\textbf{4. Impression:} Final diagnostic conclusions or disease names (e.g., ``cholecystitis'', ``pancreatitis'', ``cirrhosis'', ``renal stone''). \\
				\\
				\textbf{Instructions:} \\
				$\bullet$ Analyze the provided CT report. \\
				$\bullet$ Extract all relevant terms for each of the four categories. \\
				$\bullet$ For each category, list the unique terms found in the report. \\
				$\bullet$ If no terms are found for a category, provide an empty list \texttt{[]}. \\
				$\bullet$ Your output \textbf{MUST} be a single JSON object with four keys: \texttt{"degree"}, \texttt{"landmark"}, \texttt{"feature"}, \texttt{"impression"}. The value for each key should be a list of strings. Do not add any other text or explanations before or after the JSON object. \\
				\\
				\textbf{Example Input Report:} \\
				\texttt{``IMPRESSION: Findings consistent with right upper lobe pneumonia. FINDINGS: There is a focal area of consolidation in the right upper lobe. Mild pleural effusion is noted. The mediastinum is unremarkable. No evidence of pulmonary embolism.''} \\
				\\
				\textbf{Example Output JSON:} \\
				\texttt{\{} \\
				\texttt{\ \ "degree": ["acute", "diffuse", "mild", "small", "unremarkable", "enlarged"],} \\
				\texttt{\ \ "landmark": ["gallbladder", "pericholecystic", "pancreas", "liver"],} \\
				\texttt{\ \ "feature": ["thickening", "fat stranding", "stone"],} \\
				\texttt{\ \ "impression": ["cholecystitis", "hepatic steatosis"]} \\
				\texttt{\}} \\
				\\
				Now, process the following CT report: \\
				\\
				\textbf{CT Report:} \\
				\texttt{\{report\_text\}} \\
				\\
				\textbf{Output JSON:} \\
				\bottomrule
			\end{tabular}
			\caption{Prompt used to extract structured medical terms from raw CT reports to update FORTE keywords.
				Given a CT report, the DeepSeek is instructed to identify and categorize terms into four dimensions:
				\textit{Degree} (severity/quantity descriptors),
				\textit{Landmark} (anatomical locations),
				\textit{Feature} (visual abnormality characteristics), and
				\textit{Impression} (diagnostic conclusions).
				The extracted terms were then manually reviewed and merged with synonyms to build the final FORTE terminology lexicon.
			}
			\label{tab:forte_extraction_prompt}
		\end{table*}
		
		\clearpage
		\begin{table}[htbp]
			\centering
			\begin{tabularx}{\textwidth}{l >{\centering\arraybackslash}X >{\centering\arraybackslash}X >{\centering\arraybackslash}X}
				\toprule
				Finding & Precision & Recall & F1-Score \\
				\midrule
				submucosal\_edema               & 0.98 & 1.00 & 0.99 \\
				renal\_hypodensities            & 0.94 & 0.98 & 0.96 \\
				aortic\_valve\_calcification    & 1.00 & 1.00 & 1.00 \\
				coronary\_calcification         & 0.98 & 1.00 & 0.99 \\
				thrombosis                      & 1.00 & 1.00 & 1.00 \\
				metastatic\_disease             & 1.00 & 1.00 & 1.00 \\
				pancreatic\_atrophy             & 1.00 & 1.00 & 1.00 \\
				renal\_cyst                     & 0.96 & 0.97 & 0.96 \\
				osteopenia                      & 1.00 & 0.95 & 0.97 \\
				surgically\_absent\_gallbladder & 0.95 & 1.00 & 0.98 \\
				atelectasis                     & 1.00 & 1.00 & 1.00 \\
				abdominal\_aortic\_aneurysm     & 1.00 & 1.00 & 1.00 \\
				anasarca                        & 1.00 & 1.00 & 1.00 \\
				hiatal\_hernia                  & 1.00 & 1.00 & 1.00 \\
				lymphadenopathy                 & 0.94 & 0.99 & 0.97 \\
				prostatomegaly                  & 0.98 & 1.00 & 0.99 \\
				biliary\_ductal\_dilation       & 0.90 & 1.00 & 0.95 \\
				cardiomegaly                    & 1.00 & 1.00 & 1.00 \\
				splenomegaly                    & 0.96 & 1.00 & 0.98 \\
				hepatomegaly                    & 1.00 & 1.00 & 1.00 \\
				atherosclerosis                 & 0.99 & 1.00 & 0.99 \\
				ascites                         & 0.98 & 0.97 & 0.98 \\
				pleural\_effusion               & 0.98 & 1.00 & 0.99 \\
				hepatic\_steatosis              & 0.94 & 1.00 & 0.97 \\
				appendicitis                    & 1.00 & 0.89 & 0.94 \\
				gallstones                      & 0.95 & 1.00 & 0.98 \\
				hydronephrosis                  & 0.97 & 0.97 & 0.97 \\
				bowel\_obstruction              & 1.00 & 1.00 & 1.00 \\
				free\_air                       & 0.86 & 0.88 & 0.87 \\
				fracture                        & 1.00 & 0.97 & 0.99 \\
				\midrule
				micro avg                       & 0.97 & 0.99 & 0.98 \\
				macro avg                       & 0.98 & 0.99 & 0.98 \\
				\bottomrule
			\end{tabularx}
			\caption{\textbf{Finetuned RadBERT’s performance in abdominal CT label extraction.} This table summarizes the performance of the fine-tuned RadBERT across individual abnormalities, as evaluated by precision, recall and F1 score. These results demonstrate the model’s effectiveness in accurately and efficiently automating abnormality-specific label extraction from radiology reports.}
			\label{tab:merlin_classification}
		\end{table}

		\newpage
		\begin{table}[htbp]
			\centering
			\begin{tabularx}{\textwidth}{l >{\centering\arraybackslash}X >{\centering\arraybackslash}X >{\centering\arraybackslash}X >{\centering\arraybackslash}X >{\centering\arraybackslash}X}
				\toprule
				Organ & Total & Train & Validation & Test & Ab. Rate \\
				\midrule
				Lung                & 58,843 & 46,922 & 5,567 & 6,354 & 64.89\% \\
				Bones               & 29,077 & 21,214 & 3,869 & 3,994 & 32.07\% \\
				Liver               & 25,937 & 18,298 & 3,721 & 3,918 & 28.60\% \\
				Kidney              & 24,274 & 17,227 & 3,677 & 3,370 & 26.77\% \\
				Soft Tissue         & 20,075 & 13,626 & 3,062 & 3,387 & 22.14\% \\
				Lymph Nodes         & 18,731 & 14,112 & 2,166 & 2,453 & 20.66\% \\
				Pleura              & 18,189 & 14,202 & 1,775 & 2,212 & 20.06\% \\
				Peritoneum/Omentum  & 16,034 & 10,280 & 2,766 & 2,988 & 17.68\% \\
				Large Intestine     & 14,431 & 8,934  & 2,702 & 2,795 & 15.91\% \\
				Heart               & 13,091 & 10,214 & 1,320 & 1,557 & 14.44\% \\
				Aorta               & 12,803 & 10,421 & 1,274 & 1,108 & 14.12\% \\
				Gallbladder         & 11,638 & 7,700  & 1,950 & 1,988 & 12.83\% \\
				Pancreas            & 10,795 & 8,018  & 1,534 & 1,243 & 11.90\% \\
				Spleen              & 9,139  & 6,636  & 1,293 & 1,210 & 10.08\% \\
				Pulmonary Artery    & 7,827  & 6,535  & 425   & 867   & 8.63\% \\
				Small Intestine     & 7,230  & 4,399  & 1,359 & 1,472 & 7.97\% \\
				Esophagus           & 6,651  & 5,146  & 781   & 724   & 7.33\% \\
				Stomach             & 6,138  & 4,340  & 884   & 914   & 6.77\% \\
				Trachea/Bronchi     & 5,746  & 4,652  & 445   & 649   & 6.34\% \\
				Biliary Ducts       & 5,442  & 3,399  & 937   & 1,106 & 6.00\% \\
				Bladder             & 5,169  & 3,145  & 1,005 & 1,019 & 5.70\% \\
				Adrenal Gland       & 5,161  & 3,792  & 677   & 692   & 5.69\% \\
				Uterus              & 4,398  & 2,672  & 842   & 884   & 4.85\% \\
				Ovaries             & 3,465  & 2,148  & 664   & 653   & 3.82\% \\
				Prostate            & 3,461  & 2,093  & 705   & 663   & 3.82\% \\
				Ureter              & 2,857  & 1,780  & 484   & 593   & 3.15\% \\
				Breast              & 2,521  & 1,929  & 292   & 300   & 2.78\% \\
				Thyroid             & 2,108  & 1,804  & 113   & 191   & 2.32\% \\
				Appendix            & 1,639  & 999    & 299   & 341   & 1.81\% \\
				Thymus              & 801    & 720    & 50    & 31    & 0.88\% \\
				\bottomrule
			\end{tabularx}
			\caption{Distribution of organ-specific abnormalities across the CTRgDB. This table summarizes the number of abnormal cases for each organ system and their distribution across the training, validation and test sets, together with the corresponding abnormality rates in the full cohort.}
			\label{tab:organ_distribution}
		\end{table}

		\clearpage
		\begin{table*}[htbp]
			\label{tab:ctrate_nlg}
			\resizebox{\textwidth}{!}{%
				\begin{tabular}{lcccccc}
					\toprule
					\multirow{2}{*}{\textbf{Model}} 
					& \textbf{BLEU-1} & \textbf{BLEU-2} & \textbf{BLEU-3} & \textbf{BLEU-4} 
					& \textbf{ROUGE-L} & \textbf{METEOR} \\
					& \scriptsize{[95\% CI]} & \scriptsize{[95\% CI]} & \scriptsize{[95\% CI]} & \scriptsize{[95\% CI]} 
					& \scriptsize{[95\% CI]} & \scriptsize{[95\% CI]} \\
					\midrule
					
					\multirow{2}{*}{Astra}
					& 0.4866$^*$ & 0.3803$^*$ & 0.3052$^*$ & 0.2501$^*$ & 0.4409$^*$ & 0.2397$^*$ \\
					& \scriptsize{[0.4769, 0.4958]}
					& \scriptsize{[0.3723, 0.3878]}
					& \scriptsize{[0.2978, 0.3119]}
					& \scriptsize{[0.2433, 0.2566]}
					& \scriptsize{[0.4330, 0.4490]}
					& \scriptsize{[0.2359, 0.2434]} \\
					\thinhline 
					
					\multirow{2}{*}{Gemini-3}
					& 0.1503 & 0.1171 & 0.0982 & 0.0850 & 0.3665 & 0.1455 \\
					& \scriptsize{[0.1394, 0.1606]}
					& \scriptsize{[0.1083, 0.1257]}
					& \scriptsize{[0.0907, 0.1056]}
					& \scriptsize{[0.0785, 0.0918]}
					& \scriptsize{[0.3568, 0.3769]}
					& \scriptsize{[0.1414, 0.1496]} \\
					\thinhline 
					\multirow{2}{*}{Qwen3-VL-8B}
					& 0.2132 & 0.1602 & 0.1328 & 0.1143 & 0.3293 & 0.1407 \\
					& \scriptsize{[0.2005, 0.2265]}
					& \scriptsize{[0.1505, 0.1702]}
					& \scriptsize{[0.1245, 0.1413]}
					& \scriptsize{[0.1070, 0.1216]}
					& \scriptsize{[0.3204, 0.3384]}
					& \scriptsize{[0.1368, 0.1447]} \\
					\thinhline 
					\multirow{2}{*}{HuluMed-32B}
					& 0.3001 & 0.2324 & 0.1886 & 0.1575 & 0.3902 & 0.1826 \\
					& \scriptsize{[0.2868, 0.3137]}
					& \scriptsize{[0.2219, 0.2432]}
					& \scriptsize{[0.1798, 0.1977]}
					& \scriptsize{[0.1498, 0.1653]}
					& \scriptsize{[0.3815, 0.3988]}
					& \scriptsize{[0.1785, 0.1866]} \\
					\thinhline 
					\multirow{2}{*}{HuluMed-14B}
					& 0.3025 & 0.2322 & 0.1877 & 0.1562 & 0.3801 & 0.1794 \\
					& \scriptsize{[0.2885, 0.3166]}
					& \scriptsize{[0.2216, 0.2432]}
					& \scriptsize{[0.1786, 0.1967]}
					& \scriptsize{[0.1482, 0.1641]}
					& \scriptsize{[0.3714, 0.3886]}
					& \scriptsize{[0.1753, 0.1836]} \\
					\thinhline 
					\multirow{2}{*}{HuluMed-7B}
					& 0.2970 & 0.2269 & 0.1832 & 0.1527 & 0.3688 & 0.1756 \\
					& \scriptsize{[0.2828, 0.3107]}
					& \scriptsize{[0.2160, 0.2373]}
					& \scriptsize{[0.1741, 0.1920]}
					& \scriptsize{[0.1449, 0.1604]}
					& \scriptsize{[0.3603, 0.3777]}
					& \scriptsize{[0.1714, 0.1797]} \\
					\thinhline 
					\multirow{2}{*}{M3D}
					& 0.1841 & 0.1404 & 0.1166 & 0.1000 & 0.3219 & 0.1397 \\
					& \scriptsize{[0.1717, 0.1967]}
					& \scriptsize{[0.1306, 0.1505]}
					& \scriptsize{[0.1083, 0.1252]}
					& \scriptsize{[0.0926, 0.1076]}
					& \scriptsize{[0.3145, 0.3293]}
					& \scriptsize{[0.1356, 0.1436]} \\
					\thinhline 
					\multirow{2}{*}{RadFM}
					& 0.1065 & 0.0869 & 0.0758 & 0.0674 & 0.3376 & 0.1318 \\
					& \scriptsize{[0.0955, 0.1177]}
					& \scriptsize{[0.0778, 0.0963]}
					& \scriptsize{[0.0679, 0.0840]}
					& \scriptsize{[0.0603, 0.0747]}
					& \scriptsize{[0.3289, 0.3463]}
					& \scriptsize{[0.1274, 0.1361]} \\
					\thinhline 
					\multirow{2}{*}{Lingshu-32B}
					& 0.2873 & 0.2068 & 0.1645 & 0.1371 & 0.3020 & 0.1467 \\
					& \scriptsize{[0.2766, 0.2979]}
					& \scriptsize{[0.1984, 0.2152]}
					& \scriptsize{[0.1572, 0.1718]}
					& \scriptsize{[0.1308, 0.1436]}
					& \scriptsize{[0.2961, 0.3076]}
					& \scriptsize{[0.1433, 0.1502]} \\
					\thinhline 
					\multirow{2}{*}{Lingshu-7B}
					& 0.0611 & 0.0510 & 0.0452 & 0.0406 & 0.3539 & 0.1302 \\
					& \scriptsize{[0.0540, 0.0684]}
					& \scriptsize{[0.0450, 0.0572]}
					& \scriptsize{[0.0398, 0.0508]}
					& \scriptsize{[0.0357, 0.0457]}
					& \scriptsize{[0.3434, 0.3640]}
					& \scriptsize{[0.1259, 0.1346]} \\
					\thinhline 
					\multirow{2}{*}{MedGemma-27B}
					& 0.1174 & 0.0952 & 0.0826 & 0.0733 & 0.3540 & 0.1331 \\
					& \scriptsize{[0.1058, 0.1294]}
					& \scriptsize{[0.0858, 0.1049]}
					& \scriptsize{[0.0745, 0.0910]}
					& \scriptsize{[0.0661, 0.0809]}
					& \scriptsize{[0.3431, 0.3650]}
					& \scriptsize{[0.1290, 0.1371]} \\
					\thinhline 
					\multirow{2}{*}{MedGemma-4B}
					& 0.0257 & 0.0228 & 0.0209 & 0.0192 & 0.3730 & 0.1303 \\
					& \scriptsize{[0.0221, 0.0295]}
					& \scriptsize{[0.0195, 0.0262]}
					& \scriptsize{[0.0179, 0.0241]}
					& \scriptsize{[0.0165, 0.0222]}
					& \scriptsize{[0.3614, 0.3848]}
					& \scriptsize{[0.1257, 0.1350]} \\
					\thinhline 
					\multirow{2}{*}{R2GenGPT}
					& 0.3330 & 0.2620 & 0.2133 & 0.1779 & 0.4198 & 0.1980 \\
					& \scriptsize{[0.3197, 0.3459]}
					& \scriptsize{[0.2513, 0.2725]}
					& \scriptsize{[0.2041, 0.2221]}
					& \scriptsize{[0.1698, 0.1856]}
					& \scriptsize{[0.4120, 0.4277]}
					& \scriptsize{[0.1937, 0.2021]} \\
					\thinhline 
					\multirow{2}{*}{LLaVA}
					& 0.2012 & 0.1641 & 0.1387 & 0.1198 & 0.4172 & 0.1754 \\
					& \scriptsize{[0.1891, 0.2129]}
					& \scriptsize{[0.1541, 0.1740]}
					& \scriptsize{[0.1300, 0.1474]}
					& \scriptsize{[0.1122, 0.1275]}
					& \scriptsize{[0.4081, 0.4263]}
					& \scriptsize{[0.1709, 0.1796]} \\
					
					\bottomrule
				\end{tabular}%
			}
			\centering
			\caption{\textbf{Natural language generation metrics on the CT-Rate dataset} ($n = 1564$ cases; 95\% confidence intervals estimated via 2{,}000 bootstrap iterations). $^*$ represents a significant improvement between Astra and the second-best baseline with $P < 0.05$, otherwise shows a non-significant improvement.}
		\end{table*}
		
		\clearpage
		\begin{table*}[htbp]
			\label{tab:ctrate_clinical_prf1}
			\begin{tabularx}{\textwidth}{l >{\centering\arraybackslash}X >{\centering\arraybackslash}X >{\centering\arraybackslash}X}
				\toprule
				\textbf{Model}
				& \textbf{micro Precision} & \textbf{micro Recall} & \textbf{micro F1} \\
				& \scriptsize{[95\% CI]} & \scriptsize{[95\% CI]} & \scriptsize{[95\% CI]} \\
				\midrule
				
				\multirow{2}{*}{Astra}
				& 0.5789$^*$ & 0.5242$^*$ & 0.5502$^*$ \\
				& \scriptsize{[0.5632, 0.5948]}
				& \scriptsize{[0.5103, 0.5381]}
				& \scriptsize{[0.5380, 0.5628]} \\
				\thinhline
				
				\multirow{2}{*}{Gemini-3}
				& 0.4134 & 0.1828 & 0.2535 \\
				& \scriptsize{[0.3917, 0.4363]}
				& \scriptsize{[0.1714, 0.1944]}
				& \scriptsize{[0.2397, 0.2677]} \\
				\thinhline
				
				\multirow{2}{*}{Qwen3-VL-8B}
				& 0.3561 & 0.0695 & 0.1163 \\
				& \scriptsize{[0.3244, 0.3867]}
				& \scriptsize{[0.0611, 0.0779]}
				& \scriptsize{[0.1030, 0.1292]} \\
				\thinhline
				
				\multirow{2}{*}{HuluMed-32B}
				& 0.4687 & 0.3210 & 0.3810 \\
				& \scriptsize{[0.4494, 0.4868]}
				& \scriptsize{[0.3058, 0.3360]}
				& \scriptsize{[0.3660, 0.3952]} \\
				\thinhline
				
				\multirow{2}{*}{HuluMed-14B}
				& 0.4373 & 0.3212 & 0.3704 \\
				& \scriptsize{[0.4182, 0.4558]}
				& \scriptsize{[0.3065, 0.3366]}
				& \scriptsize{[0.3561, 0.3850]} \\
				\thinhline
				
				\multirow{2}{*}{HuluMed-7B}
				& 0.4682 & 0.2515 & 0.3272 \\
				& \scriptsize{[0.4484, 0.4882]}
				& \scriptsize{[0.2368, 0.2660]}
				& \scriptsize{[0.3113, 0.3426]} \\
				\thinhline
				
				\multirow{2}{*}{M3D}
				& 0.2446 & 0.0997 & 0.1417 \\
				& \scriptsize{[0.2262, 0.2637]}
				& \scriptsize{[0.0915, 0.1082]}
				& \scriptsize{[0.1311, 0.1528]} \\
				\thinhline
				
				\multirow{2}{*}{RadFM}
				& 0.2850 & 0.0644 & 0.1050 \\
				& \scriptsize{[0.2600, 0.3125]}
				& \scriptsize{[0.0577, 0.0716]}
				& \scriptsize{[0.0949, 0.1162]} \\
				\thinhline
				
				\multirow{2}{*}{Lingshu-32B}
				& 0.2102 & 0.0573 & 0.0901 \\
				& \scriptsize{[0.1857, 0.2347]}
				& \scriptsize{[0.0505, 0.0644]}
				& \scriptsize{[0.0796, 0.1008]} \\
				\thinhline
				
				\multirow{2}{*}{Lingshu-7B}
				& 0.2462 & 0.0449 & 0.0759 \\
				& \scriptsize{[0.2167, 0.2757]}
				& \scriptsize{[0.0389, 0.0508]}
				& \scriptsize{[0.0661, 0.0854]} \\
				\thinhline
				
				\multirow{2}{*}{MedGemma-27B}
				& 0.2907 & 0.0050 & 0.0098 \\
				& \scriptsize{[0.2048, 0.3810]}
				& \scriptsize{[0.0031, 0.0071]}
				& \scriptsize{[0.0061, 0.0140]} \\
				\thinhline
				
				\multirow{2}{*}{MedGemma-4B}
				& 0.3643 & 0.0101 & 0.0197 \\
				& \scriptsize{[0.2857, 0.4468]}
				& \scriptsize{[0.0076, 0.0129]}
				& \scriptsize{[0.0148, 0.0251]} \\
				\thinhline
				
				\multirow{2}{*}{R2GenGPT}
				& 0.5118 & 0.3580 & 0.4213 \\
				& \scriptsize{[0.4941, 0.5300]}
				& \scriptsize{[0.3448, 0.3714]}
				& \scriptsize{[0.4079, 0.4348]} \\
				\thinhline
				
				\multirow{2}{*}{LLaVA}
				& 0.5406 & 0.2751 & 0.3647 \\
				& \scriptsize{[0.5220, 0.5613]}
				& \scriptsize{[0.2630, 0.2876]}
				& \scriptsize{[0.3510, 0.3782]} \\
				
				\bottomrule
			\end{tabularx}
			\centering
			\caption{\textbf{RadBERT classification performance (micro Precision, Recall, F1) on the CT-Rate dataset} ($n = 1{,}564$ cases; 95\% confidence intervals estimated via 2{,}000 bootstrap iterations). $^*$ indicates the best performance across all models for each metric.}
		\end{table*}

		\clearpage
		\begin{table*}[htbp]
			\label{tab:ctrate_rate_forte}
			\resizebox{\textwidth}{!}{%
				\begin{tabular}{lcccccc}
					\toprule
					\multirow{2}{*}{\textbf{Model}} 
					& \textbf{RaTE-Score} & \textbf{F.-Degree} & \textbf{F.-Landmark} & \textbf{F.-Feature} & \textbf{F.-Impression} & \textbf{F.-Overall} \\
					& \scriptsize{[95\% CI]} & \scriptsize{[95\% CI]} & \scriptsize{[95\% CI]} & \scriptsize{[95\% CI]} & \scriptsize{[95\% CI]} & \scriptsize{[95\% CI]} \\
					\midrule
					
					\multirow{2}{*}{Astra}
					& 0.3510$^*$  & 0.3844$^*$  & 0.5140$^*$  & 0.4159$^*$  & 0.4457$^*$  & 0.4400$^*$  \\
					& \scriptsize{[0.3436, 0.3580]}
					& \scriptsize{[0.3739, 0.3946]}
					& \scriptsize{[0.5033, 0.5252]}
					& \scriptsize{[0.4034, 0.4273]}
					& \scriptsize{[0.4306, 0.4601]}
					& \scriptsize{[0.4300, 0.4493]} \\
					\thinhline
					
					\multirow{2}{*}{Gemini-3}
					& 0.2225 & 0.1817 & 0.2243 & 0.1409 & 0.1459 & 0.1732 \\
					& \scriptsize{[0.2190, 0.2261]}
					& \scriptsize{[0.1727, 0.1910]}
					& \scriptsize{[0.2151, 0.2340]}
					& \scriptsize{[0.1325, 0.1494]}
					& \scriptsize{[0.1340, 0.1579]}
					& \scriptsize{[0.1659, 0.1810]} \\
					\thinhline
					
					\multirow{2}{*}{Qwen3-VL-8B}
					& 0.2058 & 0.0963 & 0.1831 & 0.0648 & 0.0636 & 0.1020 \\
					& \scriptsize{[0.2022, 0.2095]}
					& \scriptsize{[0.0884, 0.1045]}
					& \scriptsize{[0.1743, 0.1918]}
					& \scriptsize{[0.0578, 0.0723]}
					& \scriptsize{[0.0544, 0.0741]}
					& \scriptsize{[0.0955, 0.1090]} \\
					\thinhline
					
					\multirow{2}{*}{HuluMed-32B}
					& 0.2528 & 0.2274 & 0.3059 & 0.2278 & 0.2388 & 0.2500 \\
					& \scriptsize{[0.2479, 0.2576]}
					& \scriptsize{[0.2177, 0.2366]}
					& \scriptsize{[0.2948, 0.3168]}
					& \scriptsize{[0.2174, 0.2380]}
					& \scriptsize{[0.2252, 0.2528]}
					& \scriptsize{[0.2414, 0.2585]} \\
					\thinhline
					
					\multirow{2}{*}{HuluMed-14B}
					& 0.2534 & 0.2276 & 0.2875 & 0.2157 & 0.2402 & 0.2427 \\
					& \scriptsize{[0.2486, 0.2584]}
					& \scriptsize{[0.2177, 0.2369]}
					& \scriptsize{[0.2769, 0.2983]}
					& \scriptsize{[0.2055, 0.2257]}
					& \scriptsize{[0.2265, 0.2541]}
					& \scriptsize{[0.2340, 0.2514]} \\
					\thinhline
					
					\multirow{2}{*}{HuluMed-7B}
					& 0.2361 & 0.1926 & 0.2708 & 0.1889 & 0.1923 & 0.2111 \\
					& \scriptsize{[0.2315, 0.2410]}
					& \scriptsize{[0.1827, 0.2026]}
					& \scriptsize{[0.2590, 0.2830]}
					& \scriptsize{[0.1781, 0.1997]}
					& \scriptsize{[0.1776, 0.2062]}
					& \scriptsize{[0.2018, 0.2202]} \\
					\thinhline
					
					\multirow{2}{*}{M3D}
					& 0.2038 & 0.1352 & 0.1989 & 0.0740 & 0.0677 & 0.1189 \\
					& \scriptsize{[0.2009, 0.2066]}
					& \scriptsize{[0.1272, 0.1435]}
					& \scriptsize{[0.1900, 0.2078]}
					& \scriptsize{[0.0677, 0.0801]}
					& \scriptsize{[0.0590, 0.0762]}
					& \scriptsize{[0.1131, 0.1248]} \\
					\thinhline
					
					\multirow{2}{*}{RadFM}
					& 0.1881 & 0.0688 & 0.0891 & 0.0556 & 0.0537 & 0.0668 \\
					& \scriptsize{[0.1856, 0.1905]}
					& \scriptsize{[0.0619, 0.0754]}
					& \scriptsize{[0.0799, 0.0978]}
					& \scriptsize{[0.0489, 0.0621]}
					& \scriptsize{[0.0450, 0.0620]}
					& \scriptsize{[0.0608, 0.0725]} \\
					\thinhline
					
					\multirow{2}{*}{Lingshu-32B}
					& 0.2232 & 0.1158 & 0.2071 & 0.0654 & 0.0730 & 0.1153 \\
					& \scriptsize{[0.2201, 0.2264]}
					& \scriptsize{[0.1083, 0.1238]}
					& \scriptsize{[0.1999, 0.2141]}
					& \scriptsize{[0.0591, 0.0718]}
					& \scriptsize{[0.0631, 0.0827]}
					& \scriptsize{[0.1100, 0.1204]} \\
					\thinhline
					
					\multirow{2}{*}{Lingshu-7B}
					& 0.1690 & 0.0635 & 0.0601 & 0.0426 & 0.0464 & 0.0532 \\
					& \scriptsize{[0.1666, 0.1715]}
					& \scriptsize{[0.0572, 0.0701]}
					& \scriptsize{[0.0542, 0.0665]}
					& \scriptsize{[0.0374, 0.0479]}
					& \scriptsize{[0.0387, 0.0542]}
					& \scriptsize{[0.0487, 0.0579]} \\
					\thinhline
					
					\multirow{2}{*}{MedGemma-27B}
					& 0.1622 & 0.0327 & 0.0546 & 0.0147 & 0.0069 & 0.0272 \\
					& \scriptsize{[0.1599, 0.1646]}
					& \scriptsize{[0.0279, 0.0378]}
					& \scriptsize{[0.0487, 0.0604]}
					& \scriptsize{[0.0113, 0.0185]}
					& \scriptsize{[0.0039, 0.0104]}
					& \scriptsize{[0.0241, 0.0305]} \\
					\thinhline
					
					\multirow{2}{*}{MedGemma-4B}
					& 0.1513 & 0.0191 & 0.0336 & 0.0132 & 0.0119 & 0.0194 \\
					& \scriptsize{[0.1490, 0.1535]}
					& \scriptsize{[0.0143, 0.0240]}
					& \scriptsize{[0.0278, 0.0398]}
					& \scriptsize{[0.0096, 0.0170]}
					& \scriptsize{[0.0073, 0.0170]}
					& \scriptsize{[0.0157, 0.0234]} \\
					\thinhline
					
					\multirow{2}{*}{R2GenGPT}
					& 0.3084 & 0.2798 & 0.3829 & 0.2886 & 0.3167 & 0.3170 \\
					& \scriptsize{[0.3021, 0.3151]}
					& \scriptsize{[0.2699, 0.2897]}
					& \scriptsize{[0.3724, 0.3941]}
					& \scriptsize{[0.2770, 0.3000]}
					& \scriptsize{[0.3023, 0.3317]}
					& \scriptsize{[0.3081, 0.3259]} \\
					\thinhline
					
					\multirow{2}{*}{LLaVA}
					& 0.2911 & 0.2658 & 0.3025 & 0.2503 & 0.2988 & 0.2793 \\
					& \scriptsize{[0.2847, 0.2972]}
					& \scriptsize{[0.2556, 0.2758]}
					& \scriptsize{[0.2908, 0.3132]}
					& \scriptsize{[0.2381, 0.2625]}
					& \scriptsize{[0.2827, 0.3143]}
					& \scriptsize{[0.2696, 0.2891]} \\
					
					\bottomrule
				\end{tabular}%
			}
			\centering
			\caption{\textbf{Fine-grained caption metrics (RaTE-Score, FORTE) on the CT-Rate dataset} ($n = 1564$ cases; 95\% confidence intervals estimated via 2{,}000 bootstrap iterations). $^*$ indicates the best performance across all models for each metric. F. means FORTE metric.}
		\end{table*}

		\clearpage
		\begin{table*}[htbp]
			\label{tab:merlin_nlg}
			\resizebox{\textwidth}{!}{%
				\begin{tabular}{lcccccc}
					\toprule
					\multirow{2}{*}{\textbf{Model}}
					& \textbf{BLEU-1} & \textbf{BLEU-2} & \textbf{BLEU-3} & \textbf{BLEU-4}
					& \textbf{ROUGE-L} & \textbf{METEOR} \\
					& \scriptsize{[95\% CI]} & \scriptsize{[95\% CI]} & \scriptsize{[95\% CI]} & \scriptsize{[95\% CI]}
					& \scriptsize{[95\% CI]} & \scriptsize{[95\% CI]} \\
					\midrule
					
					\multirow{2}{*}{Astra}
					& 0.3898$^*$ & 0.2844$^*$ & 0.2185$^*$ & 0.1759$^*$ & 0.3928 & 0.2065$^*$ \\
					& \scriptsize{[0.3786, 0.4016]}
					& \scriptsize{[0.2757, 0.2936]}
					& \scriptsize{[0.2111, 0.2264]}
					& \scriptsize{[0.1691, 0.1828]}
					& \scriptsize{[0.3845, 0.4003]}
					& \scriptsize{[0.2027, 0.2103]} \\
					\thinhline
					
					\multirow{2}{*}{Gemini-3}
					& 0.1673 & 0.1313 & 0.1085 & 0.0929 & 0.3478 & 0.1526 \\
					& \scriptsize{[0.1562, 0.1790]}
					& \scriptsize{[0.1221, 0.1406]}
					& \scriptsize{[0.1007, 0.1166]}
					& \scriptsize{[0.0860, 0.1001]}
					& \scriptsize{[0.3382, 0.3571]}
					& \scriptsize{[0.1484, 0.1568]} \\
					\thinhline
					
					\multirow{2}{*}{Qwen3-VL-8B}
					& 0.2719 & 0.1992 & 0.1591 & 0.1339 & 0.3100 & 0.1504 \\
					& \scriptsize{[0.2596, 0.2844]}
					& \scriptsize{[0.1894, 0.2089]}
					& \scriptsize{[0.1507, 0.1674]}
					& \scriptsize{[0.1266, 0.1414]}
					& \scriptsize{[0.3017, 0.3177]}
					& \scriptsize{[0.1466, 0.1541]} \\
					\thinhline
					
					\multirow{2}{*}{HuluMed-32B}
					& 0.2971 & 0.1816 & 0.1196 & 0.0859 & 0.2410 & 0.1708 \\
					& \scriptsize{[0.2911, 0.3034]}
					& \scriptsize{[0.1779, 0.1854]}
					& \scriptsize{[0.1169, 0.1225]}
					& \scriptsize{[0.0835, 0.0885]}
					& \scriptsize{[0.2375, 0.2445]}
					& \scriptsize{[0.1683, 0.1734]} \\
					\thinhline
					
					\multirow{2}{*}{HuluMed-14B}
					& 0.3453 & 0.2297 & 0.1628 & 0.1241 & 0.2555 & 0.1593 \\
					& \scriptsize{[0.3390, 0.3494]}
					& \scriptsize{[0.2253, 0.2328]}
					& \scriptsize{[0.1590, 0.1656]}
					& \scriptsize{[0.1204, 0.1269]}
					& \scriptsize{[0.2509, 0.2600]}
					& \scriptsize{[0.1561, 0.1623]} \\
					\thinhline
					
					\multirow{2}{*}{HuluMed-7B}
					& 0.2876 & 0.2111 & 0.1660 & 0.1370 & 0.3084 & 0.1587 \\
					& \scriptsize{[0.2734, 0.3013]}
					& \scriptsize{[0.2004, 0.2215]}
					& \scriptsize{[0.1571, 0.1746]}
					& \scriptsize{[0.1292, 0.1445]}
					& \scriptsize{[0.3013, 0.3154]}
					& \scriptsize{[0.1546, 0.1625]} \\
					\thinhline
					
					\multirow{2}{*}{M3D}
					& 0.2537 & 0.1869 & 0.1479 & 0.1229 & 0.3026 & 0.1514 \\
					& \scriptsize{[0.2404, 0.2673]}
					& \scriptsize{[0.1767, 0.1973]}
					& \scriptsize{[0.1391, 0.1565]}
					& \scriptsize{[0.1152, 0.1304]}
					& \scriptsize{[0.2956, 0.3097]}
					& \scriptsize{[0.1475, 0.1553]} \\
					\thinhline
					
					\multirow{2}{*}{RadFM}
					& 0.1306 & 0.1059 & 0.0894 & 0.0776 & 0.3311 & 0.1434 \\
					& \scriptsize{[0.1194, 0.1424]}
					& \scriptsize{[0.0966, 0.1154]}
					& \scriptsize{[0.0813, 0.0977]}
					& \scriptsize{[0.0704, 0.0851]}
					& \scriptsize{[0.3214, 0.3405]}
					& \scriptsize{[0.1388, 0.1478]} \\
					\thinhline
					
					\multirow{2}{*}{Lingshu-32B}
					& 0.2749 & 0.2004 & 0.1595 & 0.1336 & 0.3010 & 0.1502 \\
					& \scriptsize{[0.2627, 0.2869]}
					& \scriptsize{[0.1905, 0.2098]}
					& \scriptsize{[0.1509, 0.1677]}
					& \scriptsize{[0.1259, 0.1410]}
					& \scriptsize{[0.2941, 0.3076]}
					& \scriptsize{[0.1463, 0.1540]} \\
					\thinhline
					
					\multirow{2}{*}{Lingshu-7B}
					& 0.1236 & 0.0999 & 0.0841 & 0.0729 & 0.3377 & 0.1438 \\
					& \scriptsize{[0.1124, 0.1352]}
					& \scriptsize{[0.0906, 0.1094]}
					& \scriptsize{[0.0761, 0.0925]}
					& \scriptsize{[0.0657, 0.0802]}
					& \scriptsize{[0.3280, 0.3470]}
					& \scriptsize{[0.1394, 0.1481]} \\
					\thinhline
					
					\multirow{2}{*}{MedGemma-27B}
					& 0.1764 & 0.1368 & 0.1120 & 0.0948 & 0.3300 & 0.1484 \\
					& \scriptsize{[0.1648, 0.1888]}
					& \scriptsize{[0.1275, 0.1465]}
					& \scriptsize{[0.1040, 0.1200]}
					& \scriptsize{[0.0878, 0.1018]}
					& \scriptsize{[0.3203, 0.3394]}
					& \scriptsize{[0.1442, 0.1525]} \\
					\thinhline
					
					\multirow{2}{*}{MedGemma-4B}
					& 0.1688 & 0.1328 & 0.1090 & 0.0923 & 0.3350 & 0.1517 \\
					& \scriptsize{[0.1572, 0.1799]}
					& \scriptsize{[0.1233, 0.1421]}
					& \scriptsize{[0.1009, 0.1171]}
					& \scriptsize{[0.0852, 0.0994]}
					& \scriptsize{[0.3260, 0.3439]}
					& \scriptsize{[0.1474, 0.1558]} \\
					\thinhline
					
					\multirow{2}{*}{R2GenGPT}
					& 0.3525 & 0.2590 & 0.2007 & 0.1624 & 0.3698 & 0.1857 \\
					& \scriptsize{[0.3408, 0.3638]}
					& \scriptsize{[0.2498, 0.2679]}
					& \scriptsize{[0.1929, 0.2081]}
					& \scriptsize{[0.1554, 0.1690]}
					& \scriptsize{[0.3618, 0.3776]}
					& \scriptsize{[0.1816, 0.1897]} \\
					\thinhline
					
					\multirow{2}{*}{LLaVA}
					& 0.2607 & 0.2012 & 0.1630 & 0.1371 & 0.3899 & 0.1792 \\
					& \scriptsize{[0.2475, 0.2743]}
					& \scriptsize{[0.1907, 0.2119]}
					& \scriptsize{[0.1541, 0.1721]}
					& \scriptsize{[0.1291, 0.1451]}
					& \scriptsize{[0.3807, 0.3991]}
					& \scriptsize{[0.1750, 0.1834]} \\
					\bottomrule
				\end{tabular}%
			}
			\centering
			\caption{\textbf{Natural language generation metrics on the Merlin dataset} ($n = 1{,}000$ cases; 95\% confidence intervals estimated via 2{,}000 bootstrap iterations). $^*$ represents a significant improvement between Astra and the second-best baseline with $P < 0.05$, otherwise shows a non-significant improvement.}
		\end{table*}
		
		\clearpage
		\begin{table*}[htbp]
			\label{tab:merlin_clinical_prf1}
			\begin{tabularx}{\textwidth}{l >{\centering\arraybackslash}X >{\centering\arraybackslash}X >{\centering\arraybackslash}X}
				\toprule
				\textbf{Model}
				& \textbf{micro Precision} & \textbf{micro Recall} & \textbf{micro F1} \\
				& \scriptsize{[95\% CI]} & \scriptsize{[95\% CI]} & \scriptsize{[95\% CI]} \\
				\midrule
				
				\multirow{2}{*}{Astra}
				& 0.5263 & 0.4714$^*$ & 0.4973$^*$ \\
				& \scriptsize{[0.5069, 0.5463]}
				& \scriptsize{[0.4536, 0.4895]}
				& \scriptsize{[0.4810, 0.5137]} \\
				\thinhline
				
				\multirow{2}{*}{Gemini-3}
				& 0.2873 & 0.0845 & 0.1305 \\
				& \scriptsize{[0.2571, 0.3163]}
				& \scriptsize{[0.0740, 0.0947]}
				& \scriptsize{[0.1158, 0.1450]} \\
				\thinhline
				
				\multirow{2}{*}{Qwen3-VL-8B}
				& 0.0945 & 0.0363 & 0.0525 \\
				& \scriptsize{[0.0789, 0.1100]}
				& \scriptsize{[0.0301, 0.0425]}
				& \scriptsize{[0.0436, 0.0612]} \\
				\thinhline
				
				\multirow{2}{*}{HuluMed-32B}
				& 0.2900 & 0.1276 & 0.1772 \\
				& \scriptsize{[0.2641, 0.3170]}
				& \scriptsize{[0.1160, 0.1396]}
				& \scriptsize{[0.1620, 0.1929]} \\
				\thinhline
				
				\multirow{2}{*}{HuluMed-14B}
				& 0.2778 & 0.1542 & 0.1983 \\
				& \scriptsize{[0.2556, 0.3006]}
				& \scriptsize{[0.1415, 0.1675]}
				& \scriptsize{[0.1828, 0.2144]} \\
				\thinhline
				
				\multirow{2}{*}{HuluMed-7B}
				& 0.2320 & 0.0846 & 0.1240 \\
				& \scriptsize{[0.2076, 0.2580]}
				& \scriptsize{[0.0747, 0.0949]}
				& \scriptsize{[0.1102, 0.1384]} \\
				\thinhline
				
				\multirow{2}{*}{M3D}
				& 0.2026 & 0.0537 & 0.0849 \\
				& \scriptsize{[0.1741, 0.2295]}
				& \scriptsize{[0.0455, 0.0615]}
				& \scriptsize{[0.0723, 0.0966]} \\
				\thinhline
				
				\multirow{2}{*}{RadFM}
				& 0.2134 & 0.0202 & 0.0369 \\
				& \scriptsize{[0.1707, 0.2574]}
				& \scriptsize{[0.0156, 0.0248]}
				& \scriptsize{[0.0287, 0.0451]} \\
				\thinhline
				
				\multirow{2}{*}{Lingshu-32B}
				& 0.1406 & 0.0378 & 0.0595 \\
				& \scriptsize{[0.1183, 0.1652]}
				& \scriptsize{[0.0317, 0.0444]}
				& \scriptsize{[0.0500, 0.0698]} \\
				\thinhline
				
				\multirow{2}{*}{Lingshu-7B}
				& 0.1813 & 0.0375 & 0.0622 \\
				& \scriptsize{[0.1496, 0.2131]}
				& \scriptsize{[0.0307, 0.0446]}
				& \scriptsize{[0.0511, 0.0738]} \\
				\thinhline
				
				\multirow{2}{*}{MedGemma-27B}
				& 0.1748 & 0.0072 & 0.0139 \\
				& \scriptsize{[0.1146, 0.2381]}
				& \scriptsize{[0.0045, 0.0101]}
				& \scriptsize{[0.0088, 0.0194]} \\
				\thinhline
				
				\multirow{2}{*}{MedGemma-4B}
				& 0.2375 & 0.0329 & 0.0578 \\
				& \scriptsize{[0.1978, 0.2777]}
				& \scriptsize{[0.0271, 0.0393]}
				& \scriptsize{[0.0478, 0.0686]} \\
				\thinhline
				
				\multirow{2}{*}{R2GenGPT}
				& 0.4685 & 0.4007 & 0.4319 \\
				& \scriptsize{[0.4503, 0.4856]}
				& \scriptsize{[0.3842, 0.4174]}
				& \scriptsize{[0.4169, 0.4467]} \\
				\thinhline
				
				\multirow{2}{*}{LLaVA}
				& 0.5165 & 0.3566 & 0.4219 \\
				& \scriptsize{[0.4970, 0.5376]}
				& \scriptsize{[0.3404, 0.3731]}
				& \scriptsize{[0.4062, 0.4382]} \\
				
				\bottomrule
			\end{tabularx}
			\centering
			\caption{\textbf{RadBERT classification performance (micro Precision, Recall, F1) on the Merlin dataset} ($n = 1{,}000$ cases; 95\% confidence intervals estimated via 2{,}000 bootstrap iterations). $^*$ represents a significant improvement between Astra and the second-best baseline with $P < 0.05$, otherwise shows a non-significant improvement.}
		\end{table*}
		
		\clearpage
		\begin{table*}[htbp]
			\label{tab:merlin_rate_forte}
			\resizebox{\textwidth}{!}{%
				\begin{tabular}{lcccccc}
					\toprule
					\multirow{2}{*}{\textbf{Model}}
					& \textbf{RaTE-Score} & \textbf{F.-Degree} & \textbf{F.-Landmark} & \textbf{F.-Feature} & \textbf{F.-Impression} & \textbf{F.-Overall} \\
					& \scriptsize{[95\% CI]} & \scriptsize{[95\% CI]} & \scriptsize{[95\% CI]} & \scriptsize{[95\% CI]} & \scriptsize{[95\% CI]} & \scriptsize{[95\% CI]} \\
					\midrule
					
					\multirow{2}{*}{Astra}
					& 0.3564$^*$ & 0.3765$^*$ & 0.4107$^*$ & 0.3750$^*$ & 0.4684$^*$ & 0.4076$^*$ \\
					& \scriptsize{[0.3509, 0.3618]}
					& \scriptsize{[0.3648, 0.3872]}
					& \scriptsize{[0.4001, 0.4209]}
					& \scriptsize{[0.3654, 0.3844]}
					& \scriptsize{[0.4549, 0.4809]}
					& \scriptsize{[0.3992, 0.4157]} \\
					\thinhline
					
					\multirow{2}{*}{Gemini-3}
					& 0.2154 & 0.0975 & 0.1299 & 0.1170 & 0.1743 & 0.1296 \\
					& \scriptsize{[0.2117, 0.2193]}
					& \scriptsize{[0.0881, 0.1069]}
					& \scriptsize{[0.1219, 0.1386]}
					& \scriptsize{[0.1094, 0.1250]}
					& \scriptsize{[0.1617, 0.1870]}
					& \scriptsize{[0.1222, 0.1373]} \\
					\thinhline
					
					\multirow{2}{*}{Qwen3-VL-8B}
					& 0.2007 & 0.1001 & 0.0954 & 0.0735 & 0.0543 & 0.0808 \\
					& \scriptsize{[0.1981, 0.2035]}
					& \scriptsize{[0.0917, 0.1085]}
					& \scriptsize{[0.0890, 0.1017]}
					& \scriptsize{[0.0671, 0.0799]}
					& \scriptsize{[0.0470, 0.0618]}
					& \scriptsize{[0.0757, 0.0859]} \\
					\thinhline
					
					\multirow{2}{*}{HuluMed-32B}
					& 0.2115 & 0.1265 & 0.1968 & 0.1445 & 0.1203 & 0.1442 \\
					& \scriptsize{[0.2088, 0.2143]}
					& \scriptsize{[0.1195, 0.1336]}
					& \scriptsize{[0.1889, 0.2048]}
					& \scriptsize{[0.1378, 0.1511]}
					& \scriptsize{[0.1110, 0.1298]}
					& \scriptsize{[0.1391, 0.1496]} \\
					\thinhline
					
					\multirow{2}{*}{HuluMed-14B}
					& 0.2171 & 0.1389 & 0.1416 & 0.1382 & 0.1240 & 0.1357 \\
					& \scriptsize{[0.2143, 0.2199]}
					& \scriptsize{[0.1313, 0.1459]}
					& \scriptsize{[0.1337, 0.1496]}
					& \scriptsize{[0.1309, 0.1456]}
					& \scriptsize{[0.1135, 0.1354]}
					& \scriptsize{[0.1293, 0.1419]} \\
					\thinhline
					
					\multirow{2}{*}{HuluMed-7B}
					& 0.1963 & 0.0997 & 0.1153 & 0.0717 & 0.0744 & 0.0903 \\
					& \scriptsize{[0.1934, 0.1992]}
					& \scriptsize{[0.0916, 0.1079]}
					& \scriptsize{[0.1072, 0.1236]}
					& \scriptsize{[0.0652, 0.0788]}
					& \scriptsize{[0.0648, 0.0837]}
					& \scriptsize{[0.0843, 0.0962]} \\
					\thinhline
					
					\multirow{2}{*}{M3D}
					& 0.1891 & 0.0912 & 0.1064 & 0.0635 & 0.0476 & 0.0772 \\
					& \scriptsize{[0.1865, 0.1919]}
					& \scriptsize{[0.0827, 0.0996]}
					& \scriptsize{[0.0990, 0.1140]}
					& \scriptsize{[0.0575, 0.0697]}
					& \scriptsize{[0.0396, 0.0555]}
					& \scriptsize{[0.0716, 0.0827]} \\
					\thinhline
					
					\multirow{2}{*}{RadFM}
					& 0.1687 & 0.0431 & 0.0442 & 0.0225 & 0.0153 & 0.0313 \\
					& \scriptsize{[0.1668, 0.1708]}
					& \scriptsize{[0.0368, 0.0497]}
					& \scriptsize{[0.0385, 0.0503]}
					& \scriptsize{[0.0185, 0.0265]}
					& \scriptsize{[0.0104, 0.0205]}
					& \scriptsize{[0.0273, 0.0353]} \\
					\thinhline
					
					\multirow{2}{*}{Lingshu-32B}
					& 0.1955 & 0.0442 & 0.0880 & 0.0612 & 0.0468 & 0.0600 \\
					& \scriptsize{[0.1930, 0.1980]}
					& \scriptsize{[0.0384, 0.0498]}
					& \scriptsize{[0.0811, 0.0949]}
					& \scriptsize{[0.0547, 0.0678]}
					& \scriptsize{[0.0392, 0.0549]}
					& \scriptsize{[0.0554, 0.0648]} \\
					\thinhline
					
					\multirow{2}{*}{Lingshu-7B}
					& 0.1767 & 0.0371 & 0.0635 & 0.0382 & 0.0413 & 0.0450 \\
					& \scriptsize{[0.1743, 0.1790]}
					& \scriptsize{[0.0312, 0.0435]}
					& \scriptsize{[0.0570, 0.0704]}
					& \scriptsize{[0.0329, 0.0436]}
					& \scriptsize{[0.0335, 0.0489]}
					& \scriptsize{[0.0403, 0.0497]} \\
					\thinhline
					
					\multirow{2}{*}{MedGemma-27B}
					& 0.1719 & 0.0195 & 0.0570 & 0.0322 & 0.0218 & 0.0326 \\
					& \scriptsize{[0.1699, 0.1741]}
					& \scriptsize{[0.0154, 0.0239]}
					& \scriptsize{[0.0505, 0.0639]}
					& \scriptsize{[0.0275, 0.0368]}
					& \scriptsize{[0.0167, 0.0276]}
					& \scriptsize{[0.0291, 0.0364]} \\
					\thinhline
					
					\multirow{2}{*}{MedGemma-4B}
					& 0.1775 & 0.0392 & 0.0644 & 0.0398 & 0.0248 & 0.0421 \\
					& \scriptsize{[0.1755, 0.1797]}
					& \scriptsize{[0.0335, 0.0450]}
					& \scriptsize{[0.0577, 0.0711]}
					& \scriptsize{[0.0345, 0.0453]}
					& \scriptsize{[0.0190, 0.0307]}
					& \scriptsize{[0.0378, 0.0465]} \\
					\thinhline
					
					\multirow{2}{*}{R2GenGPT}
					& 0.3229 & 0.2965 & 0.3140 & 0.2882 & 0.3851 & 0.3210 \\
					& \scriptsize{[0.3176, 0.3285]}
					& \scriptsize{[0.2861, 0.3072]}
					& \scriptsize{[0.3043, 0.3237]}
					& \scriptsize{[0.2793, 0.2976]}
					& \scriptsize{[0.3713, 0.3985]}
					& \scriptsize{[0.3130, 0.3292]} \\
					\thinhline
					
					\multirow{2}{*}{LLaVA}
					& 0.3204 & 0.2640 & 0.3009 & 0.2749 & 0.3928 & 0.3082 \\
					& \scriptsize{[0.3144, 0.3267]}
					& \scriptsize{[0.2530, 0.2757]}
					& \scriptsize{[0.2901, 0.3110]}
					& \scriptsize{[0.2652, 0.2846]}
					& \scriptsize{[0.3769, 0.4078]}
					& \scriptsize{[0.2990, 0.3172]} \\

					\bottomrule
				\end{tabular}%
			}
			\centering
			\caption{\textbf{Fine-grained caption metrics (RaTE-Score, FORTE) on the Merlin dataset} ($n = 1{,}000$ cases; 95\% confidence intervals estimated via 2{,}000 bootstrap iterations). $^*$ represents a significant improvement between Astra and the second-best baseline with $P < 0.05$, otherwise shows a non-significant improvement. F. means FORTE metric.}
		\end{table*}

		\clearpage
		\begin{table*}[htbp]
			\label{tab:inspect_nlg}
			\resizebox{\textwidth}{!}{%
				\begin{tabular}{lcccccc}
					\toprule
					\multirow{2}{*}{\textbf{Model}}
					& \textbf{BLEU-1} & \textbf{BLEU-2} & \textbf{BLEU-3} & \textbf{BLEU-4}
					& \textbf{ROUGE-L} & \textbf{METEOR} \\
					& \scriptsize{[95\% CI]} & \scriptsize{[95\% CI]} & \scriptsize{[95\% CI]} & \scriptsize{[95\% CI]}
					& \scriptsize{[95\% CI]} & \scriptsize{[95\% CI]} \\
					\midrule
					
					\multirow{2}{*}{Astra}
					& 0.4622$^*$ & 0.3912$^*$ & 0.3479$^*$ & 0.3140$^*$ & 0.5394 & 0.2685$^*$ \\
					& \scriptsize{[0.4465, 0.4770]}
					& \scriptsize{[0.3769, 0.4047]}
					& \scriptsize{[0.3342, 0.3609]}
					& \scriptsize{[0.3010, 0.3265]}
					& \scriptsize{[0.5276, 0.5512]}
					& \scriptsize{[0.2620, 0.2753]} \\
					\thinhline
					
					\multirow{2}{*}{Gemini-3}
					& 0.3954 & 0.3092 & 0.2616 & 0.2262 & 0.4176 & 0.2216 \\
					& \scriptsize{[0.3860, 0.4052]}
					& \scriptsize{[0.3008, 0.3187]}
					& \scriptsize{[0.2537, 0.2706]}
					& \scriptsize{[0.2186, 0.2348]}
					& \scriptsize{[0.4071, 0.4286]}
					& \scriptsize{[0.2167, 0.2265]} \\
					\thinhline
					
					\multirow{2}{*}{Qwen3-VL-8B}
					& 0.3549 & 0.2845 & 0.2478 & 0.2212 & 0.4168 & 0.2216 \\
					& \scriptsize{[0.3433, 0.3676]}
					& \scriptsize{[0.2741, 0.2957]}
					& \scriptsize{[0.2382, 0.2585]}
					& \scriptsize{[0.2120, 0.2313]}
					& \scriptsize{[0.4048, 0.4294]}
					& \scriptsize{[0.2163, 0.2270]} \\
					\thinhline
					
					\multirow{2}{*}{HuluMed-32B}
					& 0.2623 & 0.1993 & 0.1651 & 0.1412 & 0.3456 & 0.2093 \\
					& \scriptsize{[0.2553, 0.2699]}
					& \scriptsize{[0.1932, 0.2059]}
					& \scriptsize{[0.1596, 0.1712]}
					& \scriptsize{[0.1359, 0.1469]}
					& \scriptsize{[0.3368, 0.3551]}
					& \scriptsize{[0.2055, 0.2135]} \\
					\thinhline
					
					\multirow{2}{*}{HuluMed-14B}
					& 0.3750 & 0.2973 & 0.2547 & 0.2228 & 0.4050 & 0.2229 \\
					& \scriptsize{[0.3639, 0.3865]}
					& \scriptsize{[0.2878, 0.3076]}
					& \scriptsize{[0.2460, 0.2641]}
					& \scriptsize{[0.2146, 0.2316]}
					& \scriptsize{[0.3954, 0.4142]}
					& \scriptsize{[0.2180, 0.2282]} \\
					\thinhline
					
					\multirow{2}{*}{HuluMed-7B}
					& 0.2800 & 0.2154 & 0.1806 & 0.1554 & 0.3467 & 0.2103 \\
					& \scriptsize{[0.2694, 0.2919]}
					& \scriptsize{[0.2067, 0.2251]}
					& \scriptsize{[0.1731, 0.1890]}
					& \scriptsize{[0.1486, 0.1631]}
					& \scriptsize{[0.3385, 0.3544]}
					& \scriptsize{[0.2060, 0.2145]} \\
					\thinhline
					
					\multirow{2}{*}{M3D}
					& 0.4314 & 0.3531 & 0.3092 & 0.2752 & 0.4237 & 0.2248 \\
					& \scriptsize{[0.4196, 0.4371]}
					& \scriptsize{[0.3425, 0.3595]}
					& \scriptsize{[0.2987, 0.3159]}
					& \scriptsize{[0.2653, 0.2819]}
					& \scriptsize{[0.4143, 0.4335]}
					& \scriptsize{[0.2193, 0.2308]} \\
					\thinhline
					
					\multirow{2}{*}{RadFM}
					& 0.4017 & 0.3451 & 0.3108 & 0.2821 & 0.4546 & 0.2291 \\
					& \scriptsize{[0.3865, 0.4162]}
					& \scriptsize{[0.3313, 0.3580]}
					& \scriptsize{[0.2980, 0.3232]}
					& \scriptsize{[0.2701, 0.2940]}
					& \scriptsize{[0.4446, 0.4645]}
					& \scriptsize{[0.2229, 0.2354]} \\
					\thinhline
					
					\multirow{2}{*}{Lingshu-32B}
					& 0.3036 & 0.2373 & 0.2020 & 0.1770 & 0.3451 & 0.2149 \\
					& \scriptsize{[0.2974, 0.3099]}
					& \scriptsize{[0.2329, 0.2419]}
					& \scriptsize{[0.1982, 0.2060]}
					& \scriptsize{[0.1735, 0.1806]}
					& \scriptsize{[0.3393, 0.3508]}
					& \scriptsize{[0.2108, 0.2191]} \\
					\thinhline
					
					\multirow{2}{*}{Lingshu-7B}
					& 0.3408 & 0.2990 & 0.2724 & 0.2495 & 0.5024 & 0.2343 \\
					& \scriptsize{[0.3227, 0.3593]}
					& \scriptsize{[0.2826, 0.3155]}
					& \scriptsize{[0.2571, 0.2876]}
					& \scriptsize{[0.2352, 0.2641]}
					& \scriptsize{[0.4903, 0.5146]}
					& \scriptsize{[0.2273, 0.2411]} \\
					\thinhline
					
					\multirow{2}{*}{MedGemma-27B}
					& 0.3974 & 0.3309 & 0.2931 & 0.2644 & 0.4372 & 0.2258 \\
					& \scriptsize{[0.3855, 0.4096]}
					& \scriptsize{[0.3194, 0.3425]}
					& \scriptsize{[0.2820, 0.3045]}
					& \scriptsize{[0.2535, 0.2754]}
					& \scriptsize{[0.4239, 0.4501]}
					& \scriptsize{[0.2197, 0.2316]} \\
					\thinhline
					
					\multirow{2}{*}{MedGemma-4B}
					& 0.2954 & 0.2681 & 0.2498 & 0.2335 & 0.5359 & 0.2411 \\
					& \scriptsize{[0.2765, 0.3151]}
					& \scriptsize{[0.2508, 0.2862]}
					& \scriptsize{[0.2334, 0.2666]}
					& \scriptsize{[0.2180, 0.2493]}
					& \scriptsize{[0.5216, 0.5502]}
					& \scriptsize{[0.2334, 0.2488]} \\
					\thinhline
					
					\multirow{2}{*}{R2GenGPT}
					& 0.4488 & 0.3824 & 0.3408 & 0.3080 & 0.5305 & 0.2587 \\
					& \scriptsize{[0.4319, 0.4643]}
					& \scriptsize{[0.3672, 0.3967]}
					& \scriptsize{[0.3264, 0.3541]}
					& \scriptsize{[0.2943, 0.3206]}
					& \scriptsize{[0.5186, 0.5426]}
					& \scriptsize{[0.2515, 0.2660]} \\
					\thinhline
					
					\multirow{2}{*}{LLaVA}
					& 0.3590 & 0.3189 & 0.2924 & 0.2703 & 0.5626 & 0.2610 \\
					& \scriptsize{[0.3403, 0.3767]}
					& \scriptsize{[0.3018, 0.3353]}
					& \scriptsize{[0.2765, 0.3080]}
					& \scriptsize{[0.2553, 0.2853]}
					& \scriptsize{[0.5491, 0.5762]}
					& \scriptsize{[0.2534, 0.2688]} \\
					
					\bottomrule
				\end{tabular}%
			}
			\centering
			\caption{\textbf{Natural language generation metrics on the Inspect dataset} ($n = 1{,}000$ cases; 95\% confidence intervals estimated via 2{,}000 bootstrap iterations). $^*$ represents a significant improvement between Astra and the second-best baseline with $P < 0.05$, otherwise shows a non-significant improvement.}
		\end{table*}

		\clearpage
		\begin{table*}[htbp]
			\label{tab:inspect_clinical_prf1}
			\begin{tabularx}{\textwidth}{l >{\centering\arraybackslash}X >{\centering\arraybackslash}X >{\centering\arraybackslash}X}
				\toprule
				\textbf{Model}
				& \textbf{micro Precision} & \textbf{micro Recall} & \textbf{micro F1} \\
				& \scriptsize{[95\% CI]} & \scriptsize{[95\% CI]} & \scriptsize{[95\% CI]} \\
				\midrule
				
				\multirow{2}{*}{Astra}
				& 0.4844 & 0.4808 & 0.4825$^*$ \\
				& \scriptsize{[0.4587, 0.5109]}
				& \scriptsize{[0.4565, 0.5055]}
				& \scriptsize{[0.4598, 0.5046]} \\
				\thinhline
				
				\multirow{2}{*}{Gemini-3}
				& 0.2693 & 0.4153$^*$ & 0.3267 \\
				& \scriptsize{[0.2482, 0.2895]}
				& \scriptsize{[0.3888, 0.4422]}
				& \scriptsize{[0.3050, 0.3471]} \\
				\thinhline
				
				\multirow{2}{*}{Qwen3-VL-8B}
				& 0.2690 & 0.2284 & 0.2471 \\
				& \scriptsize{[0.2438, 0.2957]}
				& \scriptsize{[0.2057, 0.2524]}
				& \scriptsize{[0.2246, 0.2703]} \\
				\thinhline
				
				\multirow{2}{*}{HuluMed-32B}
				& 0.1487 & 0.4004 & 0.2169 \\
				& \scriptsize{[0.1370, 0.1601]}
				& \scriptsize{[0.3742, 0.4271]}
				& \scriptsize{[0.2018, 0.2317]} \\
				\thinhline
				
				\multirow{2}{*}{HuluMed-14B}
				& 0.2180 & 0.2323 & 0.2249 \\
				& \scriptsize{[0.1970, 0.2406]}
				& \scriptsize{[0.2095, 0.2556]}
				& \scriptsize{[0.2049, 0.2456]} \\
				\thinhline
				
				\multirow{2}{*}{HuluMed-7B}
				& 0.1678 & 0.2107 & 0.1868 \\
				& \scriptsize{[0.1494, 0.1870]}
				& \scriptsize{[0.1883, 0.2337]}
				& \scriptsize{[0.1678, 0.2057]} \\
				\thinhline
				
				\multirow{2}{*}{M3D}
				& 0.1883 & 0.1458 & 0.1644 \\
				& \scriptsize{[0.1652, 0.2109]}
				& \scriptsize{[0.1277, 0.1645]}
				& \scriptsize{[0.1449, 0.1843]} \\
				\thinhline
				
				\multirow{2}{*}{RadFM}
				& 0.1814 & 0.1040 & 0.1322 \\
				& \scriptsize{[0.1541, 0.2098]}
				& \scriptsize{[0.0879, 0.1212]}
				& \scriptsize{[0.1124, 0.1520]} \\
				\thinhline
				
				\multirow{2}{*}{Lingshu-32B}
				& 0.2367 & 0.1909 & 0.2114 \\
				& \scriptsize{[0.2123, 0.2617]}
				& \scriptsize{[0.1715, 0.2119]}
				& \scriptsize{[0.1909, 0.2326]} \\
				\thinhline
				
				\multirow{2}{*}{Lingshu-7B}
				& 0.1681 & 0.1080 & 0.1315 \\
				& \scriptsize{[0.1436, 0.1945]}
				& \scriptsize{[0.0921, 0.1259]}
				& \scriptsize{[0.1128, 0.1519]} \\
				\thinhline
				
				\multirow{2}{*}{MedGemma-27B}
				& 0.2198 & 0.0270 & 0.0481 \\
				& \scriptsize{[0.1608, 0.2805]}
				& \scriptsize{[0.0185, 0.0366]}
				& \scriptsize{[0.0334, 0.0647]} \\
				\thinhline
				
				\multirow{2}{*}{MedGemma-4B}
				& 0.1939 & 0.0216 & 0.0389 \\
				& \scriptsize{[0.1346, 0.2549]}
				& \scriptsize{[0.0146, 0.0293]}
				& \scriptsize{[0.0264, 0.0522]} \\
				\thinhline
				
				\multirow{2}{*}{R2GenGPT}
				& 0.4414 & 0.3943 & 0.4165 \\
				& \scriptsize{[0.4123, 0.4699]}
				& \scriptsize{[0.3697, 0.4198]}
				& \scriptsize{[0.3925, 0.4403]} \\
				\thinhline
				
				\multirow{2}{*}{LLaVA}
				& 0.4926 & 0.3376 & 0.4006 \\
				& \scriptsize{[0.4601, 0.5244]}
				& \scriptsize{[0.3129, 0.3629]}
				& \scriptsize{[0.3743, 0.4255]} \\
				
				\bottomrule
			\end{tabularx}
			\centering
			\caption{\textbf{RadBERT classification performance (micro Precision, Recall, F1) on the Inspect dataset} ($n = 1{,}000$ cases; 95\% confidence intervals estimated via 2{,}000 bootstrap iterations). $^*$ represents a significant improvement between Astra and the second-best baseline with $P < 0.05$, otherwise shows a non-significant improvement.}
		\end{table*}
		
		\clearpage
		\begin{table*}[htbp]
			\label{tab:inspect_rate_forte}
			\resizebox{\textwidth}{!}{%
				\begin{tabular}{lcccccc}
					\toprule
					\multirow{2}{*}{\textbf{Model}}
					& \textbf{RaTE-Score} & \textbf{F.-Degree} & \textbf{F.-Landmark} & \textbf{F.-Feature} & \textbf{F.-Impression} & \textbf{F.-Overall} \\
					& \scriptsize{[95\% CI]} & \scriptsize{[95\% CI]} & \scriptsize{[95\% CI]} & \scriptsize{[95\% CI]} & \scriptsize{[95\% CI]} & \scriptsize{[95\% CI]} \\
					\midrule
					
					\multirow{2}{*}{Astra}
					& 0.3305$^*$ & 0.3601$^*$ & 0.4449$^*$ & 0.3793$^*$ & 0.4028$^*$ & 0.3968$^*$ \\
					& \scriptsize{[0.3202, 0.3408]}
					& \scriptsize{[0.3450, 0.3747]}
					& \scriptsize{[0.4296, 0.4595]}
					& \scriptsize{[0.3620, 0.3969]}
					& \scriptsize{[0.3852, 0.4190]}
					& \scriptsize{[0.3836, 0.4091]} \\
					\thinhline
					
					\multirow{2}{*}{Gemini-3}
					& 0.2335 & 0.1817 & 0.2519 & 0.2285 & 0.2074 & 0.2174 \\
					& \scriptsize{[0.2286, 0.2386]}
					& \scriptsize{[0.1707, 0.1932]}
					& \scriptsize{[0.2402, 0.2633]}
					& \scriptsize{[0.2134, 0.2427]}
					& \scriptsize{[0.1936, 0.2222]}
					& \scriptsize{[0.2078, 0.2275]} \\
					\thinhline
					
					\multirow{2}{*}{Qwen3-VL-8B}
					& 0.2140 & 0.1452 & 0.2025 & 0.1667 & 0.1767 & 0.1728 \\
					& \scriptsize{[0.2083, 0.2196]}
					& \scriptsize{[0.1334, 0.1570]}
					& \scriptsize{[0.1893, 0.2148]}
					& \scriptsize{[0.1517, 0.1809]}
					& \scriptsize{[0.1616, 0.1908]}
					& \scriptsize{[0.1617, 0.1826]} \\
					\thinhline
					
					\multirow{2}{*}{HuluMed-32B}
					& 0.2189 & 0.1819 & 0.2457 & 0.1711 & 0.2135 & 0.2024 \\
					& \scriptsize{[0.2144, 0.2233]}
					& \scriptsize{[0.1699, 0.1934]}
					& \scriptsize{[0.2337, 0.2570]}
					& \scriptsize{[0.1580, 0.1849]}
					& \scriptsize{[0.1972, 0.2291]}
					& \scriptsize{[0.1920, 0.2122]} \\
					\thinhline
					
					\multirow{2}{*}{HuluMed-14B}
					& 0.2082 & 0.1546 & 0.2144 & 0.1490 & 0.1612 & 0.1698 \\
					& \scriptsize{[0.2028, 0.2136]}
					& \scriptsize{[0.1422, 0.1672]}
					& \scriptsize{[0.2015, 0.2281]}
					& \scriptsize{[0.1354, 0.1625]}
					& \scriptsize{[0.1464, 0.1757]}
					& \scriptsize{[0.1584, 0.1805]} \\
					\thinhline
					
					\multirow{2}{*}{HuluMed-7B}
					& 0.2018 & 0.1364 & 0.1843 & 0.1167 & 0.1284 & 0.1415 \\
					& \scriptsize{[0.1969, 0.2066]}
					& \scriptsize{[0.1256, 0.1474]}
					& \scriptsize{[0.1718, 0.1965]}
					& \scriptsize{[0.1055, 0.1288]}
					& \scriptsize{[0.1154, 0.1425]}
					& \scriptsize{[0.1316, 0.1514]} \\
					\thinhline
					
					\multirow{2}{*}{M3D}
					& 0.1946 & 0.1262 & 0.1914 & 0.1008 & 0.1139 & 0.1331 \\
					& \scriptsize{[0.1896, 0.1995]}
					& \scriptsize{[0.1144, 0.1380]}
					& \scriptsize{[0.1786, 0.2043]}
					& \scriptsize{[0.0896, 0.1129]}
					& \scriptsize{[0.1021, 0.1262]}
					& \scriptsize{[0.1240, 0.1421]} \\
					\thinhline
					
					\multirow{2}{*}{RadFM}
					& 0.1920 & 0.1187 & 0.1439 & 0.0910 & 0.1040 & 0.1144 \\
					& \scriptsize{[0.1875, 0.1963]}
					& \scriptsize{[0.1052, 0.1323]}
					& \scriptsize{[0.1296, 0.1584]}
					& \scriptsize{[0.0775, 0.1045]}
					& \scriptsize{[0.0906, 0.1179]}
					& \scriptsize{[0.1035, 0.1258]} \\
					\thinhline
					
					\multirow{2}{*}{Lingshu-32B}
					& 0.2234 & 0.1203 & 0.2262 & 0.1413 & 0.1629 & 0.1627 \\
					& \scriptsize{[0.2190, 0.2279]}
					& \scriptsize{[0.1093, 0.1319]}
					& \scriptsize{[0.2152, 0.2367]}
					& \scriptsize{[0.1304, 0.1522]}
					& \scriptsize{[0.1503, 0.1756]}
					& \scriptsize{[0.1540, 0.1714]} \\
					\thinhline
					
					\multirow{2}{*}{Lingshu-7B}
					& 0.1751 & 0.0886 & 0.1038 & 0.0699 & 0.0626 & 0.0812 \\
					& \scriptsize{[0.1703, 0.1801]}
					& \scriptsize{[0.0787, 0.0982]}
					& \scriptsize{[0.0914, 0.1164]}
					& \scriptsize{[0.0589, 0.0812]}
					& \scriptsize{[0.0522, 0.0738]}
					& \scriptsize{[0.0724, 0.0900]} \\
					\thinhline
					
					\multirow{2}{*}{MedGemma-27B}
					& 0.1636 & 0.0386 & 0.0859 & 0.0473 & 0.0491 & 0.0552 \\
					& \scriptsize{[0.1593, 0.1677]}
					& \scriptsize{[0.0302, 0.0473]}
					& \scriptsize{[0.0747, 0.0967]}
					& \scriptsize{[0.0377, 0.0570]}
					& \scriptsize{[0.0389, 0.0593]}
					& \scriptsize{[0.0468, 0.0631]} \\
					\thinhline
					
					\multirow{2}{*}{MedGemma-4B}
					& 0.1443 & 0.0311 & 0.0580 & 0.0340 & 0.0317 & 0.0387 \\
					& \scriptsize{[0.1403, 0.1481]}
					& \scriptsize{[0.0231, 0.0397]}
					& \scriptsize{[0.0461, 0.0699]}
					& \scriptsize{[0.0258, 0.0437]}
					& \scriptsize{[0.0231, 0.0407]}
					& \scriptsize{[0.0312, 0.0464]} \\
					\thinhline
					
					\multirow{2}{*}{R2GenGPT}
					& 0.2907 & 0.2830 & 0.3719 & 0.3139 & 0.3130 & 0.3204 \\
					& \scriptsize{[0.2805, 0.3030]}
					& \scriptsize{[0.2695, 0.2972]}
					& \scriptsize{[0.3560, 0.3870]}
					& \scriptsize{[0.2976, 0.3310]}
					& \scriptsize{[0.2975, 0.3294]}
					& \scriptsize{[0.3088, 0.3326]} \\
					\thinhline
					
					\multirow{2}{*}{LLaVA}
					& 0.2584 & 0.2476 & 0.3278 & 0.2800 & 0.2716 & 0.2818 \\
					& \scriptsize{[0.2487, 0.2673]}
					& \scriptsize{[0.2320, 0.2630]}
					& \scriptsize{[0.3088, 0.3462]}
					& \scriptsize{[0.2608, 0.2989]}
					& \scriptsize{[0.2542, 0.2887]}
					& \scriptsize{[0.2670, 0.2960]} \\
					
					\bottomrule
				\end{tabular}%
			}
			\centering
			\caption{\textbf{Fine-grained caption metrics (RaTE-Score, FORTE) on the Inspect dataset} ($n = 1{,}000$ cases; 95\% confidence intervals estimated via 2{,}000 bootstrap iterations). $^*$ represents a significant improvement between Astra and the second-best baseline with $P < 0.05$, otherwise shows a non-significant improvement. F. means FORTE metric.}
		\end{table*}

		\begin{table*}[htbp]
			\label{tab:bimcv_nlg}
			\resizebox{\textwidth}{!}{%
				\begin{tabular}{lcccccc}
					\toprule
					\multirow{2}{*}{\textbf{Model}}
					& \textbf{BLEU-1} & \textbf{BLEU-2} & \textbf{BLEU-3} & \textbf{BLEU-4}
					& \textbf{ROUGE-L} & \textbf{METEOR} \\
					& \scriptsize{[95\% CI]} & \scriptsize{[95\% CI]} & \scriptsize{[95\% CI]} & \scriptsize{[95\% CI]}
					& \scriptsize{[95\% CI]} & \scriptsize{[95\% CI]} \\
					\midrule
					
					\multirow{2}{*}{Astra}
					& 0.4020$^*$ & 0.3032$^*$ & 0.2536$^*$ & 0.2206$^*$ & 0.3988 & 0.1968$^*$ \\
					& \scriptsize{[0.3914, 0.4117]}
					& \scriptsize{[0.2941, 0.3116]}
					& \scriptsize{[0.2456, 0.2612]}
					& \scriptsize{[0.2132, 0.2278]}
					& \scriptsize{[0.3915, 0.4058]}
					& \scriptsize{[0.1922, 0.2013]} \\
					\thinhline
					
					\multirow{2}{*}{Gemini-3}
					& 0.2749 & 0.2196 & 0.1909 & 0.1699 & 0.3903 & 0.1722 \\
					& \scriptsize{[0.2619, 0.2879]}
					& \scriptsize{[0.2089, 0.2303]}
					& \scriptsize{[0.1813, 0.2004]}
					& \scriptsize{[0.1613, 0.1786]}
					& \scriptsize{[0.3811, 0.3995]}
					& \scriptsize{[0.1679, 0.1767]} \\
					\thinhline
					
					\multirow{2}{*}{Qwen3-VL-8B}
					& 0.3272 & 0.2545 & 0.2198 & 0.1957 & 0.3655 & 0.1695 \\
					& \scriptsize{[0.3163, 0.3376]}
					& \scriptsize{[0.2455, 0.2633]}
					& \scriptsize{[0.2115, 0.2279]}
					& \scriptsize{[0.1881, 0.2032]}
					& \scriptsize{[0.3570, 0.3747]}
					& \scriptsize{[0.1654, 0.1737]} \\
					\thinhline
					
					\multirow{2}{*}{HuluMed-32B}
					& 0.3460 & 0.2613 & 0.2195 & 0.1907 & 0.3370 & 0.1700 \\
					& \scriptsize{[0.3365, 0.3552]}
					& \scriptsize{[0.2536, 0.2687]}
					& \scriptsize{[0.2127, 0.2263]}
					& \scriptsize{[0.1845, 0.1971]}
					& \scriptsize{[0.3306, 0.3433]}
					& \scriptsize{[0.1664, 0.1737]} \\
					\thinhline
					
					\multirow{2}{*}{HuluMed-14B}
					& 0.3466 & 0.2615 & 0.2197 & 0.1908 & 0.3374 & 0.1702 \\
					& \scriptsize{[0.3371, 0.3558]}
					& \scriptsize{[0.2539, 0.2690]}
					& \scriptsize{[0.2128, 0.2265]}
					& \scriptsize{[0.1846, 0.1972]}
					& \scriptsize{[0.3307, 0.3435]}
					& \scriptsize{[0.1665, 0.1739]} \\
					\thinhline
					
					\multirow{2}{*}{HuluMed-7B}
					& 0.3470 & 0.2618 & 0.2199 & 0.1911 & 0.3369 & 0.1703 \\
					& \scriptsize{[0.3377, 0.3560]}
					& \scriptsize{[0.2542, 0.2693]}
					& \scriptsize{[0.2131, 0.2267]}
					& \scriptsize{[0.1848, 0.1974]}
					& \scriptsize{[0.3304, 0.3432]}
					& \scriptsize{[0.1667, 0.1740]} \\
					\thinhline
					
					\multirow{2}{*}{M3D}
					& 0.2957 & 0.2346 & 0.2038 & 0.1816 & 0.3590 & 0.1690 \\
					& \scriptsize{[0.2831, 0.3079]}
					& \scriptsize{[0.2241, 0.2448]}
					& \scriptsize{[0.1944, 0.2131]}
					& \scriptsize{[0.1731, 0.1901]}
					& \scriptsize{[0.3519, 0.3662]}
					& \scriptsize{[0.1648, 0.1734]} \\
					\thinhline
					
					\multirow{2}{*}{RadFM}
					& 0.2496 & 0.2063 & 0.1832 & 0.1653 & 0.3838 & 0.1698 \\
					& \scriptsize{[0.2360, 0.2635]}
					& \scriptsize{[0.1946, 0.2179]}
					& \scriptsize{[0.1727, 0.1937]}
					& \scriptsize{[0.1556, 0.1749]}
					& \scriptsize{[0.3754, 0.3918]}
					& \scriptsize{[0.1653, 0.1744]} \\
					\thinhline
					
					\multirow{2}{*}{Lingshu-32B}
					& 0.3435 & 0.2622 & 0.2228 & 0.1960 & 0.3340 & 0.1687 \\
					& \scriptsize{[0.3347, 0.3522]}
					& \scriptsize{[0.2546, 0.2698]}
					& \scriptsize{[0.2158, 0.2299]}
					& \scriptsize{[0.1894, 0.2025]}
					& \scriptsize{[0.3274, 0.3405]}
					& \scriptsize{[0.1650, 0.1724]} \\
					\thinhline
					
					\multirow{2}{*}{Lingshu-7B}
					& 0.1578 & 0.1372 & 0.1252 & 0.1154 & 0.4188 & 0.1702 \\
					& \scriptsize{[0.1445, 0.1709]}
					& \scriptsize{[0.1255, 0.1485]}
					& \scriptsize{[0.1146, 0.1355]}
					& \scriptsize{[0.1056, 0.1249]}
					& \scriptsize{[0.4088, 0.4286]}
					& \scriptsize{[0.1655, 0.1753]} \\
					\thinhline
					
					\multirow{2}{*}{MedGemma-27B}
					& 0.2247 & 0.1909 & 0.1721 & 0.1577 & 0.4037 & 0.1696 \\
					& \scriptsize{[0.2083, 0.2403]}
					& \scriptsize{[0.1775, 0.2035]}
					& \scriptsize{[0.1601, 0.1836]}
					& \scriptsize{[0.1467, 0.1682]}
					& \scriptsize{[0.3932, 0.4139]}
					& \scriptsize{[0.1648, 0.1744]} \\
					\thinhline
					
					\multirow{2}{*}{MedGemma-4B}
					& 0.1141 & 0.1028 & 0.0956 & 0.0894 & 0.4350 & 0.1732 \\
					& \scriptsize{[0.1035, 0.1253]}
					& \scriptsize{[0.0932, 0.1131]}
					& \scriptsize{[0.0865, 0.1053]}
					& \scriptsize{[0.0808, 0.0985]}
					& \scriptsize{[0.4243, 0.4454]}
					& \scriptsize{[0.1681, 0.1785]} \\
					\thinhline
					
					\multirow{2}{*}{R2GenGPT}
					& 0.2116 & 0.1790 & 0.1613 & 0.1478 & 0.4433 & 0.1835 \\
					& \scriptsize{[0.1960, 0.2280]}
					& \scriptsize{[0.1656, 0.1926]}
					& \scriptsize{[0.1493, 0.1736]}
					& \scriptsize{[0.1368, 0.1593]}
					& \scriptsize{[0.4327, 0.4530]}
					& \scriptsize{[0.1783, 0.1889]} \\
					\thinhline
					
					\multirow{2}{*}{LLaVA}
					& 0.3224 & 0.2567 & 0.2220 & 0.1974 & 0.4146 & 0.1804 \\
					& \scriptsize{[0.3080, 0.3357]}
					& \scriptsize{[0.2452, 0.2677]}
					& \scriptsize{[0.2118, 0.2317]}
					& \scriptsize{[0.1883, 0.2062]}
					& \scriptsize{[0.4057, 0.4239]}
					& \scriptsize{[0.1752, 0.1855]} \\
					
					\bottomrule
				\end{tabular}%
			}
			\centering
			\caption{\textbf{Natural language generation metrics on the BIMCV dataset} ($n = 1{,}505$ cases; 95\% confidence intervals estimated via 2{,}000 bootstrap iterations). $^*$ represents a significant improvement between Astra and the second-best baseline with $P < 0.05$, otherwise shows a non-significant improvement.}
		\end{table*}

		\clearpage
		\begin{table*}[htbp]
			\label{tab:bimcv_clinical_prf1}
			\begin{tabularx}{\textwidth}{l >{\centering\arraybackslash}X >{\centering\arraybackslash}X >{\centering\arraybackslash}X}
				\toprule
				\textbf{Model}
				& \textbf{micro Precision} & \textbf{micro Recall} & \textbf{micro F1} \\
				& \scriptsize{[95\% CI]} & \scriptsize{[95\% CI]} & \scriptsize{[95\% CI]} \\
				\midrule
				
				\multirow{2}{*}{Astra}
				& 0.3572$^*$ & 0.3405$^*$ & 0.3486$^*$ \\
				& \scriptsize{[0.3373, 0.3774]}
				& \scriptsize{[0.3222, 0.3578]}
				& \scriptsize{[0.3305, 0.3652]} \\
				\thinhline
				
				\multirow{2}{*}{Gemini-3}
				& 0.2884 & 0.2419 & 0.2631 \\
				& \scriptsize{[0.2688, 0.3098]}
				& \scriptsize{[0.2251, 0.2595]}
				& \scriptsize{[0.2461, 0.2816]} \\
				\thinhline
				
				\multirow{2}{*}{Qwen3-VL-8B}
				& 0.2449 & 0.1217 & 0.1626 \\
				& \scriptsize{[0.2207, 0.2698]}
				& \scriptsize{[0.1075, 0.1355]}
				& \scriptsize{[0.1450, 0.1792]} \\
				\thinhline
				
				\multirow{2}{*}{HuluMed-32B}
				& 0.1800 & 0.1241 & 0.1469 \\
				& \scriptsize{[0.1614, 0.1994]}
				& \scriptsize{[0.1103, 0.1381]}
				& \scriptsize{[0.1319, 0.1616]} \\
				\thinhline
				
				\multirow{2}{*}{HuluMed-14B}
				& 0.1790 & 0.1233 & 0.1460 \\
				& \scriptsize{[0.1606, 0.1980]}
				& \scriptsize{[0.1097, 0.1370]}
				& \scriptsize{[0.1311, 0.1607]} \\
				\thinhline
				
				\multirow{2}{*}{HuluMed-7B}
				& 0.1809 & 0.1244 & 0.1475 \\
				& \scriptsize{[0.1629, 0.1999]}
				& \scriptsize{[0.1105, 0.1382]}
				& \scriptsize{[0.1323, 0.1624]} \\
				\thinhline
				
				\multirow{2}{*}{M3D}
				& 0.1650 & 0.0905 & 0.1169 \\
				& \scriptsize{[0.1460, 0.1836]}
				& \scriptsize{[0.0797, 0.1015]}
				& \scriptsize{[0.1037, 0.1298]} \\
				\thinhline
				
				\multirow{2}{*}{RadFM}
				& 0.1830 & 0.0827 & 0.1139 \\
				& \scriptsize{[0.1607, 0.2057]}
				& \scriptsize{[0.0717, 0.0944]}
				& \scriptsize{[0.0994, 0.1287]} \\
				\thinhline
				
				\multirow{2}{*}{Lingshu-32B}
				& 0.1691 & 0.0858 & 0.1138 \\
				& \scriptsize{[0.1499, 0.1902]}
				& \scriptsize{[0.0750, 0.0978]}
				& \scriptsize{[0.1001, 0.1287]} \\
				\thinhline
				
				\multirow{2}{*}{Lingshu-7B}
				& 0.1777 & 0.0447 & 0.0714 \\
				& \scriptsize{[0.1491, 0.2067]}
				& \scriptsize{[0.0370, 0.0530]}
				& \scriptsize{[0.0594, 0.0839]} \\
				\thinhline
				
				\multirow{2}{*}{MedGemma-27B}
				& 0.1792 & 0.0140 & 0.0260 \\
				& \scriptsize{[0.1293, 0.2335]}
				& \scriptsize{[0.0096, 0.0194]}
				& \scriptsize{[0.0178, 0.0358]} \\
				\thinhline
				
				\multirow{2}{*}{MedGemma-4B}
				& 0.2429 & 0.0126 & 0.0239 \\
				& \scriptsize{[0.1704, 0.3158]}
				& \scriptsize{[0.0083, 0.0175]}
				& \scriptsize{[0.0159, 0.0329]} \\
				\thinhline
				
				\multirow{2}{*}{R2GenGPT}
				& 0.2942 & 0.1795 & 0.2229 \\
				& \scriptsize{[0.2717, 0.3167]}
				& \scriptsize{[0.1641, 0.1960]}
				& \scriptsize{[0.2054, 0.2406]} \\
				\thinhline
				
				\multirow{2}{*}{LLaVA}
				& 0.2694 & 0.2013 & 0.2304 \\
				& \scriptsize{[0.2493, 0.2885]}
				& \scriptsize{[0.1853, 0.2165]}
				& \scriptsize{[0.2137, 0.2462]} \\
				
				\bottomrule
			\end{tabularx}
			\centering
			\caption{\textbf{RadBERT classification performance (micro Precision, Recall, F1) on the BIMCV dataset} ($n = 1{,}505$ cases; 95\% confidence intervals estimated via 2{,}000 bootstrap iterations). $^*$ represents a significant improvement between Astra and the second-best baseline with $P < 0.05$, otherwise shows a non-significant improvement.}
		\end{table*}
		
		\clearpage
		\begin{table*}[htbp]
			\label{tab:bimcv_rate_forte}
			\resizebox{\textwidth}{!}{%
				\begin{tabular}{lcccccc}
					\toprule
					\multirow{2}{*}{\textbf{Model}}
					& \textbf{RaTE-Score} & \textbf{F.-Degree} & \textbf{F.-Landmark} & \textbf{F.-Feature} & \textbf{F.-Impression} & \textbf{F.-Overall} \\
					& \scriptsize{[95\% CI]} & \scriptsize{[95\% CI]} & \scriptsize{[95\% CI]} & \scriptsize{[95\% CI]} & \scriptsize{[95\% CI]} & \scriptsize{[95\% CI]} \\
					\midrule
					
					\multirow{2}{*}{Astra}
					& 0.2624$^*$ & 0.4133$^*$ & 0.3783$^*$ & 0.2755$^*$ & 0.3448$^*$ & 0.3530$^*$ \\
					& \scriptsize{[0.2567, 0.2684]}
					& \scriptsize{[0.4023, 0.4240]}
					& \scriptsize{[0.3663, 0.3897]}
					& \scriptsize{[0.2649, 0.2863]}
					& \scriptsize{[0.3294, 0.3595]}
					& \scriptsize{[0.3441, 0.3620]} \\
					\thinhline
					
					\multirow{2}{*}{Gemini-3}
					& 0.2076 & 0.2286 & 0.2066 & 0.1586 & 0.1596 & 0.1884 \\
					& \scriptsize{[0.2035, 0.2115]}
					& \scriptsize{[0.2184, 0.2398]}
					& \scriptsize{[0.1957, 0.2167]}
					& \scriptsize{[0.1482, 0.1686]}
					& \scriptsize{[0.1471, 0.1720]}
					& \scriptsize{[0.1802, 0.1963]} \\
					\thinhline
					
					\multirow{2}{*}{Qwen3-VL-8B}
					& 0.1954 & 0.1683 & 0.1669 & 0.1041 & 0.1043 & 0.1359 \\
					& \scriptsize{[0.1918, 0.1990]}
					& \scriptsize{[0.1586, 0.1786]}
					& \scriptsize{[0.1572, 0.1770]}
					& \scriptsize{[0.0957, 0.1127]}
					& \scriptsize{[0.0940, 0.1141]}
					& \scriptsize{[0.1287, 0.1429]} \\
					\thinhline
					
					\multirow{2}{*}{HuluMed-32B}
					& 0.1904 & 0.1728 & 0.1747 & 0.0836 & 0.0814 & 0.1281 \\
					& \scriptsize{[0.1870, 0.1939]}
					& \scriptsize{[0.1632, 0.1833]}
					& \scriptsize{[0.1643, 0.1849]}
					& \scriptsize{[0.0767, 0.0906]}
					& \scriptsize{[0.0712, 0.0910]}
					& \scriptsize{[0.1214, 0.1350]} \\
					\thinhline
					
					\multirow{2}{*}{HuluMed-14B}
					& 0.1904 & 0.1730 & 0.1753 & 0.0832 & 0.0819 & 0.1284 \\
					& \scriptsize{[0.1870, 0.1939]}
					& \scriptsize{[0.1630, 0.1836]}
					& \scriptsize{[0.1647, 0.1855]}
					& \scriptsize{[0.0762, 0.0905]}
					& \scriptsize{[0.0717, 0.0913]}
					& \scriptsize{[0.1215, 0.1353]} \\
					\thinhline
					
					\multirow{2}{*}{HuluMed-7B}
					& 0.1907 & 0.1702 & 0.1736 & 0.0832 & 0.0817 & 0.1272 \\
					& \scriptsize{[0.1873, 0.1941]}
					& \scriptsize{[0.1603, 0.1805]}
					& \scriptsize{[0.1631, 0.1841]}
					& \scriptsize{[0.0762, 0.0902]}
					& \scriptsize{[0.0715, 0.0911]}
					& \scriptsize{[0.1204, 0.1339]} \\
					\thinhline
					
					\multirow{2}{*}{M3D}
					& 0.1803 & 0.1755 & 0.1552 & 0.0665 & 0.0547 & 0.1130 \\
					& \scriptsize{[0.1772, 0.1834]}
					& \scriptsize{[0.1651, 0.1860]}
					& \scriptsize{[0.1458, 0.1647]}
					& \scriptsize{[0.0598, 0.0730]}
					& \scriptsize{[0.0463, 0.0637]}
					& \scriptsize{[0.1066, 0.1189]} \\
					\thinhline
					
					\multirow{2}{*}{RadFM}
					& 0.1769 & 0.1280 & 0.1153 & 0.0729 & 0.0468 & 0.0907 \\
					& \scriptsize{[0.1739, 0.1798]}
					& \scriptsize{[0.1187, 0.1378]}
					& \scriptsize{[0.1060, 0.1253]}
					& \scriptsize{[0.0652, 0.0806]}
					& \scriptsize{[0.0385, 0.0548]}
					& \scriptsize{[0.0846, 0.0971]} \\
					\thinhline
					
					\multirow{2}{*}{Lingshu-32B}
					& 0.2010 & 0.1571 & 0.1709 & 0.0739 & 0.0780 & 0.1200 \\
					& \scriptsize{[0.1975, 0.2046]}
					& \scriptsize{[0.1481, 0.1665]}
					& \scriptsize{[0.1620, 0.1797]}
					& \scriptsize{[0.0672, 0.0804]}
					& \scriptsize{[0.0690, 0.0877]}
					& \scriptsize{[0.1140, 0.1262]} \\
					\thinhline
					
					\multirow{2}{*}{Lingshu-7B}
					& 0.1598 & 0.0927 & 0.0766 & 0.0394 & 0.0448 & 0.0634 \\
					& \scriptsize{[0.1566, 0.1630]}
					& \scriptsize{[0.0840, 0.1017]}
					& \scriptsize{[0.0679, 0.0855]}
					& \scriptsize{[0.0340, 0.0453]}
					& \scriptsize{[0.0361, 0.0534]}
					& \scriptsize{[0.0577, 0.0693]} \\
					\thinhline
					
					\multirow{2}{*}{MedGemma-27B}
					& 0.1479 & 0.0571 & 0.0500 & 0.0239 & 0.0277 & 0.0397 \\
					& \scriptsize{[0.1453, 0.1506]}
					& \scriptsize{[0.0492, 0.0650]}
					& \scriptsize{[0.0428, 0.0572]}
					& \scriptsize{[0.0190, 0.0293]}
					& \scriptsize{[0.0211, 0.0347]}
					& \scriptsize{[0.0345, 0.0449]} \\
					\thinhline
					
					\multirow{2}{*}{MedGemma-4B}
					& 0.1400 & 0.0417 & 0.0508 & 0.0175 & 0.0165 & 0.0317 \\
					& \scriptsize{[0.1374, 0.1428]}
					& \scriptsize{[0.0346, 0.0495]}
					& \scriptsize{[0.0427, 0.0596]}
					& \scriptsize{[0.0130, 0.0226]}
					& \scriptsize{[0.0113, 0.0224]}
					& \scriptsize{[0.0267, 0.0368]} \\
					\thinhline
					
					\multirow{2}{*}{R2GenGPT}
					& 0.1823 & 0.1947 & 0.1752 & 0.1284 & 0.1856 & 0.1710 \\
					& \scriptsize{[0.1782, 0.1862]}
					& \scriptsize{[0.1813, 0.2086]}
					& \scriptsize{[0.1635, 0.1870]}
					& \scriptsize{[0.1170, 0.1400]}
					& \scriptsize{[0.1694, 0.2029]}
					& \scriptsize{[0.1601, 0.1821]} \\
					\thinhline
					
					\multirow{2}{*}{LLaVA}
					& 0.2189 & 0.2636 & 0.2513 & 0.1430 & 0.1970 & 0.2137 \\
					& \scriptsize{[0.2143, 0.2235]}
					& \scriptsize{[0.2525, 0.2749]}
					& \scriptsize{[0.2396, 0.2632]}
					& \scriptsize{[0.1333, 0.1530]}
					& \scriptsize{[0.1817, 0.2114]}
					& \scriptsize{[0.2052, 0.2225]} \\
					
					\bottomrule
				\end{tabular}%
			}
			\centering
			\caption{\textbf{Fine-grained caption metrics (RaTE-Score, FORTE) on the BIMCV dataset} ($n = 1{,}505$ cases; 95\% confidence intervals estimated via 2{,}000 bootstrap iterations). $^*$ represents a significant improvement between Astra and the second-best baseline with $P < 0.05$, otherwise shows a non-significant improvement. F. means FORTE metric.}
		\end{table*}

		\clearpage
		\begin{table*}[htbp]
			\label{tab:ctrg_clinical_prf1}
			\begin{tabularx}{\textwidth}{l >{\centering\arraybackslash}X >{\centering\arraybackslash}X >{\centering\arraybackslash}X}
				\toprule
				\textbf{Model}
				& \textbf{micro Precision} & \textbf{micro Recall} & \textbf{micro F1} \\
				& \scriptsize{[95\% CI]} & \scriptsize{[95\% CI]} & \scriptsize{[95\% CI]} \\
				\midrule
				
				\multirow{2}{*}{Astra}
				& 0.6618 & 0.3203$^*$ & 0.4317$^*$ \\
				& \scriptsize{[0.5781, 0.7379]}
				& \scriptsize{[0.2684, 0.3742]}
				& \scriptsize{[0.3718, 0.4907]} \\
				\thinhline
				
				\multirow{2}{*}{Gemini-3}
				& 0.5529 & 0.1673 & 0.2568 \\
				& \scriptsize{[0.4456, 0.6566]}
				& \scriptsize{[0.1237, 0.2098]}
				& \scriptsize{[0.1954, 0.3137]} \\
				\thinhline
				
				\multirow{2}{*}{Qwen3-VL-8B}
				& 0.4000 & 0.0702 & 0.1194 \\
				& \scriptsize{[0.2667, 0.5349]}
				& \scriptsize{[0.0418, 0.1003]}
				& \scriptsize{[0.0734, 0.1667]} \\
				\thinhline
				
				\multirow{2}{*}{HuluMed-32B}
				& 0.5676 & 0.2989 & 0.3916 \\
				& \scriptsize{[0.4890, 0.6457]}
				& \scriptsize{[0.2474, 0.3522]}
				& \scriptsize{[0.3333, 0.4485]} \\
				\thinhline
				
				\multirow{2}{*}{HuluMed-14B}
				& 0.5060 & 0.1495 & 0.2308 \\
				& \scriptsize{[0.4024, 0.6087]}
				& \scriptsize{[0.1079, 0.1937]}
				& \scriptsize{[0.1717, 0.2881]} \\
				\thinhline
				
				\multirow{2}{*}{HuluMed-7B}
				& 0.4615 & 0.1068 & 0.1734 \\
				& \scriptsize{[0.3414, 0.5902]}
				& \scriptsize{[0.0722, 0.1444]}
				& \scriptsize{[0.1212, 0.2274]} \\
				\thinhline
				
				\multirow{2}{*}{M3D}
				& 0.4121 & 0.2669 & 0.3240 \\
				& \scriptsize{[0.3403, 0.4850]}
				& \scriptsize{[0.2162, 0.3185]}
				& \scriptsize{[0.2685, 0.3782]} \\
				\thinhline
				
				\multirow{2}{*}{RadFM}
				& 0.4483 & 0.0463 & 0.0839 \\
				& \scriptsize{[0.2727, 0.6471]}
				& \scriptsize{[0.0238, 0.0727]}
				& \scriptsize{[0.0444, 0.1286]} \\
				\thinhline
				
				\multirow{2}{*}{Lingshu-32B}
				& 0.5062 & 0.1439 & 0.2240 \\
				& \scriptsize{[0.3944, 0.6104]}
				& \scriptsize{[0.1018, 0.1851]}
				& \scriptsize{[0.1636, 0.2798]} \\
				\thinhline
				
				\multirow{2}{*}{Lingshu-7B}
				& 0.6154 & 0.1993 & 0.3011 \\
				& \scriptsize{[0.5195, 0.7093]}
				& \scriptsize{[0.1536, 0.2455]}
				& \scriptsize{[0.2402, 0.3602]} \\
				\thinhline
				
				\multirow{2}{*}{MedGemma-27B}
				& 0.6250 & 0.0356 & 0.0673 \\
				& \scriptsize{[0.4000, 0.8336]}
				& \scriptsize{[0.0146, 0.0590]}
				& \scriptsize{[0.0282, 0.1100]} \\
				\thinhline
				
				\multirow{2}{*}{MedGemma-4B}
				& 0.8000 & 0.0285 & 0.0550 \\
				& \scriptsize{[0.5000, 1.0000]}
				& \scriptsize{[0.0107, 0.0491]}
				& \scriptsize{[0.0211, 0.0930]} \\
				
				\bottomrule
			\end{tabularx}
			\centering
			\caption{\textbf{RadBERT classification performance (micro Precision, Recall, F1) on the external CTRG-Chest dataset} ($n = 324$ cases; 95\% confidence intervals estimated via 2{,}000 bootstrap iterations). $^*$ represents a significant improvement between Astra and the second-best baseline with $P < 0.05$, otherwise shows a non-significant improvement.}
		\end{table*}

		\clearpage
		\begin{table*}[htbp]
			\label{tab:ctrg_rate_forte}
			\resizebox{\textwidth}{!}{%
				\begin{tabular}{lcccccc}
					\toprule
					\multirow{2}{*}{\textbf{Model}}
					& \textbf{RaTE-Score} & \textbf{F.-Degree} & \textbf{F.-Landmark} & \textbf{F.-Feature} & \textbf{F.-Impression} & \textbf{F.-Overall} \\
					& \scriptsize{[95\% CI]} & \scriptsize{[95\% CI]} & \scriptsize{[95\% CI]} & \scriptsize{[95\% CI]} & \scriptsize{[95\% CI]} & \scriptsize{[95\% CI]} \\
					\midrule
					
					\multirow{2}{*}{Astra}
					& 0.2434 & 0.2550$^*$ & 0.3918$^*$ & 0.1408 & 0.1346 & 0.2306$^*$ \\
					& \scriptsize{[0.2348, 0.2525]}
					& \scriptsize{[0.2324, 0.2783]}
					& \scriptsize{[0.3690, 0.4131]}
					& \scriptsize{[0.1195, 0.1634]}
					& \scriptsize{[0.1063, 0.1649]}
					& \scriptsize{[0.2151, 0.2467]} \\
					\thinhline
					
					\multirow{2}{*}{Gemini-3}
					& 0.2386 & 0.2326 & 0.2693 & 0.1072 & 0.0962 & 0.1763 \\
					& \scriptsize{[0.2284, 0.2489]}
					& \scriptsize{[0.2096, 0.2542]}
					& \scriptsize{[0.2461, 0.2897]}
					& \scriptsize{[0.0844, 0.1302]}
					& \scriptsize{[0.0679, 0.1256]}
					& \scriptsize{[0.1582, 0.1938]} \\
					\thinhline
					
					\multirow{2}{*}{Qwen3-VL-8B}
					& 0.2149 & 0.1483 & 0.2199 & 0.0792 & 0.0588 & 0.1265 \\
					& \scriptsize{[0.2055, 0.2243]}
					& \scriptsize{[0.1258, 0.1710]}
					& \scriptsize{[0.1944, 0.2443]}
					& \scriptsize{[0.0582, 0.1008]}
					& \scriptsize{[0.0344, 0.0861]}
					& \scriptsize{[0.1080, 0.1443]} \\
					\thinhline
					
					\multirow{2}{*}{HuluMed-32B}
					& 0.2187 & 0.2056 & 0.2660 & 0.1466 & 0.1771 & 0.1988 \\
					& \scriptsize{[0.2102, 0.2279]}
					& \scriptsize{[0.1819, 0.2295]}
					& \scriptsize{[0.2376, 0.2929]}
					& \scriptsize{[0.1245, 0.1705]}
					& \scriptsize{[0.1418, 0.2129]}
					& \scriptsize{[0.1801, 0.2172]} \\
					\thinhline
					
					\multirow{2}{*}{HuluMed-14B}
					& 0.2148 & 0.1608 & 0.2148 & 0.1033 & 0.1216 & 0.1501 \\
					& \scriptsize{[0.2051, 0.2250]}
					& \scriptsize{[0.1373, 0.1873]}
					& \scriptsize{[0.1897, 0.2393]}
					& \scriptsize{[0.0763, 0.1317]}
					& \scriptsize{[0.0851, 0.1600]}
					& \scriptsize{[0.1276, 0.1742]} \\
					\thinhline
					
					\multirow{2}{*}{HuluMed-7B}
					& 0.2054 & 0.1620 & 0.2085 & 0.0663 & 0.0767 & 0.1284 \\
					& \scriptsize{[0.1977, 0.2138]}
					& \scriptsize{[0.1352, 0.1874]}
					& \scriptsize{[0.1850, 0.2327]}
					& \scriptsize{[0.0496, 0.0836]}
					& \scriptsize{[0.0475, 0.1070]}
					& \scriptsize{[0.1102, 0.1467]} \\
					\thinhline
					
					\multirow{2}{*}{M3D}
					& 0.2200 & 0.1925 & 0.2466 & 0.1019 & 0.1183 & 0.1648 \\
					& \scriptsize{[0.2116, 0.2286]}
					& \scriptsize{[0.1701, 0.2147]}
					& \scriptsize{[0.2230, 0.2704]}
					& \scriptsize{[0.0814, 0.1229]}
					& \scriptsize{[0.0860, 0.1539]}
					& \scriptsize{[0.1467, 0.1833]} \\
					\thinhline
					
					\multirow{2}{*}{RadFM}
					& 0.1997 & 0.0166 & 0.0525 & 0.0307 & 0.0220 & 0.0304 \\
					& \scriptsize{[0.1947, 0.2047]}
					& \scriptsize{[0.0086, 0.0254]}
					& \scriptsize{[0.0368, 0.0682]}
					& \scriptsize{[0.0163, 0.0459]}
					& \scriptsize{[0.0051, 0.0442]}
					& \scriptsize{[0.0198, 0.0420]} \\
					\thinhline
					
					\multirow{2}{*}{Lingshu-32B}
					& 0.2380 & 0.1656 & 0.2873 & 0.0969 & 0.1178 & 0.1669 \\
					& \scriptsize{[0.2295, 0.2473]}
					& \scriptsize{[0.1474, 0.1843]}
					& \scriptsize{[0.2662, 0.3083]}
					& \scriptsize{[0.0777, 0.1157]}
					& \scriptsize{[0.0855, 0.1510]}
					& \scriptsize{[0.1509, 0.1827]} \\
					\thinhline
					
					\multirow{2}{*}{Lingshu-7B}
					& 0.1977 & 0.1300 & 0.1566 & 0.1085 & 0.1996 & 0.1486 \\
					& \scriptsize{[0.1889, 0.2068]}
					& \scriptsize{[0.1063, 0.1554]}
					& \scriptsize{[0.1292, 0.1842]}
					& \scriptsize{[0.0862, 0.1326]}
					& \scriptsize{[0.1521, 0.2495]}
					& \scriptsize{[0.1239, 0.1742]} \\
					\thinhline
					
					\multirow{2}{*}{MedGemma-27B}
					& 0.1554 & 0.0408 & 0.0562 & 0.0322 & 0.0400 & 0.0423 \\
					& \scriptsize{[0.1495, 0.1619]}
					& \scriptsize{[0.0216, 0.0628]}
					& \scriptsize{[0.0335, 0.0829]}
					& \scriptsize{[0.0156, 0.0521]}
					& \scriptsize{[0.0155, 0.0686]}
					& \scriptsize{[0.0247, 0.0629]} \\
					\thinhline
					
					\multirow{2}{*}{MedGemma-4B}
					& 0.1586 & 0.0331 & 0.0624 & 0.0244 & 0.0119 & 0.0329 \\
					& \scriptsize{[0.1533, 0.1640]}
					& \scriptsize{[0.0181, 0.0506]}
					& \scriptsize{[0.0422, 0.0854]}
					& \scriptsize{[0.0118, 0.0380]}
					& \scriptsize{[0.0000, 0.0297]}
					& \scriptsize{[0.0219, 0.0462]} \\
					
					\bottomrule
				\end{tabular}%
			}
			\centering
			\caption{\textbf{Fine-grained caption metrics (RaTE-Score, FORTE) on the external CTRG-Chest dataset} ($n = 324$ cases; 95\% confidence intervals estimated via 2{,}000 bootstrap iterations). $^*$ represents a significant improvement between Astra and the second-best baseline with $P < 0.05$, otherwise shows a non-significant improvement. F. means FORTE metric.}
		\end{table*}

		\clearpage
		\begin{table*}[htbp]
			\label{tab:amos_clinical_prf1}
			\begin{tabularx}{\textwidth}{l >{\centering\arraybackslash}X >{\centering\arraybackslash}X >{\centering\arraybackslash}X}
				\toprule
				\textbf{Model}
				& \textbf{micro Precision} & \textbf{micro Recall} & \textbf{micro F1} \\
				& \scriptsize{[95\% CI]} & \scriptsize{[95\% CI]} & \scriptsize{[95\% CI]} \\
				\midrule
				
				\multirow{2}{*}{Astra}
				& 0.4051 & 0.4211$^*$ & 0.4129$^*$ \\
				& \scriptsize{[0.3590, 0.4534]}
				& \scriptsize{[0.3658, 0.4744]}
				& \scriptsize{[0.3694, 0.4569]} \\
				\thinhline
				
				\multirow{2}{*}{Gemini-3}
				& 0.3387 & 0.0553 & 0.0950 \\
				& \scriptsize{[0.2203, 0.4584]}
				& \scriptsize{[0.0332, 0.0788]}
				& \scriptsize{[0.0582, 0.1339]} \\
				\thinhline
				
				\multirow{2}{*}{Qwen3-VL-8B}
				& 0.1389 & 0.0526 & 0.0763 \\
				& \scriptsize{[0.0861, 0.1953]}
				& \scriptsize{[0.0312, 0.0755]}
				& \scriptsize{[0.0463, 0.1081]} \\
				\thinhline
				
				\multirow{2}{*}{HuluMed-32B}
				& 0.2949 & 0.1684 & 0.2144 \\
				& \scriptsize{[0.2335, 0.3580]}
				& \scriptsize{[0.1305, 0.2055]}
				& \scriptsize{[0.1696, 0.2573]} \\
				\thinhline
				
				\multirow{2}{*}{HuluMed-14B}
				& 0.2075 & 0.1158 & 0.1486 \\
				& \scriptsize{[0.1594, 0.2651]}
				& \scriptsize{[0.0879, 0.1470]}
				& \scriptsize{[0.1146, 0.1868]} \\
				\thinhline
				
				\multirow{2}{*}{HuluMed-7B}
				& 0.3392 & 0.1526 & 0.2105 \\
				& \scriptsize{[0.2697, 0.4214]}
				& \scriptsize{[0.1191, 0.1893]}
				& \scriptsize{[0.1667, 0.2575]} \\
				\thinhline
				
				\multirow{2}{*}{M3D}
				& 0.2230 & 0.0868 & 0.1250 \\
				& \scriptsize{[0.1600, 0.2883]}
				& \scriptsize{[0.0598, 0.1156]}
				& \scriptsize{[0.0880, 0.1621]} \\
				\thinhline
				
				\multirow{2}{*}{RadFM}
				& 0.1724 & 0.0132 & 0.0244 \\
				& \scriptsize{[0.0435, 0.3333]}
				& \scriptsize{[0.0027, 0.0261]}
				& \scriptsize{[0.0050, 0.0481]} \\
				\thinhline
				
				\multirow{2}{*}{Lingshu-32B}
				& 0.1985 & 0.0684 & 0.1018 \\
				& \scriptsize{[0.1357, 0.2696]}
				& \scriptsize{[0.0452, 0.0944]}
				& \scriptsize{[0.0672, 0.1387]} \\
				\thinhline
				
				\multirow{2}{*}{Lingshu-7B}
				& 0.4815 & 0.0342 & 0.0639 \\
				& \scriptsize{[0.3000, 0.6667]}
				& \scriptsize{[0.0183, 0.0526]}
				& \scriptsize{[0.0347, 0.0964]} \\
				\thinhline
				
				\multirow{2}{*}{MedGemma-27B}
				& 0.4286 & 0.0158 & 0.0305 \\
				& \scriptsize{[0.1538, 0.6923]}
				& \scriptsize{[0.0050, 0.0294]}
				& \scriptsize{[0.0097, 0.0557]} \\
				\thinhline
				
				\multirow{2}{*}{MedGemma-4B}
				& 0.4070 & 0.0921 & 0.1502 \\
				& \scriptsize{[0.3068, 0.5165]}
				& \scriptsize{[0.0643, 0.1250]}
				& \scriptsize{[0.1071, 0.1982]} \\
				
				\bottomrule
			\end{tabularx}
			\centering
			\caption{\textbf{RadBERT classification performance (micro Precision, Recall, F1) on the external AMOS-MM dataset} ($n = 400$ cases; 95\% confidence intervals estimated via 2{,}000 bootstrap iterations). $^*$ represents a significant improvement between Astra and the second-best baseline with $P < 0.05$, otherwise shows a non-significant improvement.}
		\end{table*}

		\clearpage
		\begin{table*}[htbp]
			\label{tab:amos_rate_forte}
			\resizebox{\textwidth}{!}{%
				\begin{tabular}{lcccccc}
					\toprule
					\multirow{2}{*}{\textbf{Model}}
					& \textbf{RaTE-Score} & \textbf{F.-Degree} & \textbf{F.-Landmark} & \textbf{F.-Feature} & \textbf{F.-Impression} & \textbf{F.-Overall} \\
					& \scriptsize{[95\% CI]} & \scriptsize{[95\% CI]} & \scriptsize{[95\% CI]} & \scriptsize{[95\% CI]} & \scriptsize{[95\% CI]} & \scriptsize{[95\% CI]} \\
					\midrule
					
					\multirow{2}{*}{Astra}
					& 0.2507$^*$ & 0.2880$^*$ & 0.3084$^*$ & 0.1973$^*$ & 0.1036$^*$ & 0.2243$^*$ \\
					& \scriptsize{[0.2449, 0.2561]}
					& \scriptsize{[0.2684, 0.3070]}
					& \scriptsize{[0.2904, 0.3258]}
					& \scriptsize{[0.1807, 0.2138]}
					& \scriptsize{[0.0826, 0.1258]}
					& \scriptsize{[0.2117, 0.2370]} \\
					\thinhline
					
					\multirow{2}{*}{Gemini-3}
					& 0.1847 & 0.0752 & 0.0977 & 0.0680 & 0.0429 & 0.0710 \\
					& \scriptsize{[0.1799, 0.1897]}
					& \scriptsize{[0.0578, 0.0930]}
					& \scriptsize{[0.0816, 0.1134]}
					& \scriptsize{[0.0536, 0.0819]}
					& \scriptsize{[0.0246, 0.0632]}
					& \scriptsize{[0.0584, 0.0833]} \\
					\thinhline
					
					\multirow{2}{*}{Qwen3-VL-8B}
					& 0.1872 & 0.0845 & 0.0895 & 0.0565 & 0.0240 & 0.0636 \\
					& \scriptsize{[0.1826, 0.1920]}
					& \scriptsize{[0.0683, 0.1012]}
					& \scriptsize{[0.0770, 0.1017]}
					& \scriptsize{[0.0453, 0.0680]}
					& \scriptsize{[0.0128, 0.0373]}
					& \scriptsize{[0.0543, 0.0728]} \\
					\thinhline
					
					\multirow{2}{*}{HuluMed-32B}
					& 0.1984 & 0.1662 & 0.1619 & 0.1154 & 0.0253 & 0.1172 \\
					& \scriptsize{[0.1932, 0.2040]}
					& \scriptsize{[0.1451, 0.1886]}
					& \scriptsize{[0.1420, 0.1820]}
					& \scriptsize{[0.0993, 0.1322]}
					& \scriptsize{[0.0082, 0.0466]}
					& \scriptsize{[0.1024, 0.1320]} \\
					\thinhline
					
					\multirow{2}{*}{HuluMed-14B}
					& 0.1894 & 0.1521 & 0.1284 & 0.0903 & 0.0282 & 0.0997 \\
					& \scriptsize{[0.1843, 0.1947]}
					& \scriptsize{[0.1312, 0.1720]}
					& \scriptsize{[0.1118, 0.1459]}
					& \scriptsize{[0.0742, 0.1073]}
					& \scriptsize{[0.0108, 0.0473]}
					& \scriptsize{[0.0866, 0.1133]} \\
					\thinhline
					
					\multirow{2}{*}{HuluMed-7B}
					& 0.1987 & 0.1518 & 0.1520 & 0.0949 & 0.0365 & 0.1088 \\
					& \scriptsize{[0.1932, 0.2043]}
					& \scriptsize{[0.1291, 0.1739]}
					& \scriptsize{[0.1329, 0.1724]}
					& \scriptsize{[0.0797, 0.1113]}
					& \scriptsize{[0.0200, 0.0556]}
					& \scriptsize{[0.0952, 0.1236]} \\
					\thinhline
					
					\multirow{2}{*}{M3D}
					& 0.1877 & 0.1021 & 0.1047 & 0.0726 & 0.0238 & 0.0758 \\
					& \scriptsize{[0.1835, 0.1919]}
					& \scriptsize{[0.0840, 0.1209]}
					& \scriptsize{[0.0897, 0.1194]}
					& \scriptsize{[0.0602, 0.0860]}
					& \scriptsize{[0.0080, 0.0426]}
					& \scriptsize{[0.0650, 0.0871]} \\
					\thinhline
					
					\multirow{2}{*}{RadFM}
					& 0.1745 & 0.0763 & 0.0741 & 0.0524 & 0.0132 & 0.0540 \\
					& \scriptsize{[0.1697, 0.1796]}
					& \scriptsize{[0.0604, 0.0928]}
					& \scriptsize{[0.0590, 0.0900]}
					& \scriptsize{[0.0406, 0.0642]}
					& \scriptsize{[0.0000, 0.0306]}
					& \scriptsize{[0.0440, 0.0647]} \\
					\thinhline
					
					\multirow{2}{*}{Lingshu-32B}
					& 0.1846 & 0.0672 & 0.0928 & 0.0621 & 0.0245 & 0.0617 \\
					& \scriptsize{[0.1804, 0.1887]}
					& \scriptsize{[0.0541, 0.0818]}
					& \scriptsize{[0.0777, 0.1085]}
					& \scriptsize{[0.0510, 0.0738]}
					& \scriptsize{[0.0103, 0.0396]}
					& \scriptsize{[0.0520, 0.0715]} \\
					\thinhline
					
					\multirow{2}{*}{Lingshu-7B}
					& 0.1614 & 0.0293 & 0.0337 & 0.0131 & 0.0135 & 0.0224 \\
					& \scriptsize{[0.1579, 0.1650]}
					& \scriptsize{[0.0170, 0.0424]}
					& \scriptsize{[0.0234, 0.0448]}
					& \scriptsize{[0.0072, 0.0197]}
					& \scriptsize{[0.0000, 0.0301]}
					& \scriptsize{[0.0146, 0.0317]} \\
					\thinhline
					
					\multirow{2}{*}{MedGemma-27B}
					& 0.1764 & 0.0329 & 0.0779 & 0.0436 & 0.0164 & 0.0427 \\
					& \scriptsize{[0.1722, 0.1804]}
					& \scriptsize{[0.0223, 0.0438]}
					& \scriptsize{[0.0636, 0.0914]}
					& \scriptsize{[0.0329, 0.0542]}
					& \scriptsize{[0.0028, 0.0320]}
					& \scriptsize{[0.0338, 0.0518]} \\
					\thinhline
					
					\multirow{2}{*}{MedGemma-4B}
					& 0.1800 & 0.0325 & 0.0860 & 0.0654 & 0.0249 & 0.0522 \\
					& \scriptsize{[0.1762, 0.1837]}
					& \scriptsize{[0.0223, 0.0430]}
					& \scriptsize{[0.0699, 0.1034]}
					& \scriptsize{[0.0522, 0.0798]}
					& \scriptsize{[0.0081, 0.0451]}
					& \scriptsize{[0.0415, 0.0632]} \\
					
					\bottomrule
				\end{tabular}%
			}
			\centering
			\caption{\textbf{Fine-grained caption metrics (RaTE-Score, FORTE) on the external AMOS-MM dataset} ($n = 400$ cases; 95\% confidence intervals estimated via 2{,}000 bootstrap iterations). $^*$ represents a significant improvement between Astra and the second-best baseline with $P < 0.05$, otherwise shows a non-significant improvement. F. means FORTE metric.}
		\end{table*}

		\clearpage
		\begin{table*}[htbp]
			\label{tab:inhouse_chest_clinical_prf1}
			\begin{tabularx}{\textwidth}{l >{\centering\arraybackslash}X >{\centering\arraybackslash}X >{\centering\arraybackslash}X}
				\toprule
				\textbf{Model}
				& \textbf{micro Precision} & \textbf{micro Recall} & \textbf{micro F1} \\
				& \scriptsize{[95\% CI]} & \scriptsize{[95\% CI]} & \scriptsize{[95\% CI]} \\
				\midrule
				
				\multirow{2}{*}{Astra}
				& 0.5675 & 0.4355$^*$ & 0.4928$^*$ \\
				& \scriptsize{[0.5361, 0.6002]}
				& \scriptsize{[0.4094, 0.4615]}
				& \scriptsize{[0.4679, 0.5176]} \\
				\thinhline
				
				\multirow{2}{*}{Gemini-3}
				& 0.2158 & 0.1076 & 0.1436 \\
				& \scriptsize{[0.1834, 0.2508]}
				& \scriptsize{[0.0908, 0.1267]}
				& \scriptsize{[0.1219, 0.1678]} \\
				\thinhline
				
				\multirow{2}{*}{Qwen3-VL-8B}
				& 0.3049 & 0.0601 & 0.1005 \\
				& \scriptsize{[0.2510, 0.3622]}
				& \scriptsize{[0.0451, 0.0762]}
				& \scriptsize{[0.0771, 0.1244]} \\
				\thinhline
				
				\multirow{2}{*}{HuluMed-32B}
				& 0.3467 & 0.2638 & 0.2996 \\
				& \scriptsize{[0.3153, 0.3778]}
				& \scriptsize{[0.2358, 0.2911]}
				& \scriptsize{[0.2728, 0.3265]} \\
				\thinhline
				
				\multirow{2}{*}{HuluMed-14B}
				& 0.3298 & 0.2999 & 0.3142 \\
				& \scriptsize{[0.3024, 0.3567]}
				& \scriptsize{[0.2710, 0.3288]}
				& \scriptsize{[0.2877, 0.3394]} \\
				\thinhline
				
				\multirow{2}{*}{HuluMed-7B}
				& 0.3488 & 0.1508 & 0.2105 \\
				& \scriptsize{[0.3089, 0.3867]}
				& \scriptsize{[0.1274, 0.1740]}
				& \scriptsize{[0.1818, 0.2378]} \\
				\thinhline
				
				\multirow{2}{*}{M3D}
				& 0.3481 & 0.1323 & 0.1917 \\
				& \scriptsize{[0.3054, 0.3907]}
				& \scriptsize{[0.1124, 0.1537]}
				& \scriptsize{[0.1664, 0.2187]} \\
				\thinhline
				
				\multirow{2}{*}{RadFM}
				& 0.6170 & 0.0930 & 0.1617 \\
				& \scriptsize{[0.5538, 0.6869]}
				& \scriptsize{[0.0782, 0.1094]}
				& \scriptsize{[0.1380, 0.1876]} \\
				\thinhline
				
				\multirow{2}{*}{Lingshu-32B}
				& 0.2735 & 0.0489 & 0.0830 \\
				& \scriptsize{[0.2136, 0.3391]}
				& \scriptsize{[0.0366, 0.0618]}
				& \scriptsize{[0.0629, 0.1040]} \\
				\thinhline
				
				\multirow{2}{*}{Lingshu-7B}
				& 0.2743 & 0.0497 & 0.0842 \\
				& \scriptsize{[0.2120, 0.3394]}
				& \scriptsize{[0.0371, 0.0628]}
				& \scriptsize{[0.0632, 0.1056]} \\
				\thinhline
				
				\multirow{2}{*}{MedGemma-27B}
				& 0.3529 & 0.0048 & 0.0095 \\
				& \scriptsize{[0.1429, 0.5385]}
				& \scriptsize{[0.0008, 0.0094]}
				& \scriptsize{[0.0016, 0.0184]} \\
				\thinhline
				
				\multirow{2}{*}{MedGemma-4B}
				& 0.3571 & 0.0080 & 0.0157 \\
				& \scriptsize{[0.1818, 0.5263]}
				& \scriptsize{[0.0034, 0.0130]}
				& \scriptsize{[0.0066, 0.0253]} \\
				
				\bottomrule
			\end{tabularx}
			\centering
			\caption{\textbf{RadBERT classification performance (micro Precision, Recall, F1) on the external Inhouse-Chest-1 dataset} ($n = 400$ cases; 95\% confidence intervals estimated via 2{,}000 bootstrap iterations). $^*$ represents a significant improvement between Astra and the second-best baseline with $P < 0.05$, otherwise shows a non-significant improvement.}
		\end{table*}
		
		\clearpage
		\begin{table*}[htbp]
			\label{tab:inhouse_chest1_rate_forte}
			\resizebox{\textwidth}{!}{%
				\begin{tabular}{lcccccc}
					\toprule
					\multirow{2}{*}{\textbf{Model}}
					& \textbf{RaTE-Score} & \textbf{F.-Degree} & \textbf{F.-Landmark} & \textbf{F.-Feature} & \textbf{F.-Impression} & \textbf{F.-Overall} \\
					& \scriptsize{[95\% CI]} & \scriptsize{[95\% CI]} & \scriptsize{[95\% CI]} & \scriptsize{[95\% CI]} & \scriptsize{[95\% CI]} & \scriptsize{[95\% CI]} \\
					\midrule
					
					\multirow{2}{*}{Astra}
					& 0.2874$^*$ & 0.2745$^*$ & 0.4805$^*$ & 0.2641$^*$ & 0.3227$^*$ & 0.3355$^*$ \\
					& \scriptsize{[0.2792, 0.2957]}
					& \scriptsize{[0.2605, 0.2890]}
					& \scriptsize{[0.4647, 0.4957]}
					& \scriptsize{[0.2489, 0.2788]}
					& \scriptsize{[0.2888, 0.3562]}
					& \scriptsize{[0.3231, 0.3473]} \\
					\thinhline
					
					\multirow{2}{*}{Gemini-3}
					& 0.2275 & 0.2086 & 0.2678 & 0.1006 & 0.0925 & 0.1674 \\
					& \scriptsize{[0.2219, 0.2334]}
					& \scriptsize{[0.1937, 0.2237]}
					& \scriptsize{[0.2522, 0.2839]}
					& \scriptsize{[0.0866, 0.1154]}
					& \scriptsize{[0.0690, 0.1167]}
					& \scriptsize{[0.1555, 0.1796]} \\
					\thinhline
					
					\multirow{2}{*}{Qwen3-VL-8B}
					& 0.1985 & 0.1060 & 0.1597 & 0.0647 & 0.0588 & 0.0973 \\
					& \scriptsize{[0.1921, 0.2059]}
					& \scriptsize{[0.0903, 0.1212]}
					& \scriptsize{[0.1404, 0.1796]}
					& \scriptsize{[0.0511, 0.0796]}
					& \scriptsize{[0.0371, 0.0829]}
					& \scriptsize{[0.0833, 0.1117]} \\
					\thinhline
					
					\multirow{2}{*}{HuluMed-32B}
					& 0.2313 & 0.1891 & 0.2863 & 0.1656 & 0.1717 & 0.2032 \\
					& \scriptsize{[0.2238, 0.2389]}
					& \scriptsize{[0.1737, 0.2047]}
					& \scriptsize{[0.2623, 0.3095]}
					& \scriptsize{[0.1498, 0.1810]}
					& \scriptsize{[0.1437, 0.1970]}
					& \scriptsize{[0.1885, 0.2166]} \\
					\thinhline
					
					\multirow{2}{*}{HuluMed-14B}
					& 0.2440 & 0.2086 & 0.3118 & 0.1845 & 0.1782 & 0.2208 \\
					& \scriptsize{[0.2367, 0.2514]}
					& \scriptsize{[0.1950, 0.2233]}
					& \scriptsize{[0.2895, 0.3331]}
					& \scriptsize{[0.1695, 0.1995]}
					& \scriptsize{[0.1520, 0.2039]}
					& \scriptsize{[0.2072, 0.2343]} \\
					\thinhline
					
					\multirow{2}{*}{HuluMed-7B}
					& 0.2156 & 0.1496 & 0.2366 & 0.1265 & 0.1324 & 0.1613 \\
					& \scriptsize{[0.2089, 0.2227]}
					& \scriptsize{[0.1314, 0.1680]}
					& \scriptsize{[0.2158, 0.2560]}
					& \scriptsize{[0.1096, 0.1432]}
					& \scriptsize{[0.1056, 0.1588]}
					& \scriptsize{[0.1440, 0.1768]} \\
					\thinhline
					
					\multirow{2}{*}{M3D}
					& 0.2037 & 0.1456 & 0.2111 & 0.0987 & 0.0986 & 0.1385 \\
					& \scriptsize{[0.1985, 0.2093]}
					& \scriptsize{[0.1299, 0.1611]}
					& \scriptsize{[0.1927, 0.2298]}
					& \scriptsize{[0.0832, 0.1146]}
					& \scriptsize{[0.0723, 0.1258]}
					& \scriptsize{[0.1248, 0.1527]} \\
					\thinhline
					
					\multirow{2}{*}{RadFM}
					& 0.1953 & 0.0826 & 0.1806 & 0.1004 & 0.1572 & 0.1302 \\
					& \scriptsize{[0.1904, 0.2006]}
					& \scriptsize{[0.0700, 0.0962]}
					& \scriptsize{[0.1620, 0.1988]}
					& \scriptsize{[0.0858, 0.1151]}
					& \scriptsize{[0.1215, 0.1934]}
					& \scriptsize{[0.1165, 0.1441]} \\
					\thinhline
					
					\multirow{2}{*}{Lingshu-32B}
					& 0.1741 & 0.0798 & 0.0653 & 0.0238 & 0.0326 & 0.0504 \\
					& \scriptsize{[0.1697, 0.1787]}
					& \scriptsize{[0.0690, 0.0914]}
					& \scriptsize{[0.0524, 0.0785]}
					& \scriptsize{[0.0163, 0.0314]}
					& \scriptsize{[0.0165, 0.0499]}
					& \scriptsize{[0.0419, 0.0595]} \\
					\thinhline
					
					\multirow{2}{*}{Lingshu-7B}
					& 0.1735 & 0.0786 & 0.0650 & 0.0238 & 0.0325 & 0.0500 \\
					& \scriptsize{[0.1693, 0.1782]}
					& \scriptsize{[0.0679, 0.0896]}
					& \scriptsize{[0.0522, 0.0779]}
					& \scriptsize{[0.0163, 0.0314]}
					& \scriptsize{[0.0165, 0.0497]}
					& \scriptsize{[0.0415, 0.0589]} \\
					\thinhline
					
					\multirow{2}{*}{MedGemma-27B}
					& 0.1541 & 0.0125 & 0.0217 & 0.0083 & 0.0109 & 0.0134 \\
					& \scriptsize{[0.1500, 0.1586]}
					& \scriptsize{[0.0067, 0.0193]}
					& \scriptsize{[0.0134, 0.0313]}
					& \scriptsize{[0.0026, 0.0149]}
					& \scriptsize{[0.0000, 0.0286]}
					& \scriptsize{[0.0071, 0.0213]} \\
					\thinhline
					
					\multirow{2}{*}{MedGemma-4B}
					& 0.1672 & 0.0380 & 0.0584 & 0.0220 & 0.0182 & 0.0341 \\
					& \scriptsize{[0.1624, 0.1725]}
					& \scriptsize{[0.0297, 0.0470]}
					& \scriptsize{[0.0454, 0.0715]}
					& \scriptsize{[0.0143, 0.0303]}
					& \scriptsize{[0.0037, 0.0344]}
					& \scriptsize{[0.0261, 0.0422]} \\
					
					\bottomrule
				\end{tabular}%
			}
			\centering
			\caption{\textbf{Fine-grained caption metrics (RaTE-Score, FORTE) on the external Inhouse-Chest-1 dataset} ($n = 400$ cases; 95\% confidence intervals estimated via 2{,}000 bootstrap iterations). $^*$ represents a significant improvement between Astra and the second-best baseline with $P < 0.05$, otherwise shows a non-significant improvement. F. means FORTE metric.}
		\end{table*}

		\clearpage
		\begin{table*}[htbp]
			\label{tab:inhouse_abdomen_clinical_prf1}
			\begin{tabularx}{\textwidth}{l >{\centering\arraybackslash}X >{\centering\arraybackslash}X >{\centering\arraybackslash}X}
				\toprule
				\textbf{Model}
				& \textbf{micro Precision} & \textbf{micro Recall} & \textbf{micro F1} \\
				& \scriptsize{[95\% CI]} & \scriptsize{[95\% CI]} & \scriptsize{[95\% CI]} \\
				\midrule
				
				\multirow{2}{*}{Astra}
				& 0.2453 & 0.5239$^*$ & 0.3342$^*$ \\
				& \scriptsize{[0.2263, 0.2654]}
				& \scriptsize{[0.4840, 0.5626]}
				& \scriptsize{[0.3109, 0.3575]} \\
				\thinhline
				
				\multirow{2}{*}{Gemini-3}
				& 0.0522 & 0.0279 & 0.0364 \\
				& \scriptsize{[0.0281, 0.0808]}
				& \scriptsize{[0.0143, 0.0439]}
				& \scriptsize{[0.0190, 0.0560]} \\
				\thinhline
				
				\multirow{2}{*}{Qwen3-VL-8B}
				& 0.0095 & 0.0100 & 0.0097 \\
				& \scriptsize{[0.0019, 0.0183]}
				& \scriptsize{[0.0020, 0.0192]}
				& \scriptsize{[0.0020, 0.0189]} \\
				\thinhline
				
				\multirow{2}{*}{HuluMed-32B}
				& 0.2165 & 0.1833 & 0.1985 \\
				& \scriptsize{[0.1816, 0.2511]}
				& \scriptsize{[0.1508, 0.2175]}
				& \scriptsize{[0.1654, 0.2304]} \\
				\thinhline
				
				\multirow{2}{*}{HuluMed-14B}
				& 0.0870 & 0.0837 & 0.0853 \\
				& \scriptsize{[0.0629, 0.1134]}
				& \scriptsize{[0.0609, 0.1078]}
				& \scriptsize{[0.0620, 0.1097]} \\
				\thinhline
				
				\multirow{2}{*}{HuluMed-7B}
				& 0.1119 & 0.0956 & 0.1031 \\
				& \scriptsize{[0.0842, 0.1415]}
				& \scriptsize{[0.0720, 0.1201]}
				& \scriptsize{[0.0779, 0.1286]} \\
				\thinhline
				
				\multirow{2}{*}{M3D}
				& 0.0914 & 0.0737 & 0.0816 \\
				& \scriptsize{[0.0640, 0.1190]}
				& \scriptsize{[0.0509, 0.0967]}
				& \scriptsize{[0.0570, 0.1061]} \\
				\thinhline
				
				\multirow{2}{*}{RadFM}
				& 0.0503 & 0.0179 & 0.0264 \\
				& \scriptsize{[0.0227, 0.0798]}
				& \scriptsize{[0.0079, 0.0306]}
				& \scriptsize{[0.0119, 0.0435]} \\
				\thinhline
				
				\multirow{2}{*}{Lingshu-32B}
				& 0.0603 & 0.0478 & 0.0533 \\
				& \scriptsize{[0.0382, 0.0842]}
				& \scriptsize{[0.0294, 0.0672]}
				& \scriptsize{[0.0333, 0.0747]} \\
				\thinhline
				
				\multirow{2}{*}{Lingshu-7B}
				& 0.0830 & 0.0398 & 0.0538 \\
				& \scriptsize{[0.0506, 0.1216]}
				& \scriptsize{[0.0235, 0.0589]}
				& \scriptsize{[0.0321, 0.0790]} \\
				\thinhline
				
				\multirow{2}{*}{MedGemma-27B}
				& 0.2105 & 0.0080 & 0.0154 \\
				& \scriptsize{[0.0556, 0.4000]}
				& \scriptsize{[0.0019, 0.0168]}
				& \scriptsize{[0.0037, 0.0323]} \\
				\thinhline
				
				\multirow{2}{*}{MedGemma-4B}
				& 0.4762 & 0.1992 & 0.2809 \\
				& \scriptsize{[0.4174, 0.5397]}
				& \scriptsize{[0.1651, 0.2343]}
				& \scriptsize{[0.2397, 0.3234]} \\
				
				\bottomrule
			\end{tabularx}
			\centering
			\caption{\textbf{RadBERT classification performance (micro Precision, Recall, F1) on the external Inhouse-Abdomen-1 dataset} ($n = 400$ cases; 95\% confidence intervals estimated via 2{,}000 bootstrap iterations). $^*$ represents a significant improvement between Astra and the second-best baseline with $P < 0.05$, otherwise shows a non-significant improvement.}
		\end{table*}

		\clearpage
		\begin{table*}[htbp]
			\label{tab:inhouse_abdomen1_rate_forte}
			\resizebox{\textwidth}{!}{%
				\begin{tabular}{lcccccc}
					\toprule
					\multirow{2}{*}{\textbf{Model}}
					& \textbf{RaTE-Score} & \textbf{F.-Degree} & \textbf{F.-Landmark} & \textbf{F.-Feature} & \textbf{F.-Impression} & \textbf{F.-Overall} \\
					& \scriptsize{[95\% CI]} & \scriptsize{[95\% CI]} & \scriptsize{[95\% CI]} & \scriptsize{[95\% CI]} & \scriptsize{[95\% CI]} & \scriptsize{[95\% CI]} \\
					\midrule
					
					\multirow{2}{*}{Astra}
					& 0.2509$^*$ & 0.3029$^*$ & 0.3500$^*$ & 0.2169$^*$ & 0.0631 & 0.2332$^*$ \\
					& \scriptsize{[0.2452, 0.2564]}
					& \scriptsize{[0.2883, 0.3185]}
					& \scriptsize{[0.3345, 0.3664]}
					& \scriptsize{[0.2020, 0.2308]}
					& \scriptsize{[0.0471, 0.0807]}
					& \scriptsize{[0.2243, 0.2422]} \\
					\thinhline
					
					\multirow{2}{*}{Gemini-3}
					& 0.1659 & 0.0391 & 0.0558 & 0.0322 & 0.0141 & 0.0353 \\
					& \scriptsize{[0.1619, 0.1697]}
					& \scriptsize{[0.0285, 0.0508]}
					& \scriptsize{[0.0418, 0.0703]}
					& \scriptsize{[0.0222, 0.0420]}
					& \scriptsize{[0.0027, 0.0296]}
					& \scriptsize{[0.0269, 0.0440]} \\
					\thinhline
					
					\multirow{2}{*}{Qwen3-VL-8B}
					& 0.1699 & 0.0405 & 0.0631 & 0.0285 & 0.0000 & 0.0331 \\
					& \scriptsize{[0.1659, 0.1739]}
					& \scriptsize{[0.0301, 0.0523]}
					& \scriptsize{[0.0532, 0.0739]}
					& \scriptsize{[0.0220, 0.0357]}
					& \scriptsize{[0.0000, 0.0000]}
					& \scriptsize{[0.0278, 0.0390]} \\
					\thinhline
					
					\multirow{2}{*}{HuluMed-32B}
					& 0.2035 & 0.1684 & 0.1722 & 0.1396 & 0.0598 & 0.1350 \\
					& \scriptsize{[0.1973, 0.2098]}
					& \scriptsize{[0.1492, 0.1885]}
					& \scriptsize{[0.1530, 0.1918]}
					& \scriptsize{[0.1209, 0.1583]}
					& \scriptsize{[0.0361, 0.0850]}
					& \scriptsize{[0.1202, 0.1504]} \\
					\thinhline
					
					\multirow{2}{*}{HuluMed-14B}
					& 0.1626 & 0.0502 & 0.0546 & 0.0414 & 0.0134 & 0.0399 \\
					& \scriptsize{[0.1586, 0.1669]}
					& \scriptsize{[0.0369, 0.0640]}
					& \scriptsize{[0.0430, 0.0670]}
					& \scriptsize{[0.0298, 0.0531]}
					& \scriptsize{[0.0000, 0.0312]}
					& \scriptsize{[0.0308, 0.0497]} \\
					\thinhline
					
					\multirow{2}{*}{HuluMed-7B}
					& 0.1853 & 0.1032 & 0.1330 & 0.0780 & 0.0365 & 0.0877 \\
					& \scriptsize{[0.1806, 0.1900]}
					& \scriptsize{[0.0862, 0.1212]}
					& \scriptsize{[0.1164, 0.1495]}
					& \scriptsize{[0.0641, 0.0920]}
					& \scriptsize{[0.0174, 0.0588]}
					& \scriptsize{[0.0757, 0.1002]} \\
					\thinhline
					
					\multirow{2}{*}{M3D}
					& 0.1840 & 0.0944 & 0.1145 & 0.0615 & 0.0260 & 0.0741 \\
					& \scriptsize{[0.1790, 0.1888]}
					& \scriptsize{[0.0776, 0.1125]}
					& \scriptsize{[0.0979, 0.1305]}
					& \scriptsize{[0.0497, 0.0740]}
					& \scriptsize{[0.0113, 0.0430]}
					& \scriptsize{[0.0633, 0.0855]} \\
					\thinhline
					
					\multirow{2}{*}{RadFM}
					& 0.1513 & 0.0133 & 0.0268 & 0.0266 & 0.0155 & 0.0205 \\
					& \scriptsize{[0.1477, 0.1549]}
					& \scriptsize{[0.0061, 0.0217]}
					& \scriptsize{[0.0178, 0.0357]}
					& \scriptsize{[0.0169, 0.0370]}
					& \scriptsize{[0.0000, 0.0332]}
					& \scriptsize{[0.0123, 0.0299]} \\
					\thinhline
					
					\multirow{2}{*}{Lingshu-32B}
					& 0.1841 & 0.0594 & 0.0921 & 0.0656 & 0.0118 & 0.0572 \\
					& \scriptsize{[0.1799, 0.1887]}
					& \scriptsize{[0.0464, 0.0728]}
					& \scriptsize{[0.0783, 0.1066]}
					& \scriptsize{[0.0545, 0.0776]}
					& \scriptsize{[0.0024, 0.0229]}
					& \scriptsize{[0.0494, 0.0654]} \\
					\thinhline
					
					\multirow{2}{*}{Lingshu-7B}
					& 0.1605 & 0.0247 & 0.0474 & 0.0174 & 0.0000 & 0.0224 \\
					& \scriptsize{[0.1572, 0.1639]}
					& \scriptsize{[0.0159, 0.0344]}
					& \scriptsize{[0.0348, 0.0600]}
					& \scriptsize{[0.0103, 0.0257]}
					& \scriptsize{[0.0000, 0.0000]}
					& \scriptsize{[0.0169, 0.0287]} \\
					\thinhline
					
					\multirow{2}{*}{MedGemma-27B}
					& 0.1871 & 0.0315 & 0.1184 & 0.0373 & 0.0252 & 0.0531 \\
					& \scriptsize{[0.1818, 0.1924]}
					& \scriptsize{[0.0200, 0.0441]}
					& \scriptsize{[0.1021, 0.1348]}
					& \scriptsize{[0.0263, 0.0489]}
					& \scriptsize{[0.0000, 0.0580]}
					& \scriptsize{[0.0423, 0.0647]} \\
					\thinhline
					
					\multirow{2}{*}{MedGemma-4B}
					& 0.1876 & 0.0463 & 0.1456 & 0.0901 & 0.0218 & 0.0760 \\
					& \scriptsize{[0.1835, 0.1918]}
					& \scriptsize{[0.0351, 0.0583]}
					& \scriptsize{[0.1256, 0.1646]}
					& \scriptsize{[0.0755, 0.1050]}
					& \scriptsize{[0.0071, 0.0391]}
					& \scriptsize{[0.0653, 0.0865]} \\
					
					\bottomrule
				\end{tabular}%
			}
			\centering
			\caption{\textbf{Fine-grained caption metrics (RaTE-Score, FORTE) on the external Inhouse-Abdomen-1 dataset} ($n = 400$ cases; 95\% confidence intervals estimated via 2{,}000 bootstrap iterations). $^*$ represents a significant improvement between Astra and the second-best baseline with $P < 0.05$, otherwise shows a non-significant improvement. F. means FORTE metric.}
		\end{table*}

		\clearpage
		\begin{table*}[htbp]
			\label{tab:inhouse_abdomen2_clinical_prf1}
			\begin{tabularx}{\textwidth}{l >{\centering\arraybackslash}X >{\centering\arraybackslash}X >{\centering\arraybackslash}X}
				\toprule
				\textbf{Model}
				& \textbf{micro Precision} & \textbf{micro Recall} & \textbf{micro F1} \\
				& \scriptsize{[95\% CI]} & \scriptsize{[95\% CI]} & \scriptsize{[95\% CI]} \\
				\midrule
				
				\multirow{2}{*}{Astra}
				& 0.3659$^*$ & 0.7599$^*$ & 0.4939$^*$ \\
				& \scriptsize{[0.3437, 0.3885]}
				& \scriptsize{[0.7204, 0.7995]}
				& \scriptsize{[0.4706, 0.5167]} \\
				\thinhline
				
				\multirow{2}{*}{Gemini-3}
				& 0.1491 & 0.1119 & 0.1278 \\
				& \scriptsize{[0.1029, 0.1994]}
				& \scriptsize{[0.0769, 0.1502]}
				& \scriptsize{[0.0884, 0.1690]} \\
				\thinhline
				
				\multirow{2}{*}{Qwen3-VL-8B}
				& 0.0569 & 0.0723 & 0.0637 \\
				& \scriptsize{[0.0348, 0.0797]}
				& \scriptsize{[0.0438, 0.1002]}
				& \scriptsize{[0.0385, 0.0886]} \\
				\thinhline
				
				\multirow{2}{*}{HuluMed-32B}
				& 0.1890 & 0.1515 & 0.1682 \\
				& \scriptsize{[0.1527, 0.2267]}
				& \scriptsize{[0.1189, 0.1860]}
				& \scriptsize{[0.1342, 0.2020]} \\
				\thinhline
				
				\multirow{2}{*}{HuluMed-14B}
				& 0.1650 & 0.1538 & 0.1592 \\
				& \scriptsize{[0.1316, 0.1978]}
				& \scriptsize{[0.1190, 0.1879]}
				& \scriptsize{[0.1258, 0.1899]} \\
				\thinhline
				
				\multirow{2}{*}{HuluMed-7B}
				& 0.1178 & 0.1049 & 0.1110 \\
				& \scriptsize{[0.0859, 0.1503]}
				& \scriptsize{[0.0762, 0.1353]}
				& \scriptsize{[0.0813, 0.1410]} \\
				\thinhline
				
				\multirow{2}{*}{M3D}
				& 0.1278 & 0.1212 & 0.1244 \\
				& \scriptsize{[0.0974, 0.1624]}
				& \scriptsize{[0.0925, 0.1517]}
				& \scriptsize{[0.0951, 0.1562]} \\
				\thinhline
				
				\multirow{2}{*}{RadFM}
				& 0.0508 & 0.0210 & 0.0297 \\
				& \scriptsize{[0.0213, 0.0860]}
				& \scriptsize{[0.0091, 0.0341]}
				& \scriptsize{[0.0128, 0.0486]} \\
				\thinhline
				
				\multirow{2}{*}{Lingshu-32B}
				& 0.1656 & 0.1235 & 0.1415 \\
				& \scriptsize{[0.1278, 0.2046]}
				& \scriptsize{[0.0939, 0.1555]}
				& \scriptsize{[0.1093, 0.1739]} \\
				\thinhline
				
				\multirow{2}{*}{Lingshu-7B}
				& 0.1176 & 0.0420 & 0.0619 \\
				& \scriptsize{[0.0714, 0.1654]}
				& \scriptsize{[0.0242, 0.0603]}
				& \scriptsize{[0.0362, 0.0887]} \\
				\thinhline
				
				\multirow{2}{*}{MedGemma-27B}
				& 0.1768 & 0.0207 & 0.0371 \\
				& \scriptsize{[0.1355, 0.2128]}
				& \scriptsize{[0.0136, 0.0299]}
				& \scriptsize{[0.0247, 0.0524]} \\
				\thinhline
				
				\multirow{2}{*}{MedGemma-4B}
				& 0.2429 & 0.0396 & 0.0681 \\
				& \scriptsize{[0.1481, 0.3455]}
				& \scriptsize{[0.0221, 0.0590]}
				& \scriptsize{[0.0385, 0.0989]} \\
				
				\bottomrule
			\end{tabularx}
			\centering
			\caption{\textbf{RadBERT classification performance (micro Precision, Recall, F1) on the external Inhouse-Abdomen-2 dataset} ($n = 400$ cases; 95\% confidence intervals estimated via 2{,}000 bootstrap iterations). $^*$ represents a significant improvement between Astra and the second-best baseline with $P < 0.05$, otherwise shows a non-significant improvement.}
		\end{table*}
		
		\clearpage
		\begin{table*}[htbp]
			\label{tab:inhouse_abdomen2_rate_forte}
			\resizebox{\textwidth}{!}{%
				\begin{tabular}{lcccccc}
					\toprule
					\multirow{2}{*}{\textbf{Model}}
					& \textbf{RaTE-Score} & \textbf{F.-Degree} & \textbf{F.-Landmark} & \textbf{F.-Feature} & \textbf{F.-Impression} & \textbf{F.-Overall} \\
					& \scriptsize{[95\% CI]} & \scriptsize{[95\% CI]} & \scriptsize{[95\% CI]} & \scriptsize{[95\% CI]} & \scriptsize{[95\% CI]} & \scriptsize{[95\% CI]} \\
					\midrule
					
					\multirow{2}{*}{Astra}
					& 0.3157$^*$ & 0.4701$^*$ & 0.5270$^*$ & 0.4234$^*$ & 0.3871$^*$ & 0.4519$^*$ \\
					& \scriptsize{[0.3097, 0.3216]}
					& \scriptsize{[0.4538, 0.4869]}
					& \scriptsize{[0.5125, 0.5415]}
					& \scriptsize{[0.4078, 0.4403]}
					& \scriptsize{[0.3562, 0.4175]}
					& \scriptsize{[0.4426, 0.4613]} \\
					\thinhline
					
					\multirow{2}{*}{Gemini-3}
					& 0.1790 & 0.0947 & 0.1156 & 0.0788 & 0.0988 & 0.0970 \\
					& \scriptsize{[0.1732, 0.1848]}
					& \scriptsize{[0.0743, 0.1147]}
					& \scriptsize{[0.0969, 0.1334]}
					& \scriptsize{[0.0617, 0.0961]}
					& \scriptsize{[0.0663, 0.1284]}
					& \scriptsize{[0.0794, 0.1136]} \\
					\thinhline
					
					\multirow{2}{*}{Qwen3-VL-8B}
					& 0.1771 & 0.0599 & 0.1056 & 0.0630 & 0.0356 & 0.0660 \\
					& \scriptsize{[0.1712, 0.1830]}
					& \scriptsize{[0.0462, 0.0743]}
					& \scriptsize{[0.0889, 0.1224]}
					& \scriptsize{[0.0502, 0.0758]}
					& \scriptsize{[0.0218, 0.0499]}
					& \scriptsize{[0.0551, 0.0772]} \\
					\thinhline
					
					\multirow{2}{*}{HuluMed-32B}
					& 0.1967 & 0.1482 & 0.1810 & 0.1435 & 0.1042 & 0.1442 \\
					& \scriptsize{[0.1897, 0.2045]}
					& \scriptsize{[0.1246, 0.1727]}
					& \scriptsize{[0.1583, 0.2033]}
					& \scriptsize{[0.1193, 0.1689]}
					& \scriptsize{[0.0745, 0.1367]}
					& \scriptsize{[0.1239, 0.1649]} \\
					\thinhline
					
					\multirow{2}{*}{HuluMed-14B}
					& 0.1989 & 0.1585 & 0.1925 & 0.1489 & 0.1017 & 0.1504 \\
					& \scriptsize{[0.1916, 0.2063]}
					& \scriptsize{[0.1364, 0.1805]}
					& \scriptsize{[0.1700, 0.2146]}
					& \scriptsize{[0.1277, 0.1705]}
					& \scriptsize{[0.0727, 0.1320]}
					& \scriptsize{[0.1326, 0.1685]} \\
					\thinhline
					
					\multirow{2}{*}{HuluMed-7B}
					& 0.1989 & 0.1323 & 0.1856 & 0.1181 & 0.0907 & 0.1317 \\
					& \scriptsize{[0.1925, 0.2054]}
					& \scriptsize{[0.1128, 0.1515]}
					& \scriptsize{[0.1646, 0.2058]}
					& \scriptsize{[0.0998, 0.1376]}
					& \scriptsize{[0.0645, 0.1167]}
					& \scriptsize{[0.1164, 0.1466]} \\
					\thinhline
					
					\multirow{2}{*}{M3D}
					& 0.1802 & 0.1002 & 0.1132 & 0.0714 & 0.0449 & 0.0824 \\
					& \scriptsize{[0.1747, 0.1857]}
					& \scriptsize{[0.0816, 0.1201]}
					& \scriptsize{[0.0953, 0.1320]}
					& \scriptsize{[0.0567, 0.0866]}
					& \scriptsize{[0.0263, 0.0660]}
					& \scriptsize{[0.0698, 0.0959]} \\
					\thinhline
					
					\multirow{2}{*}{RadFM}
					& 0.1642 & 0.0735 & 0.0994 & 0.0520 & 0.0239 & 0.0622 \\
					& \scriptsize{[0.1584, 0.1707]}
					& \scriptsize{[0.0550, 0.0931]}
					& \scriptsize{[0.0800, 0.1193]}
					& \scriptsize{[0.0368, 0.0679]}
					& \scriptsize{[0.0073, 0.0426]}
					& \scriptsize{[0.0493, 0.0753]} \\
					\thinhline
					
					\multirow{2}{*}{Lingshu-32B}
					& 0.1843 & 0.0525 & 0.1466 & 0.0744 & 0.0493 & 0.0807 \\
					& \scriptsize{[0.1793, 0.1898]}
					& \scriptsize{[0.0393, 0.0670]}
					& \scriptsize{[0.1258, 0.1670]}
					& \scriptsize{[0.0594, 0.0898]}
					& \scriptsize{[0.0311, 0.0681]}
					& \scriptsize{[0.0689, 0.0929]} \\
					\thinhline
					
					\multirow{2}{*}{Lingshu-7B}
					& 0.1666 & 0.0505 & 0.0921 & 0.0447 & 0.0255 & 0.0532 \\
					& \scriptsize{[0.1609, 0.1724]}
					& \scriptsize{[0.0344, 0.0678]}
					& \scriptsize{[0.0751, 0.1095]}
					& \scriptsize{[0.0310, 0.0583]}
					& \scriptsize{[0.0074, 0.0484]}
					& \scriptsize{[0.0409, 0.0668]} \\
					\thinhline
					
					\multirow{2}{*}{MedGemma-27B}
					& 0.1554 & 0.0084 & 0.0371 & 0.0202 & 0.0087 & 0.0186 \\
					& \scriptsize{[0.1515, 0.1597]}
					& \scriptsize{[0.0019, 0.0166]}
					& \scriptsize{[0.0250, 0.0505]}
					& \scriptsize{[0.0091, 0.0323]}
					& \scriptsize{[0.0000, 0.0213]}
					& \scriptsize{[0.0109, 0.0272]} \\
					\thinhline
					
					\multirow{2}{*}{MedGemma-4B}
					& 0.1731 & 0.0349 & 0.1040 & 0.0744 & 0.0351 & 0.0621 \\
					& \scriptsize{[0.1676, 0.1792]}
					& \scriptsize{[0.0227, 0.0477]}
					& \scriptsize{[0.0816, 0.1270]}
					& \scriptsize{[0.0560, 0.0929]}
					& \scriptsize{[0.0154, 0.0584]}
					& \scriptsize{[0.0478, 0.0771]} \\
					
					\bottomrule
				\end{tabular}%
			}
			\centering
			\caption{\textbf{Fine-grained caption metrics (RaTE-Score, FORTE) on the external Inhouse-Abdomen-2 dataset} ($n = 400$ cases; 95\% confidence intervals estimated via 2{,}000 bootstrap iterations). $^*$ represents a significant improvement between Astra and the second-best baseline with $P < 0.05$, otherwise shows a non-significant improvement. F. means FORTE metric.}
		\end{table*}

		\clearpage
		\begin{table*}[htbp]
			\centering
			\label{tab:inhouse_abdomen3_clinical_prf1}
			\begin{tabularx}{\textwidth}{l >{\centering\arraybackslash}X >{\centering\arraybackslash}X >{\centering\arraybackslash}X}
				\toprule
				\textbf{Model}
				& \textbf{micro Precision} & \textbf{micro Recall} & \textbf{micro F1} \\
				& \scriptsize{[95\% CI]} & \scriptsize{[95\% CI]} & \scriptsize{[95\% CI]} \\
				\midrule
				
				\multirow{2}{*}{Astra}
				& 0.3811 & 0.6046$^*$ & 0.4675$^*$ \\
				& \scriptsize{[0.3576, 0.4041]}
				& \scriptsize{[0.5736, 0.6377]}
				& \scriptsize{[0.4443, 0.4898]} \\
				\thinhline
				
				\multirow{2}{*}{Gemini-3}
				& 0.1544 & 0.1207 & 0.1354 \\
				& \scriptsize{[0.1223, 0.1851]}
				& \scriptsize{[0.0964, 0.1449]}
				& \scriptsize{[0.1087, 0.1617]} \\
				\thinhline
				
				\multirow{2}{*}{Qwen3-VL-8B}
				& 0.0989 & 0.0706 & 0.0824 \\
				& \scriptsize{[0.0742, 0.1232]}
				& \scriptsize{[0.0524, 0.0890]}
				& \scriptsize{[0.0615, 0.1025]} \\
				\thinhline
				
				\multirow{2}{*}{HuluMed-32B}
				& 0.1906 & 0.1297 & 0.1543 \\
				& \scriptsize{[0.1567, 0.2256]}
				& \scriptsize{[0.1057, 0.1555]}
				& \scriptsize{[0.1269, 0.1821]} \\
				\thinhline
				
				\multirow{2}{*}{HuluMed-14B}
				& 0.1316 & 0.0988 & 0.1129 \\
				& \scriptsize{[0.1018, 0.1622]}
				& \scriptsize{[0.0766, 0.1216]}
				& \scriptsize{[0.0884, 0.1376]} \\
				\thinhline
				
				\multirow{2}{*}{HuluMed-7B}
				& 0.1670 & 0.1078 & 0.1310 \\
				& \scriptsize{[0.1333, 0.2008]}
				& \scriptsize{[0.0861, 0.1283]}
				& \scriptsize{[0.1058, 0.1555]} \\
				\thinhline
				
				\multirow{2}{*}{M3D}
				& 0.1497 & 0.0719 & 0.0971 \\
				& \scriptsize{[0.1140, 0.1847]}
				& \scriptsize{[0.0535, 0.0911]}
				& \scriptsize{[0.0733, 0.1216]} \\
				\thinhline
				
				\multirow{2}{*}{RadFM}
				& 0.0704 & 0.0180 & 0.0286 \\
				& \scriptsize{[0.0372, 0.1075]}
				& \scriptsize{[0.0092, 0.0277]}
				& \scriptsize{[0.0147, 0.0438]} \\
				\thinhline
				
				\multirow{2}{*}{Lingshu-32B}
				& 0.1241 & 0.0629 & 0.0835 \\
				& \scriptsize{[0.0923, 0.1562]}
				& \scriptsize{[0.0459, 0.0809]}
				& \scriptsize{[0.0616, 0.1054]} \\
				\thinhline
				
				\multirow{2}{*}{Lingshu-7B}
				& 0.1034 & 0.0270 & 0.0428 \\
				& \scriptsize{[0.0642, 0.1449]}
				& \scriptsize{[0.0165, 0.0379]}
				& \scriptsize{[0.0263, 0.0599]} \\
				\thinhline
				
				\multirow{2}{*}{MedGemma-27B}
				& 0.2222 & 0.0103 & 0.0196 \\
				& \scriptsize{[0.0909, 0.3500]}
				& \scriptsize{[0.0038, 0.0177]}
				& \scriptsize{[0.0074, 0.0336]} \\
				\thinhline
				
				\multirow{2}{*}{MedGemma-4B}
				& 0.5135 & 0.0244 & 0.0466 \\
				& \scriptsize{[0.3448, 0.6744]}
				& \scriptsize{[0.0138, 0.0365]}
				& \scriptsize{[0.0266, 0.0686]} \\
				
				\bottomrule
			\end{tabularx}
			\caption{\textbf{RadBERT classification performance (micro Precision, Recall, F1) on the external Inhouse-Abdomen-3 dataset} ($n = 400$ cases; 95\% confidence intervals estimated via 2{,}000 bootstrap iterations). $^*$ represents a significant improvement between Astra and the second-best baseline with $P < 0.05$, otherwise shows a non-significant improvement.}
		\end{table*}
		
		\clearpage
		\begin{table*}[htbp]
			\label{tab:inhouse_abdomen3_rate_forte}
			\resizebox{\textwidth}{!}{%
				\begin{tabular}{lcccccc}
					\toprule
					\multirow{2}{*}{\textbf{Model}}
					& \textbf{RaTE-Score} & \textbf{F.-Degree} & \textbf{F.-Landmark} & \textbf{F.-Feature} & \textbf{F.-Impression} & \textbf{F.-Overall} \\
					& \scriptsize{[95\% CI]} & \scriptsize{[95\% CI]} & \scriptsize{[95\% CI]} & \scriptsize{[95\% CI]} & \scriptsize{[95\% CI]} & \scriptsize{[95\% CI]} \\
					\midrule
					
					\multirow{2}{*}{Astra}
					& 0.3143$^*$ & 0.4053$^*$ & 0.4512$^*$ & 0.3215$^*$ & 0.2640$^*$ & 0.3605$^*$ \\
					& \scriptsize{[0.3077, 0.3216]}
					& \scriptsize{[0.3927, 0.4182]}
					& \scriptsize{[0.4376, 0.4650]}
					& \scriptsize{[0.3069, 0.3353]}
					& \scriptsize{[0.2352, 0.2916]}
					& \scriptsize{[0.3506, 0.3704]} \\
					\thinhline
					
					\multirow{2}{*}{Gemini-3}
					& 0.2034 & 0.1050 & 0.1289 & 0.0919 & 0.1012 & 0.1068 \\
					& \scriptsize{[0.1976, 0.2092]}
					& \scriptsize{[0.0911, 0.1201]}
					& \scriptsize{[0.1151, 0.1438]}
					& \scriptsize{[0.0798, 0.1043]}
					& \scriptsize{[0.0782, 0.1231]}
					& \scriptsize{[0.0942, 0.1191]} \\
					\thinhline
					
					\multirow{2}{*}{Qwen3-VL-8B}
					& 0.1938 & 0.0829 & 0.0929 & 0.0677 & 0.0277 & 0.0678 \\
					& \scriptsize{[0.1891, 0.1985]}
					& \scriptsize{[0.0694, 0.0966]}
					& \scriptsize{[0.0806, 0.1053]}
					& \scriptsize{[0.0579, 0.0771]}
					& \scriptsize{[0.0162, 0.0393]}
					& \scriptsize{[0.0591, 0.0757]} \\
					\thinhline
					
					\multirow{2}{*}{HuluMed-32B}
					& 0.2058 & 0.1536 & 0.1657 & 0.1140 & 0.0871 & 0.1301 \\
					& \scriptsize{[0.1998, 0.2122]}
					& \scriptsize{[0.1324, 0.1754]}
					& \scriptsize{[0.1457, 0.1849]}
					& \scriptsize{[0.0978, 0.1295]}
					& \scriptsize{[0.0608, 0.1139]}
					& \scriptsize{[0.1132, 0.1460]} \\
					\thinhline
					
					\multirow{2}{*}{HuluMed-14B}
					& 0.1925 & 0.1002 & 0.1215 & 0.0687 & 0.0628 & 0.0883 \\
					& \scriptsize{[0.1872, 0.1981]}
					& \scriptsize{[0.0844, 0.1172]}
					& \scriptsize{[0.1039, 0.1391]}
					& \scriptsize{[0.0555, 0.0819]}
					& \scriptsize{[0.0416, 0.0853]}
					& \scriptsize{[0.0748, 0.1021]} \\
					\thinhline
					
					\multirow{2}{*}{HuluMed-7B}
					& 0.2028 & 0.1212 & 0.1513 & 0.0891 & 0.0605 & 0.1055 \\
					& \scriptsize{[0.1972, 0.2079]}
					& \scriptsize{[0.1048, 0.1378]}
					& \scriptsize{[0.1355, 0.1665]}
					& \scriptsize{[0.0745, 0.1028]}
					& \scriptsize{[0.0398, 0.0798]}
					& \scriptsize{[0.0928, 0.1172]} \\
					\thinhline
					
					\multirow{2}{*}{M3D}
					& 0.1860 & 0.0941 & 0.1053 & 0.0691 & 0.0453 & 0.0785 \\
					& \scriptsize{[0.1818, 0.1903]}
					& \scriptsize{[0.0802, 0.1087]}
					& \scriptsize{[0.0925, 0.1183]}
					& \scriptsize{[0.0568, 0.0815]}
					& \scriptsize{[0.0274, 0.0656]}
					& \scriptsize{[0.0682, 0.0891]} \\
					\thinhline
					
					\multirow{2}{*}{RadFM}
					& 0.1690 & 0.0517 & 0.0555 & 0.0369 & 0.0171 & 0.0403 \\
					& \scriptsize{[0.1650, 0.1733]}
					& \scriptsize{[0.0397, 0.0641]}
					& \scriptsize{[0.0443, 0.0669]}
					& \scriptsize{[0.0273, 0.0480]}
					& \scriptsize{[0.0052, 0.0304]}
					& \scriptsize{[0.0323, 0.0489]} \\
					\thinhline
					
					\multirow{2}{*}{Lingshu-32B}
					& 0.1929 & 0.0648 & 0.1190 & 0.0646 & 0.0381 & 0.0716 \\
					& \scriptsize{[0.1886, 0.1978]}
					& \scriptsize{[0.0535, 0.0760]}
					& \scriptsize{[0.1044, 0.1341]}
					& \scriptsize{[0.0536, 0.0754]}
					& \scriptsize{[0.0227, 0.0545]}
					& \scriptsize{[0.0629, 0.0805]} \\
					\thinhline
					
					\multirow{2}{*}{Lingshu-7B}
					& 0.1734 & 0.0383 & 0.0591 & 0.0234 & 0.0149 & 0.0339 \\
					& \scriptsize{[0.1696, 0.1774]}
					& \scriptsize{[0.0283, 0.0496]}
					& \scriptsize{[0.0487, 0.0707]}
					& \scriptsize{[0.0164, 0.0308]}
					& \scriptsize{[0.0048, 0.0280]}
					& \scriptsize{[0.0275, 0.0414]} \\
					\thinhline
					
					\multirow{2}{*}{MedGemma-27B}
					& 0.1730 & 0.0313 & 0.0615 & 0.0320 & 0.0205 & 0.0363 \\
					& \scriptsize{[0.1692, 0.1773]}
					& \scriptsize{[0.0221, 0.0415]}
					& \scriptsize{[0.0498, 0.0739]}
					& \scriptsize{[0.0231, 0.0417]}
					& \scriptsize{[0.0078, 0.0357]}
					& \scriptsize{[0.0287, 0.0447]} \\
					\thinhline
					
					\multirow{2}{*}{MedGemma-4B}
					& 0.1689 & 0.0210 & 0.0429 & 0.0157 & 0.0181 & 0.0244 \\
					& \scriptsize{[0.1654, 0.1722]}
					& \scriptsize{[0.0126, 0.0308]}
					& \scriptsize{[0.0312, 0.0559]}
					& \scriptsize{[0.0093, 0.0235]}
					& \scriptsize{[0.0059, 0.0339]}
					& \scriptsize{[0.0169, 0.0329]} \\
					
					\bottomrule
				\end{tabular}%
			}
			\centering
			\caption{\textbf{Fine-grained caption metrics (RaTE-Score, FORTE) on the external Inhouse-Abdomen-3 dataset} ($n = 400$ cases; 95\% confidence intervals estimated via 2{,}000 bootstrap iterations). $^*$ represents a significant improvement between Astra and the second-best baseline with $P < 0.05$, otherwise shows a non-significant improvement. F. means FORTE metric.}
		\end{table*}

	\end{appendices}
	
\end{document}